\documentclass[10pt,twocolumn,letterpaper]{article}

\usepackage{cvpr}              

\usepackage{multirow}
\usepackage{adjustbox}
\usepackage{bbm}
\usepackage{placeins}
\usepackage{balance}

\definecolor{cvprblue}{rgb}{0.21,0.49,0.74}
\usepackage[pagebackref,breaklinks,colorlinks,allcolors=cvprblue]{hyperref}
\usepackage[accsupp]{axessibility}
\usepackage{pifont}
\newcommand{\xmark}{\ding{55}}
\def\paperID{41286} 
\def\confName{CVPR}
\def\confYear{2026}

\title{Human Interaction-Aware 3D Reconstruction from a Single Image}

\author{
Gwanghyun Kim$^1$\thanks{Authors contributed equally. $^{\dagger}$Corresponding author.}  \quad Junghun James Kim$^2$\footnotemark[\value{footnote}]  \quad Suh Yoon Jeon$^1$\footnotemark[\value{footnote}] \quad Jason Park$^1$ \quad Se Young Chun$^{1,2,3\dagger}$ \\
  \normalfont $^1$Dept. of Electrical and Computer Engineering, $^2$IPAI, $^3$INMC \\
Seoul National University, Republic of Korea \\
{\tt\small \{gwang.kim, jonghean12,  euniejeon, jsp11235,  sychun\}@snu.ac.kr}
}

\begin{document}
\begin{figure*}
\twocolumn[{
\maketitle
\begin{center}
\vspace{-3em}
\includegraphics[width=0.95\textwidth]{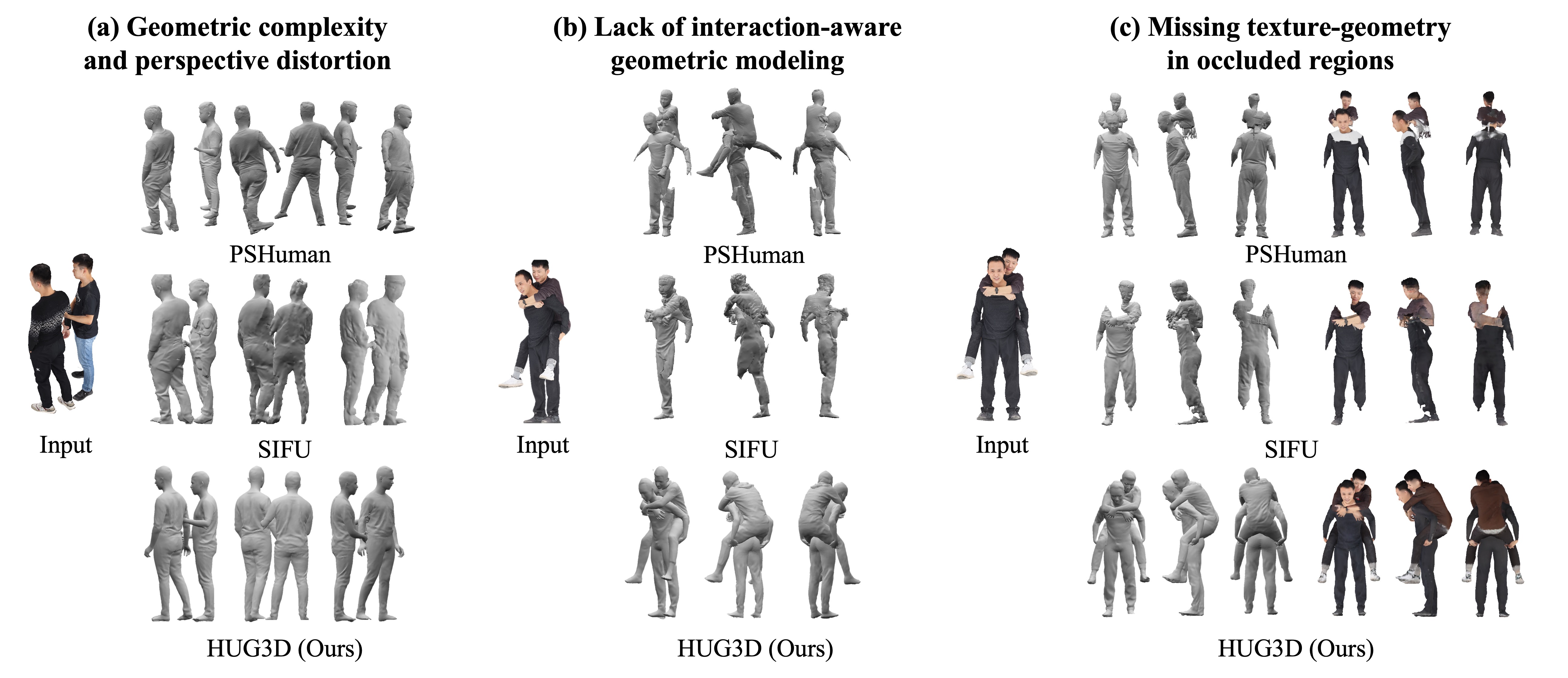}
\vspace{-1.em}
\caption{Core challenges in monocular multi-human 3D reconstruction: (a) geometric complexity and perspective distortion, (b) lack of interaction-aware geometric modeling, and 
(c) missing texture and geometry in occluded regions. HUG3D addresses all three challenges.}
\vspace{-0.em}
\label{fig_motivating}
\end{center}
}]
\end{figure*}
{
  \renewcommand{\thefootnote}%
    {\fnsymbol{footnote}}
    \footnotetext[1]{Authors contributed equally. \ $^\dagger$Corresponding author.}
  
}

\begin{abstract}
Reconstructing textured 3D human models from a single image is fundamental for AR/VR and digital human applications. However, existing methods mostly focus on single individuals and thus fail in multi-human scenes, where naive composition of individual reconstructions often leads to artifacts such as unrealistic overlaps, missing geometry in occluded regions, and distorted interactions. These limitations highlight the need for approaches that incorporate group-level context and interaction priors. We introduce a holistic method that explicitly models both group- and instance-level information. To mitigate perspective-induced geometric distortions, we first transform the input into a canonical orthographic space. Our primary component, Human Group-Instance Multi-View Diffusion (HUG-MVD), then generates complete multi-view normals and images by jointly modeling individuals and group context to resolve occlusions and proximity. Subsequently, the Human Group-Instance Geometric Reconstruction (HUG-GR) module optimizes the geometry by leveraging explicit, physics-based interaction priors to enforce physical plausibility and accurately model inter-human contact. Finally, the multi-view images are fused into a high-fidelity texture. Together, these components form our complete framework, HUG3D.
Extensive experiments show that HUG3D significantly outperforms both single-human and existing multi-human methods, producing physically plausible, high-fidelity 3D reconstructions of interacting people from a single image. Project page: \href{https://jongheean11.github.io/HUG3D_project}{\textcolor{magenta}{jongheean11.github.io/HUG3D\_project}}
\vspace{-1.5em}
\end{abstract}


\section{Introduction}
\label{sec:intro}

Reconstructing detailed 3D human models from visual input~\citep{ho2024sith, zhang2024sifu, jiang2024multiply, zheng2021deepmulticap, li2024pshuman, saito2019pifu, saito2020pifuhd, xiu2022icon, chupa} is a fundamental task in computer vision, supporting applications in augmented and virtual reality (AR/VR)~\citep{ma2021pixel, orts2016holoportation}, digital humans, and social behavior understanding. While monocular 3D reconstruction from a single RGB image~\citep{ho2024sith, zhang2024sifu, li2024pshuman} has made substantial progress, most existing methods are limited to isolated individuals in controlled environments. However, these approaches often fail to generalize to real-world scenes involving multiple interacting people, where occlusion, perspective distortion, and spatial entanglement introduce significant ambiguity and modeling challenges.

In particular, we identify three core challenges in monocular multi-human 3D reconstruction: 
\textbf{(1) Geometric complexity and perspective distortion.} 
Multi-human scenes often involve substantial depth variation and complex spatial layouts, which can lead to perspective distortions. While most methods assume orthographic views leading to distortions on real-world perspective inputs (Fig.~\ref{fig_motivating}(a)), a few perspective-aware models~\citep{li2024era3d, wang2025meat} remain limited to single-object cases and struggle with multi-human complexity. The scarcity of annotated multi-human data~\citep{yin2023hi4d, zheng2021deepmulticap} further hinders generalization across camera poses and interaction patterns.
\textbf{(2) Lack of interaction-aware geometric modeling.} 
Most methods reconstruct individuals independently, overlooking contextual cues like contact, occlusion, and spatial proximity (Fig.~\ref{fig_motivating}(b)). This often results in unrealistic outputs such as overlapping limbs or unnatural distances. Although group-wise SMPL-X approaches~\citep{baradel2024multi, muller2024generative, sun2022putting} offer early signs of interaction modeling, full-surface, textured reconstruction remains underexplored.
\textbf{(3) Missing geometry and texture in occluded regions.} 
Occlusions between people obscure critical body parts, causing incomplete geometry and textures (Fig.~\ref{fig_motivating}(c)). While generative inpainting~\citep{rombach2022high, flux2024} can hallucinate plausible content, only a few approaches~\citep{cao2024mvinpainter, barda2024instant3dit} jointly address geometry and appearance, and even fewer try handling multi-view consistency under occlusion.

To address these challenges, we propose a holistic method for human interaction-aware 3D reconstruction from a single image. Our method incorporates both group- and instance-level information with three main components: 
(1) \textit{Canonical Perspective-to-Orthographic View Transform (Pers2Ortho).} 
To make multi-view diffusion tractable under severe geometric distortion, we transform the input perspective image into a canonical orthographic space. 
From the image, we estimate a partial 3D textured geometry and reproject it to obtain consistent multi-view RGB and normal representations.
(2) \textit{Human Group-Instance Multi-View Diffusion (HUG-MVD).} This diffusion model jointly completes missing geometry and textures while serving as an implicit prior that enforces physically plausible human interactions. 
(3) \textit{Textured Mesh Reconstruction.} Our \textit{Human Group-Instance Geometry Reconstruction (HUG-GR)} module first optimizes the mesh using group-level, instance-level, and physics-based supervision. 
Subsequently, the multi-view images are fused into a final, high-fidelity texture using occlusion-aware blending. Collectively, these components form our holistic framework, \textbf{HUG3D}.

Experiments show that HUG3D outperforms baselines, producing physically plausible, high-fidelity textured reconstructions of interacting people from a single image.

\section{Related Works}
\label{sec:related}

\noindent \textbf{Single-Human 3D Reconstruction.}
Significant advances in single-human 3D reconstruction have been driven by parametric models such as SMPL~\citep{loper2015smpl,pavlakos2019expressive}. Early methods fit models to 2D cues~\citep{bogo2016keep}, while later works regress parameters directly from images using deep networks~\citep{kanazawa2018end}. Implicit and volumetric approaches~\citep{saito2019pifu,saito2020pifuhd,mustafa2021multi,fieraru2021remips} further improve geometric detail.
Recent efforts enhance view consistency and texture. SIFU~\citep{zhang2024sifu} optimizes UVs via SDF, and SiTH~\citep{ho2024sith} uses back-view synthesis. PSHuman~\citep{li2024pshuman} generates multi-view RGBs via diffusion, and LHM~\citep{qiu2025lhm} uses transformers over image and SMPL-X tokens.
Though effective for isolated individuals, these methods struggle in multi-human settings. Na\"ive application per instance leads to artifacts like overlapping meshes and inconsistent scale, underscoring the need for interaction-aware models.

\noindent \textbf{Multi-Human 3D Reconstruction.}  
Reconstructing multiple humans in 3D remains challenging due to occlusions, inter-person interactions, and depth ambiguities. Early approaches reconstructed each individual independently~\citep{li2019crowdpose,sun2022putting,zheng2021deepmulticap}, but this often led to physically implausible results. Later methods improved spatial coherence by introducing global constraints~\citep{hassan2021populating} or jointly regressing shapes and poses in a shared coordinate system~\citep{mustafa2021multi}. Multi-view and video-based approaches~\citep{kocabas2020vibe,jiang2024multiply,lee2024guess} further enhanced reconstruction accuracy, though they require video or multi-view inputs, which can be costly to acquire and process.  
More recently, learning-based methods have sought to tackle multi-human reconstruction from a single image. While interaction-aware networks~\citep{fieraru2021remips,cha20243d,chanrobust} and  model-free architectures~\citep{mustafa2021multi} improve plausibility, their performance remains limited in single-view, single-frame scenarios.

\noindent \textbf{Scene-Level Human Reconstruction and Interaction Modeling.}
Several methods incorporate human-scene or human-human interaction to improve physical plausibility. POSA~\citep{hassan2021populating} enforces realistic body-ground contact but focuses on single-person scenarios. Group-level priors~\citep{muller2024generative} enhance pose coherence, while BUDDI~\citep{muller2024generative} learns a diffusion-based prior for plausible two-person interactions. Other approaches rely on explicit contact labels~\citep{joo2018total}, but typically focus on coarse geometry or pose estimation. However, these methods often yield results with low texture fidelity or limited to SMPL-X mesh predictions, falling short of producing fully textured, detailed 3D reconstructions of multi-human interacting.

\begin{figure*}[!t]
     \centering
    \includegraphics[width=0.9\textwidth]{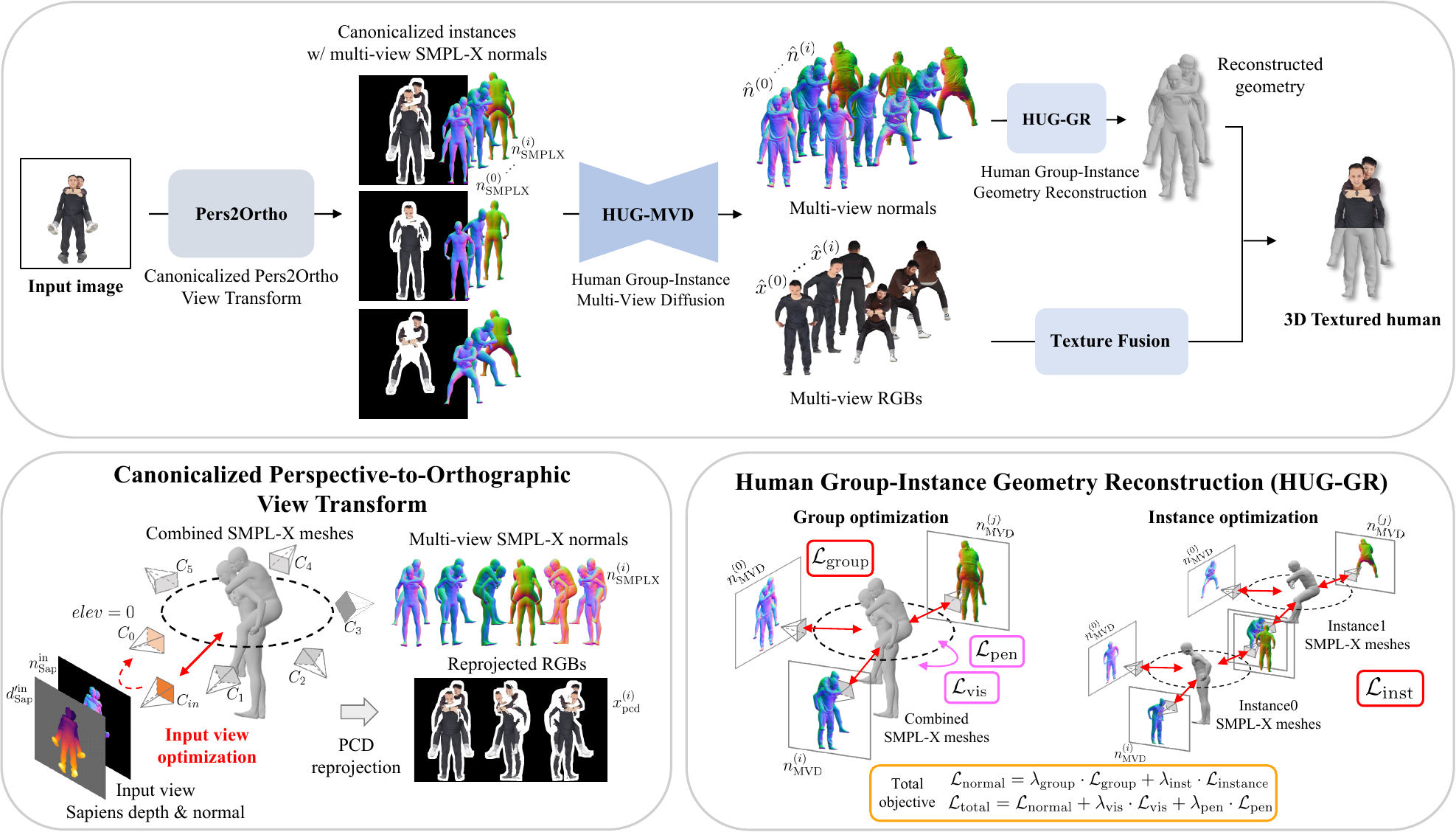}
    \vspace{-.5em} 
    \caption{Overview of our {HUG3D} framework. Given a single perspective image, 
    (1) the \textit{Canonical Perspective-to-Orthographic View Transform (Pers2Ortho)} 
    converts it into a canonical multi-view orthographic representation to resolve scale ambiguity and enable consistent multi-view reasoning. 
    (2) The \textit{Human Group-Instance Multi-View Diffusion (HUG-MVD)} model completes 
    occluded geometry and texture while maintaining plausible interactions. 
    (3) The \textit{Textured Mesh Reconstruction} stage refines the mesh and generates high-fidelity textures with our 
    physics-based \textit{Human Group-Instance Geometry Reconstruction (HUG-GR)}, which enforces physical consistency via optimization with multi-view normal cues and interaction constraints.
    }
    \vspace{-1.5em}
    \label{fig_method}
\end{figure*}

\begin{figure}[t]
    \centering
    \includegraphics[width=\columnwidth]{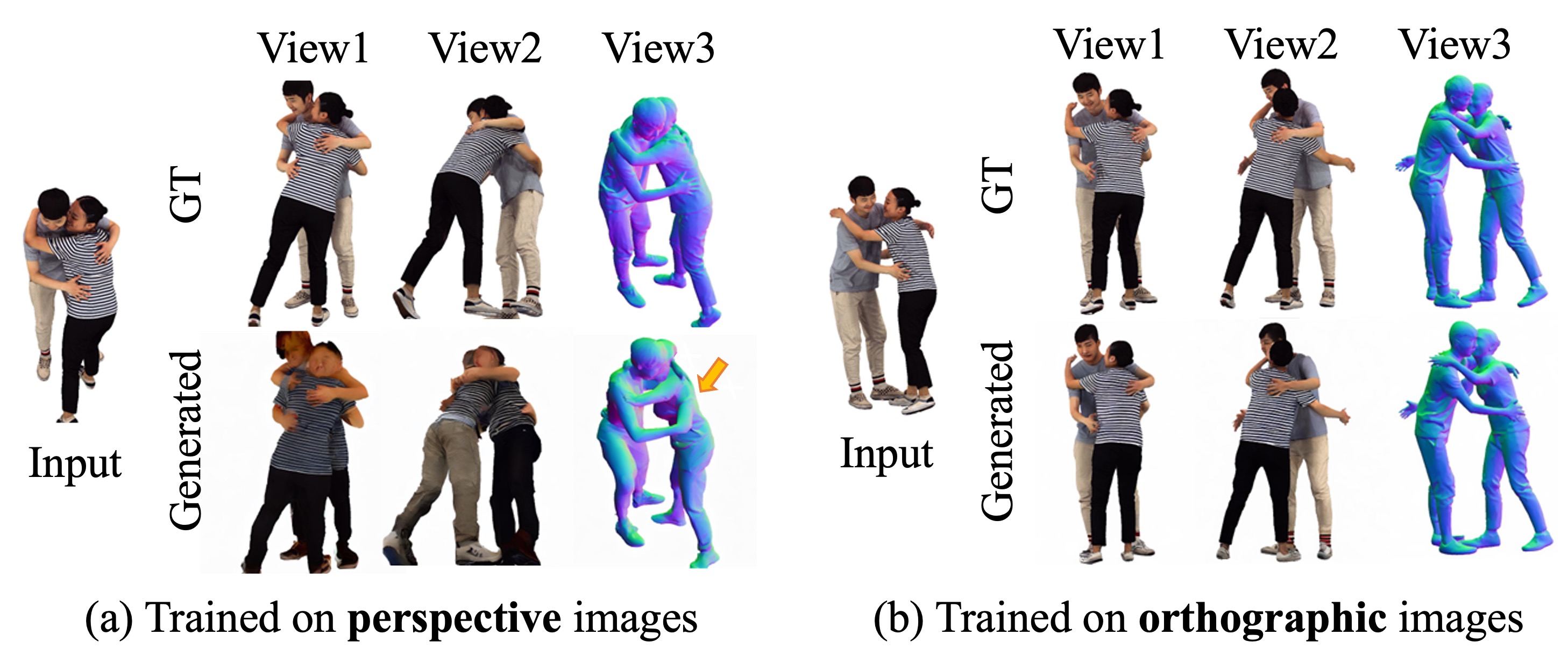}
    \vspace{-2em}
    \caption{Comparison of results from multi-view diffusion trained on perspective vs. orthographic images.}
    \label{fig_ablation_persortho}
    \vspace{-1.5em}
\end{figure}
\section{Human Interaction-Aware 3D Reconstruction from a Single Image}
\label{sec:method}

Our HUG3D framework operates in three main stages, as illustrated in Fig.~\ref{fig_method}. 
First, the Canonical Perspective-to-Orthographic View Transform (\textit{Pers2Ortho}) module converts the input perspective image into a consistent multi-view orthographic representation. Next, our Human Group-Instance Multi-View Diffusion (\textit{HUG-MVD}) model completes occluded geometry and texture while ensuring plausible interactions. 
Finally, the Textured Mesh Reconstruction stage refines the mesh with our physics-based \textit{HUG-GR} module and synthesizes a high-fidelity texture.

\subsection{Canonical Perspective-to-Orthographic View Transform}
Multi-human scenes exhibit severe depth variation and occlusion, which violate the orthographic assumptions typically adopted in multi-view diffusion models. As a result, directly learning interactions from a single perspective-view image is challenging due to extreme geometric complexity (see Fig.~\ref{fig_ablation_persortho}). To overcome this limitation, we introduce Pers2Ortho module, a canonical view transformation module that estimates partial 3D geometry from the input perspective view and constructs consistent multi-view representations robust to geometric distortion. 

\noindent \textbf{Initial Geometry, Segmentation, and Camera Setup.}
Given a single RGB input, we first estimate an initial SMPL-X mesh, instance segmentation masks, and the perspective camera parameters $(\mathbf{K}_{\text{in}}, \mathbf{R}_{\text{in}}, \mathbf{T}_{\text{in}})$ of $\mathcal{C}_{\text{in}}$. We employ RoBUDDI, our more robust variant of BUDDI~\citep{muller2024generative}, as detailed in supplementary Sec. A.1. 
To define a canonical processing space, the SMPL-X mesh is first normalized to a tight bounding box. Around this normalized geometry, we place six orthographic cameras $\{\mathcal{C}_0, \ldots, \mathcal{C}_5\}$ at fixed azimuths $\{0^\circ, 45^\circ, 90^\circ, 180^\circ, 270^\circ, 315^\circ\}$ with zero elevation. Each camera has extrinsic parameters $(\mathbf{R}_i, \mathbf{T}_i)$. This canonical camera rig ensures spatial alignment across all instances, providing a stable basis for downstream multi-view generation.

\noindent \textbf{Partial 3D Construction.}
To enhance the fidelity of the canonical representation, we initialize a partial 3D mesh $\mathcal{M}$ with the initial SMPL-X mesh and refine it using geometric supervision from Sapiens~\citep{khirodkar2024sapiens}. Specifically, Sapiens predicts affine-invariant depth ($d'^{\text{in}}_{\text{Sap}}$) and surface normals ($n^{\text{in}}_{\text{Sap}}$) from the perspective input view $\mathcal{C}_{\text{in}}$. The mesh vertices are optimized to align with these predictions by minimizing the geometry loss:
\begin{equation}
\mathcal{L}_{\text{geo}} = 
  \left\| d'^{\text{in}}_{\text{Sap}} - d'^{\text{in}}_{\mathcal{M}} \right\|_2^2 + 
  1 - \left\langle n^{\text{in}}_{\text{Sap}}, n^{\text{in}}_{\mathcal{M}} \right\rangle,
\label{eq:geo_loss_affine}
\end{equation}
where $d'^{\text{in}}_{\mathcal{M}}$ and $n^{\text{in}}_{\mathcal{M}}$ denote the depth and normal maps rendered from the mesh. The first term enforces depth consistency via L2 distance, while the second term enforces orientation consistency via cosine similarity. To handle possible topology mismatches between the initial mesh and Sapiens predictions, we adopt a remeshing strategy~\citep{li2024pshuman}.  

\noindent \textbf{Multi-View Input Generation via PCD Reprojection.}
For the multi-view diffusion stage, we first render a complete set of normal maps $\{n_{\text{SMPLX}}^{(i)}\}_{i=0}^{5}$ from all six canonical views $\{\mathcal{C}_i\}$ from the initial SMPL-X mesh for the pose guidance. 

Second, we generate partial RGB inputs $\{x^{(0)}_{\text{pcd}}, x^{(1)}_{\text{pcd}}, x^{(5)}_{\text{pcd}}\}$ by reprojecting the input image onto the refined partial 3D mesh $\mathcal{M}$ from three key canonical views ($0^\circ$, $45^\circ$, and $315^\circ$).  
Concretely, we render a depth map $d'^{\text{in}}_{\mathcal{M}}$ from $\mathcal{M}$ and build a dense point cloud (PCD) $\mathcal{P}$ in the coordinate system of $\mathcal{C}_{\text{in}}$. This PCD is reprojected into each orthographic view $\mathcal{C}_i$ as:
\begin{equation}
\mathcal{P}_{i} = \Pi_{\text{ortho}}^{i}(\mathcal{P}) = \mathbf{R}_{i} \cdot \mathcal{P} + \mathbf{T}_{i},
\label{eq:pcd_reproject}
\end{equation}
where $\Pi_{\text{ortho}}^{i}$ denotes the orthographic projection. RGB values from $\mathcal{C}_{\text{in}}$ are then transferred onto $\mathcal{P}_i$, yielding partial RGB maps $x^{(i)}_{\text{pcd}}$. Unlike mesh vertex coloring, which often produces sparse and low-quality textures, our PCD reprojection preserves dense appearance details in visible regions while maintaining spatial consistency across canonical views. These inputs serve as robust conditioning signals for multi-view diffusion. Further details can be found in Sec. A.2 
of the supplementary.

\begin{figure*}[!t]
     \centering
    \includegraphics[width=0.9\textwidth]{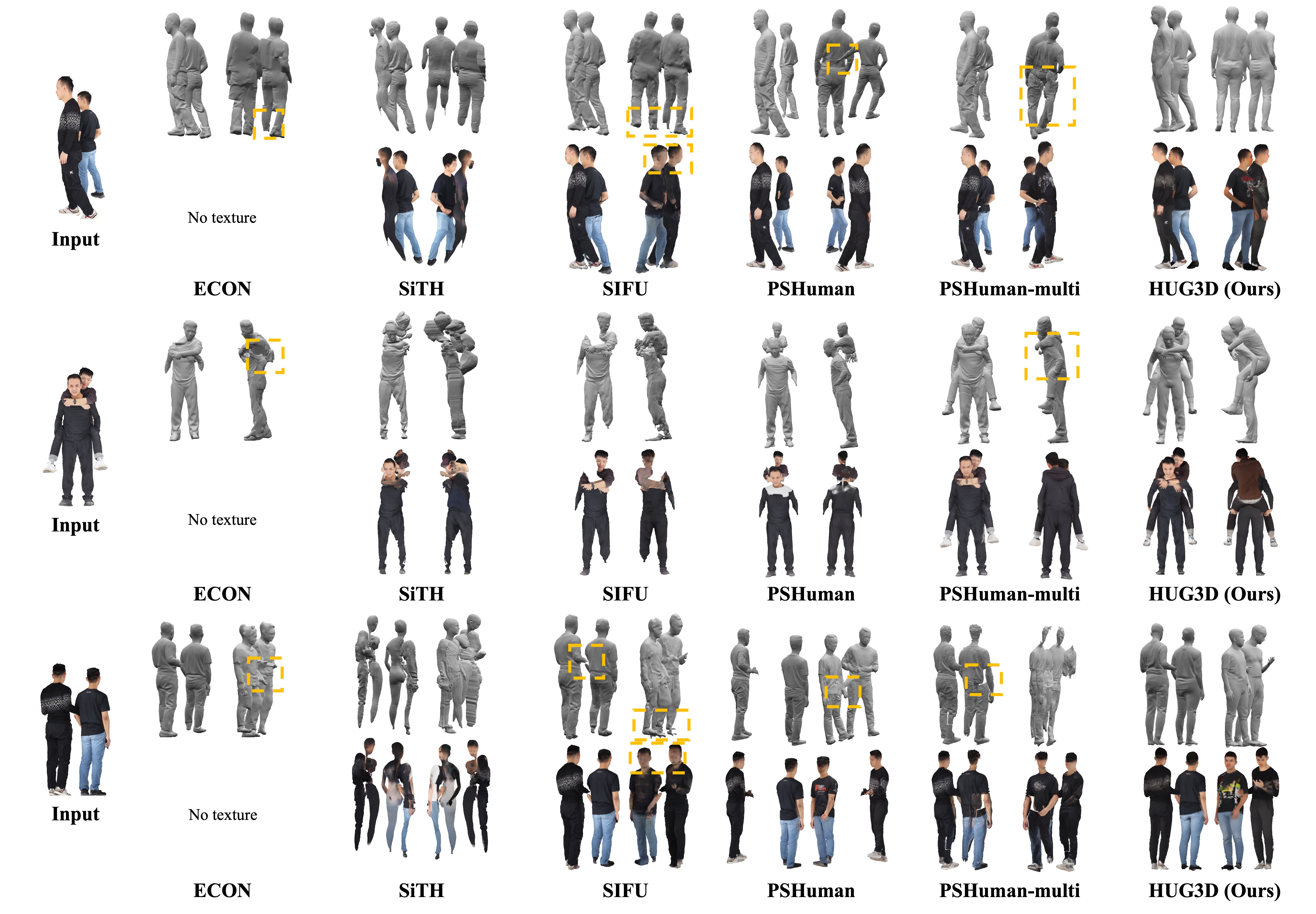}
    \vspace{-1.em}
    \caption{Qualitative comparison on multi-human 3D reconstruction from a single image. HUG3D outperforms baselines by correcting perspective distortion, preserving inter-human contact, and hallucinating plausible textures under heavy occlusion.}
    \vspace{-1.5em}
    \label{fig_quali_multihuman}
\end{figure*}

\begin{figure*}[]
     \centering
    \includegraphics[width=0.9\textwidth]{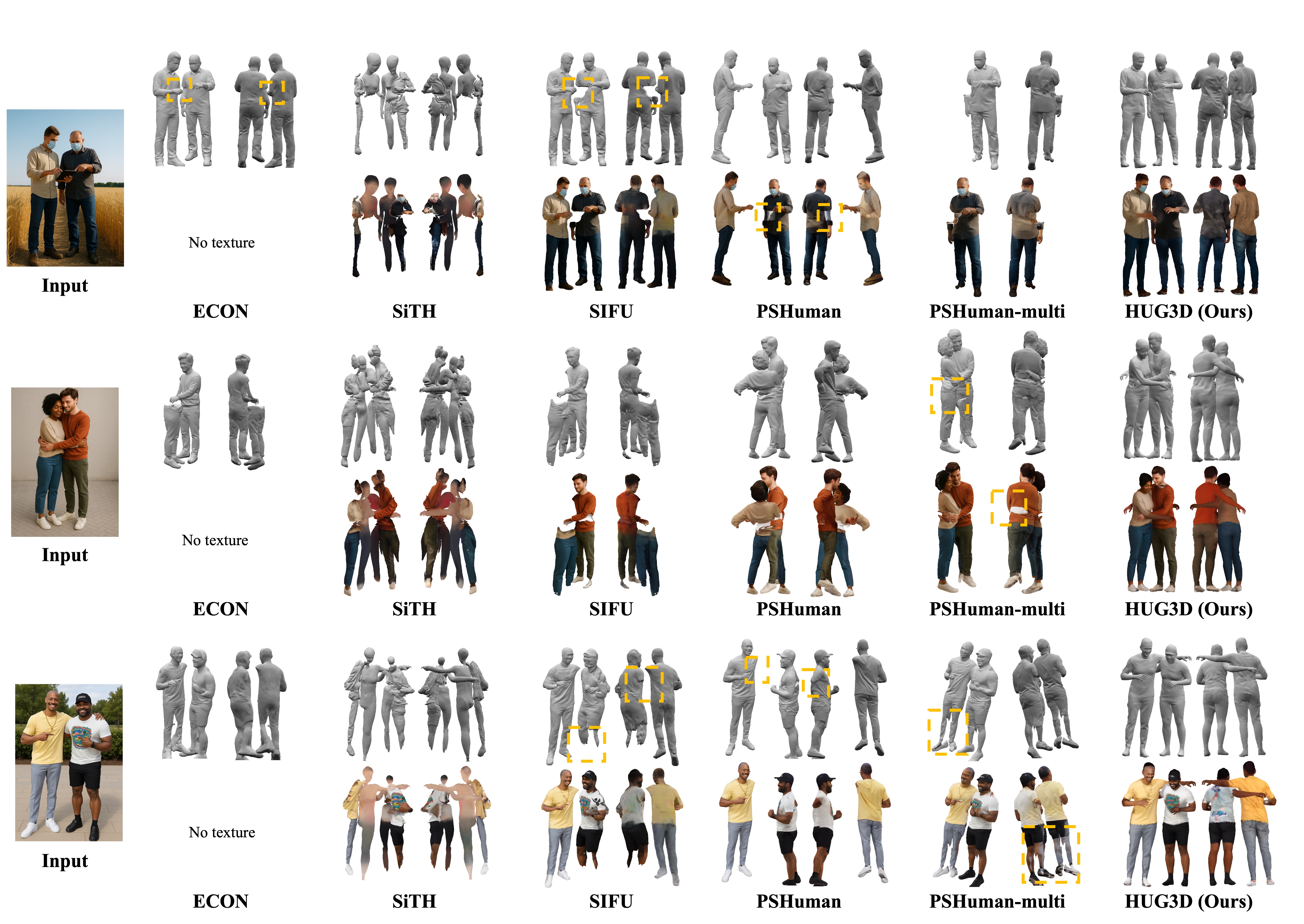}
    \vspace{-1.em}
    \caption{Qualitative results on in-the-wild images. Our method demonstrates robust multi-human reconstructions, outperforming baseline approaches and highlighting its practical applicability.}
    \vspace{-1.em}
    \label{fig_quali_inthewild}
\end{figure*}


\subsection{Human Group-Instance Multi-View Diffusion}
We introduce Human Group-Instance Multi-View Diffusion (HUG-MVD), a diffusion model that leverages both group-level and instance-level priors to resolve occlusions and inter-person interactions in multi-human scenes.

\noindent \textbf{Interaction- and Occlusion-Aware Multi-View Diffusion.}  
Given reprojected partial RGB inputs and canonical-view SMPL-X normal maps, HUG-MVD reconstructs geometry and appearance missing due to occlusion. Unlike standard single-human diffusion models, our formulation incorporates group-level priors by jointly training on two complementary sources. First, diverse single-human datasets with full supervision~\citep{ho2023learning,yu2021function4d} provide coverage over a wide range of body shapes and identities. Second, multi-human datasets with partial geometry~\citep{yin2023hi4d} capture realistic inter-person occlusions and interactions, though with limited identity variation. This combination enables the model to generalize across diverse individuals while remaining robust to complex group-level occlusions.

To simulate realistic occlusions for single-human training data, we generate masked inputs $x^{(i)}_{\text{mask}}$ from the reprojected point clouds $x^{(i)}_{\text{pcd}}$ and visibility masks $M_{\text{vis}}$, as described in Sec. A.3.3. 
These masks indicate unobserved regions for each canonical view $\mathcal{C}_i$. The model is conditioned on SMPL-X normal maps $\{n^{(i)}_{\text{SMPLX}}\}_{i=0}^{5}$, which provide geometric guidance via ControlNet~\citep{zhang2023adding}, steering the denoising process toward consistent, human-like reconstructions.

\noindent \textbf{Multimodal Training Objective.}  
Given 6 canonical views, the model predicts complete RGB images $\{\hat{x}^{(i)}\}$ and surface normals $\{\hat{n}^{(i)}\}$ simultaneously using a shared denoising diffusion process. To train the model, we minimize the following objective $\mathcal{L}_{\text{diff}}$ as below:
\begin{equation}
\begin{split}
\sum_{i=0}^{5} \Bigg( 
\mathbb{E}_{t,\epsilon} \Big[ 
\left\| \epsilon - \epsilon_\theta(z_{t, \text{rgb}}^{(i)}, t, x^{(i)}_{\text{mask}}, n^{(i)}_{\text{SMPLX}}) \right\|_2^2 
\Big] \\
+ \mathbb{E}_{t,\epsilon} \Big[ 
\left\| \epsilon - \epsilon_\theta(z_{t, \text{normal}}^{(i)}, t, x^{(i)}_{\text{mask}}, n^{(i)}_{\text{SMPLX}}) \right\|_2^2 
\Big]
\Bigg),
\label{eq:diffusion_loss}
\end{split}
\end{equation} 
where $z_{t,\text{rgb}}^{(i)}$ and $z_{t,\text{normal}}^{(i)}$ are the noisy latents for the RGB and normal map at timestep $t$, respectively. The denoising network $\epsilon_\theta$ learns to estimate the clean signals for both modalities, conditioned on the masked RGB inputs $x^{(i)}_{\text{mask}}$ and SMPL-X normal maps $n^{(i)}_{\text{SMPLX}}$. By jointly optimizing RGB and normal prediction, the model enforces consistency between appearance and geometry.

\noindent \textbf{Joint Group-Instance Inference.}  
At inference, a single HUG-MVD model jointly performs group- and instance-level reconstruction, predicting complete normals $\{\hat{n}^{(i)}\}$ and RGB images $\{\hat{x}^{(i)}\}$ from partial RGB inputs and SMPL-X guidance.  

To maintain consistency between group- and instance-level latent representations, instance-specific latents $z_{t,\text{inst}(k)}^{(i)}$ are injected into the group latent $z_{t,\text{group}}^{(i)}$ at their corresponding spatial regions. Formally, at diffusion timestep $t$:
\begin{eqnarray}
z_{t, \text{group}}^{(i)} &\leftarrow& \sum_{k=1}^{K} \left[
\alpha \cdot \mathbbm{1}_{\text{inst}}^{(i,k)} \cdot z_{t, \text{inst}(k)}^{(i)} +  (1 - \alpha)\cdot \mathbbm{1}_{\text{inst}}^{(i,k)} \right. \nonumber \\
&& \left.  \cdot z_{t, \text{group}}^{(i)}
\right]
+ \left(1 - \sum_{k=1}^{K} \mathbbm{1}_{\text{inst}}^{(i,k)}\right) \cdot z_{t, \text{group}}^{(i)}, 
\end{eqnarray}
where $\mathbbm{1}_{\text{inst}}^{(i,k)}$ indicates the spatial extent of instance $k$ in view $\mathcal{C}_i$, and $\alpha \in [0,1]$ controls blending strength. This composition allows fine-grained instance details to inform the global group representation, improving geometric consistency across overlapping surfaces.

Further details with detailed illustrations of HUG-MVD are provided in Sec. A.3 
of the supplementary.

\subsection{Textured Mesh Reconstruction}\label{hug_mr}

\subsubsection{Human Group-Instance Geometry Reconstruction}\label{hug_gr}
We refine the initial SMPL-X mesh to produce physically plausible geometry for multi-human scenes, leveraging both group- and instance-level supervision and physics-inspired constraints.

\noindent \textbf{Group-Instance Normal Supervision.}  
The mesh $\mathcal{M}$ is optimized to match predicted normals from HUG-MVD at both group and instance levels. Let \( n^{(i)}_{\text{MVD}, \text{group}} \) and \( n^{(i)}_{\mathcal{M}, \text{group}} \) denote predicted and rendered normals for the group, and \( n^{(i,k)}_{\text{MVD}, \text{inst}} \) and \( n^{(i,k)}_{\mathcal{M}, \text{inst}} \) for instance \( k \). We define:
\begin{equation}
\mathcal{L}_{\text{normal}} = \lambda_{\text{group}} \cdot \mathcal{L}_{\text{group}} + \lambda_{\text{inst}} \cdot \mathcal{L}_{\text{instance}},
\end{equation}
\begin{equation}
\begin{split}
\mathcal{L}_{\text{group}} = \sum_{i=0}^{5} \left(1 - \langle n^{(i)}_{\text{MVD}, \text{group}}, n^{(i)}_{\mathcal{M}, \text{group}} \rangle \right), \\ \quad
\mathcal{L}_{\text{instance}} = \sum_{i=0}^{5} \sum_{k=1}^{K} \left(1 - \langle n^{(i,k)}_{\text{MVD}, \text{inst}}, n^{(i,k)}_{\mathcal{M}, \text{inst}} \rangle \right),
\end{split}
\end{equation}
where $n_\mathcal{M}$ is rendered from the currently optimized mesh via a differentiable renderer at each step.

\noindent \textbf{Interpenetration Loss.}  
To prevent implausible overlaps between body parts, we define an interpenetration loss over tolerance pairs $(i,j)\in\mathrm{V}$, where $\mathrm{V}$ denotes the set of part pairs derived from contact regions in the initial SMPL-X meshes. 
For each pair, let $s_1^{i,j}$ and $s_2^{i,j}$ be the closest points on the surfaces of parts $i$ and $j$. 
The loss penalizes distances below a threshold $tol$, with $\xi$ acting as a smoothing temperature that controls the softness of the penalty:
\vspace{-0.5em}
\begin{equation}
\mathcal{L}_{\mathrm{pen}} = \operatorname{mean}_{V}\!\Big[\,\xi\,\ln\!\big(1 + e^{(\mathrm{tol}-\lvert s_1^{i,j}-s_2^{i,j}\rvert)/\xi}\big)\Big].
\end{equation}

\noindent \textbf{Visibility Loss.}  
We enforce consistency between rendered visibility and ground-truth masks:
\begin{equation}
\mathcal{L}_{\mathrm{vis}} =
  \frac{1}{2B} \sum_{k=1}^{K} \sum_{b=1}^{B}
    \frac{E^k_b}{M^k_b + \epsilon},
\end{equation}
where $E^k_b$ and $M^k_b$ denote incorrectly occluded and total visible pixels for body part $b$ in instance $k$. This loss encourages each body part in the rendered mesh to match its expected visible region, ensuring accurate silhouettes and occlusion boundaries, especially in complex multi-human interactions.

\noindent \textbf{Total Optimization Objective.}  
The final mesh is optimized using:
\begin{equation}
\mathcal{L}_{\text{total}} = \mathcal{L}_{\text{normal}} + \lambda_{\mathrm{vis}} \cdot \mathcal{L}_{\text{vis}} + \lambda_{\mathrm{pen}} \cdot \mathcal{L}_{\text{pen}}.
\end{equation}
We apply finer learning rates to high-frequency semantic regions (e.g., hands, face) to improve local accuracy while maintaining overall shape stability.

Additional details for HUG-GR are provided in Sec. A.4 
of the supplementary.

\subsubsection{Texture Construction}
Full-body vertex texture is generated by projecting multi-view RGBs onto the optimized mesh. To improve fidelity, occluded and low-confidence regions are blended using view-aware confidence masks. High-fidelity face restoration is applied for oblique or occluded views. 
Please refer to Sec. A.5 
of the supplementary for additional details.

\section{Experiments}
\label{sec:experiments}

\begin{table*}[!t]\centering
\caption{Quantitative comparison of geometric metrics for multi-human 3D reconstruction. HUG3D achieves the best overall scores in all metrics including CD, P2S, and NC, and also outperforms other baselines in CP, indicating better interaction-aware reconstruction.}
    \vspace{-0.5em}
\label{tab_geometry_main}
\scriptsize
\resizebox{0.7\textwidth}{!}{
\centering
\begin{tabular}{cccccccc}
\toprule
Method & CD ↓ & P2S ↓ & NC ↑ & F-score ↑ & bbox-IoU ↑ & Norm $L2$ ↓ & CP ↑  
\\
\midrule
SIFU & 5.644 & 2.284 & 0.754 & 29.244 & 0.778 & 0.028 & 0.089 \\
SiTH  & 9.251 & 3.185 & 0.709 & 21.037 & 0.708 & 0.040 & 0.135 \\
PSHuman & 15.579 & 6.088 & 0.617 & 9.749 & 0.659 & 0.069 & 0.027 \\
DeepMultiCap & 13.719 & 2.555 & 0.749 & 18.125 & 0.513 & 0.049 & 0.083 \\
\textbf{Ours} & \textbf{3.631} & \textbf{1.752} & \textbf{0.811} & \textbf{41.504} & \textbf{0.847} & \textbf{0.019} & \textbf{0.240} \\
\bottomrule
\end{tabular}
}
\vspace{-0.5em}
\end{table*}
\begin{table*}[!t]\centering
\scriptsize
    \begin{minipage}[t]{0.48\linewidth}
        \centering
        \caption{{Quantitative evaluation on texture quality of multi-human 3D reconstruction.}}
            \vspace{-0.5em}
        \label{tab_texture_main}
        \resizebox{0.8\textwidth}{!}{
        \begin{tabular}{cccc}
        
        \toprule
        Method & PSNR ↑ & SSIM ↑ & LPIPS ↓
        \\
        \midrule
        SIFU & 15.202 & 0.793 & 0.202 \\
        SiTH  & 13.798 & 0.789 & 0.233 \\
        PSHuman & 11.418 & 0.742 & 0.304\\
        \textbf{Ours} & \textbf{16.456} & \textbf{0.809} & \textbf{0.168}\\
        \bottomrule
        \end{tabular}
        }
    \end{minipage}
    \hfill
  \begin{minipage}[t]{0.48\linewidth}
    \centering
    \caption{Quantitive evaluation on geometry and texture quality within occluded region. }
        \vspace{-0.5em}
    \label{tab_occluded_main}
    \resizebox{0.8\textwidth}{!}{
    \begin{tabular}{ccccc}
      \toprule
      Method & Norm $L2$ ↓ & PSNR ↑ & SSIM ↑ \\
      \midrule
      SIFU              & 0.197 & 6.157 & 0.559 \\
      SiTH              & 0.197 & 6.355 & 0.539 \\
      PSHuman           & 0.252 & 4.621  & 0.510  \\
      DeepMultiCap      & 0.217 & - & - \\
      \textbf{Ours}         & \textbf{0.140} & \textbf{8.388} & \textbf{0.602}  \\
      \bottomrule
    \end{tabular}
    }
  \end{minipage}
  \vspace{-1.0em}
\end{table*}
\begin{table*}[!t]\centering
    \caption{Ablation study of key components in HUG3D. We report geometry (CD, P2S, Norm $L2$) and texture (PSNR, LPIPS) metrics under various configurations. }
    \vspace{-0.5em}
    \label{tab_ablation}
    \resizebox{0.85\textwidth}{!}{
    \begin{tabular}{cccccccc}
      \toprule
      Module & Method & CD ↓ & P2S ↓ & Occ.Norm $L2$ ↓ & PSNR ↑ & LPIPS ↓ & Occ.PSNR ↑\\
      \midrule
      \multirow{3}{*}{HUG-MVD}
        & Trained on group-only data      & 4.564 & 2.203 & 0.157 & 15.641 & 0.191 & 7.423 \\
        & Trained on instance-only data   & 4.645 & 2.245 & 0.156 & 15.840 & 0.185 & 7.726 \\
        & w/o Instance-to-group latent composition & 4.646 & 2.249 & 0.159 & 16.198 & 0.183 & 7.916 \\
      \cmidrule(lr){1-8}
    \multirow{2}{*}{HUG-GR}
      & Instance-only normal supervision  &  4.642 & 2.250 & 0.156 & 16.180 & 0.183 & 7.902 \\
      & Group-only normal supervision  &  4.620 & 2.230 & 0.159 & 16.169 & 0.183 & 7.678 \\
    \midrule
    \multicolumn{2}{c}{\textbf{Ours (Full)}} & \textbf{4.316} & \textbf{2.122} & \textbf{0.153} & \textbf{16.454} & \textbf{0.179} & \textbf{8.082} \\
      \bottomrule
    \end{tabular}
    }
\vspace{-1em}
  \end{table*}
 
\subsection{Experimental Setup}

\noindent \textbf{Implementation Details.}  
The \textit{Pers2Ortho} module is based on BUDDI~\citep{muller2024generative} for SMPL-X fitting and camera parameter estimation. For partial 3D construction, the initial SMPL-X mesh is optimized for 200 iterations using Adam~\citep{kingma2014adam} with a learning rate of $0.02$. 
\textit{HUG-MVD} is initialized from PSHuman~\citep{li2024pshuman}, based on Stable Diffusion 2.1~\cite{rombach2022high}, and integrates a frozen normal-map ControlNet~\citep{zhang2023adding}. Training is performed on a single NVIDIA A100 (80GB) GPU with a batch size of 16 and gradient accumulation of 8 steps, using Adam (lr=$5\times10^{-6}$, $\beta_1=0.9$, $\beta_2=0.999$). A two-stage curriculum is employed: (1) 1,000 steps without inpainting masks, and (2) 1,000 steps with inpainting masks to simulate occlusion. Training takes approximately two days. A DDPM scheduler with 1{,}000 diffusion steps is used during training, while inference uses a DDIM scheduler ($\eta=1.0$) with 40 denoising steps. A blending factor of $\alpha=0.8$ balances diversity and fidelity.
\textit{HUG-GR} optimizes the mesh over 200 iterations with a learning rate of 0.01 and loss weights: $\lambda_{\text{group}}=1.0$, $\lambda_{\text{inst}}=0.2$, $\lambda_{\text{pen}}=2.0$, and $\lambda_{\text{vis}}=1.0$.
Further implementation details are provided in Sec. A 
of the supplementary.

\noindent \textbf{Training Data.} Training samples of HUG-MVD are rendered from  raw scans which are composed of textured mesh sequences. We train our model using the Hi4D dataset~\citep{yin2023hi4d} for multi-human supervision, along with THuman2.0~\citep{yu2021function4d} and CustomHumans~\citep{ho2023learning} for diverse single-human poses and appearances, enabling robust learning across varied interactions and body configurations. Additional details are provided in Sec. A.3.1 
of the supplementary. 

\noindent \textbf{Evaluation Dataset.} 
We conduct experiments on the MultiHuman~\citep{zheng2021deepmulticap} dataset, which provides multi-person, multi-view sequences with full 3D mesh supervision. For quantitative evaluation, we focused on two-person interacting scenes, including 20 sequences - six of these feature closely interacting pairs with heavy occlusions and complex spatial entanglements, while the remaining sequences capture more natural interactions. For each scene, we select a random initial viewpoint and render four perspective views, to evaluate under natural camera distortion, at azimuth offsets of  \(\{0^\circ,\,90^\circ,\,180^\circ,\,270^\circ\}\), yielding 80 images in total. See Sec. B.2 
of the supplementary for the details.

\begin{figure*}[!t]
     \centering
    \includegraphics[width=0.9\textwidth]{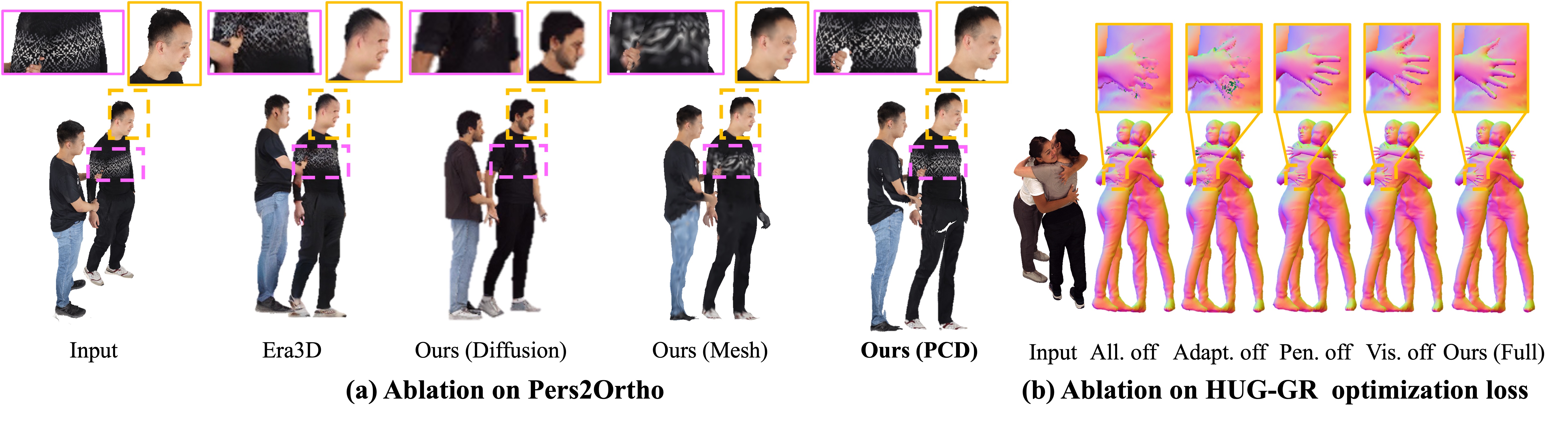}
    \vspace{-1.em}
\caption{Ablation study on key components. (a) Pers2Ortho improves projection sharpness.(b) Each HUG-GR loss boosts geometric plausibility.}

    \vspace{-2.em}
    \label{fig_ablation}
\end{figure*}

\noindent \textbf{Baselines.} 
To the best of our knowledge, we are the \textit{first} to tackle \textit{multi-human 3D reconstruction with both geometry and texture}, and no existing public baselines directly address this task (see Tab. S5 in supplementary). 
We therefore compare our method against two categories of prior works: (i) single-human reconstruction from a single image, and (ii) multi-human reconstruction from multi-view images or videos. 
To ensure a fair comparison, we follow the evaluation protocol of~\citep{cha20243d}.  
For single-human methods—ECON~\citep{xiu2023econ}, SIFU~\citep{zhang2024sifu}, SiTH~\citep{ho2024sith}, and PSHuman~\citep{li2024pshuman}—we crop each person using the dataset's ground-truth instance masks, reconstruct them independently, and then align the results into a shared coordinate frame and composite to form the complete scene. 
We also report PSHuman's performance when applied directly to uncropped multi-person images. 
For multi-human methods, we evaluate DeepMultiCap~\citep{zheng2021deepmulticap}, designed for multi-view images, and Multiply~\citep{jiang2024multiply}, designed for videos, under single image setting. 
All SMPL-based methods use ground-truth SMPL-X poses to isolate reconstruction quality from pose estimation errors. 
Further details are provided in supplementary Sec. B.1 
and results on multi-human baselines can be found in Sec. C.1 
of the supplementary material.

\noindent \textbf{Evaluation Metrics.}  We evaluate geometry with $L_1$ Chamfer distance (CD) [cm] and 1-directional point-to-surface distance (P2S) [cm], normal consistency (NC), F-score, and bbox-IoU between reconstructed and ground-truth meshes. To assess surface detail consistency, we compute $L_2$ normal error between predicted and ground-truth normal renders - across four rotated views at \(\{0^\circ,\,90^\circ,\,180^\circ,\,270^\circ\}\) relative to the input view. This is also separately computed specifically within occluded regions to evaluate robustness under occlusions. Physical realism is quantified through contact precision score (CP) defined by the
overlap between the estimated and ground-truth inter-body contact map. Texture fidelity is assessed using PSNR, SSIM, and LPIPS, computed across same views as $L_2$ normal error. Also separately computed specifically within occluded regions. 
More details are provided in Sec. B.3 
of the supplementary.

\subsection{Results}

\noindent \textbf{Quantitative Results.} As shown in Tab.~\ref{tab_geometry_main}, our method outperforms on all geometric metrics, with the lowest CD, P2S and highest NC. CP is markedly higher than other baselines, indicating superior physical realism and fidelity of inter-instance contacts. Tab.~\ref{tab_texture_main} shows our method achieved the highest PSNR and SSIM scores along with the lowest LPIPS. Moreover, in the occluded-region evaluation, Tab.~\ref{tab_occluded_main}, we significantly outperform the baselines in both Normal $L2$ error for geometry and PSNR/SSIM for texture. This indicates better perceptual quality of hallucinated textures where baselines typically fall.

\noindent \textbf{Qualitative Results.} 
Fig.~\ref{fig_quali_multihuman} illustrates that our approach surpasses all baselines in multi-human 3D reconstruction from a single image. Pers2Ortho enables accurate shape recovery in challenging viewpoints such as elevated scenes, while baseline models produce distorted results. Through HUG-MVD and HUG-GR, we model inter-human interactions, producing meshes that faithfully align with contact regions in the input image. In contrast, baselines either suffer from interpenetration or fail to preserve contact. Visual inspections confirm that HUG3D successfully reconstructs complete human meshes even under substantial occlusion. Hallucinated textures plausibly infer clothing and shape details not visible in the input. Competing methods, by contrast, exhibit broken surfaces, floating limbs, and missing textures.
Fig.~\ref{fig_quali_inthewild} shows results on in-the-wild images. Our approach consistently reconstructs multiple humans with high fidelity, surpassing baseline methods and demonstrating real-world applicability.

Additional results can be found in Secs. C and D 
of the supplementary.

\subsection{Ablation Study}

\noindent \textbf{Pers2Ortho.} 
As shown in Fig.~\ref{fig_ablation}(a), we evaluate our PCD-based Pers2Ortho against several baselines: the generative method Era3D~\citep{li2024era3d}, and our custom diffusion and mesh reprojection variants. These competing methods often produce blurry results with lost facial details, whereas our strategy consistently preserves high-fidelity features from the original view.

\noindent \textbf{HUG-MVD.} 
Tab.~\ref{tab_ablation} shows that training HUG-MVD on group-only data~\citep{yin2023hi4d} suffers from limited geometry and texture diversity, while instance-only data~\citep{ho2023learning, yu2021function4d} fails to capture inter-person interactions. Our full setting, combining both sources, significantly outperforms either alone, highlighting their complementary benefits.
Also, disabling instance-to-group latent composition leads to inconsistent surfaces and breaks continuity across instances.

\noindent \textbf{HUG-GR.} 
We analyze the impact of each component in HUG-GR. As shown in Fig.~\ref{fig_ablation}(b), omitting adaptive-region specific optimization degrades quality in high-frequency areas, such as hands. Removing interpenetration loss leads to mesh interpenetration and unrealistic overlaps, demonstrating its importance for physically grounded reconstruction. Excluding the visibility loss impairs alignment and produces unnatural surface at contact regions. In Tab.~\ref{tab_ablation}, by comparing variants using only group-level or only instance-level losses, we found that incorporating both losses simultaneously yields the best performance. 

More ablations and analysis can be found in Sec. E 
of the supplementary.

\section{Discussion}
\label{sec:discussion}

\noindent \textbf{Limitations.} 
HUG3D focuses on inter-human occlusion and does not handle occlusions from external objects, which we plan to address in future work. Although robust in the wild, our method can fail under severe depth ambiguity or heavy occlusion due to the inherent limitations of the single-image setting. Further discussion appears in Sec. G 
of the supplementary.

\noindent \textbf{Conclusion.} 
We presented {HUG3D}, a three-stage framework for high-fidelity 3D reconstruction of interacting people from a single RGB image. HUG3D addresses key challenges such as occlusions, complex interactions, and geometric distortions by combining a canonical view transform (Pers2Ortho), a group-instance multi-view diffusion model (HUG-MVD), and a physics-based reconstruction stage (HUG-GR) for refined geometry and texture. 
Extensive experiments show that HUG3D significantly outperforms single-human and prior multi-human methods, producing visually accurate, physically plausible reconstructions. Our framework enables reliable applications in AR/VR, telepresence, and digital human modeling, advancing realistic multi-human 3D reconstruction from a single image.

\section*{Acknowledgements}
This work was supported in part by Institute of Information \& communications Technology Planning \& Evaluation (IITP) grant funded by the Korea government(MSIT) [NO.RS-2021-II211343, Artificial Intelligence Graduate School Program (Seoul National University)], the National Research Foundation of Korea(NRF) grants funded by the Korea government(MSIT) (Nos. RS-2022-NR067592, RS-2025-02263628), the BK21 FOUR program of the Education and Research Program for Future ICT Pioneers, Seoul National University and AI-Bio Research Grant through Seoul National University.

{
    \small
    \bibliographystyle{cvpr2026_ieeenat_fullname}
    \bibliography{cvpr2026_main}

@String(CVPR= {IEEE Conf. Comput. Vis. Pattern Recog.})

@String(ICCV= {Int. Conf. Comput. Vis.})

@String(ECCV= {Eur. Conf. Comput. Vis.})

@String(TOG= {ACM Trans. Graph.})

@String(ICLR = {Int. Conf. Learn. Represent.})

@String(CVPR  = {CVPR})

@String(ICCV  = {ICCV})

@String(ECCV  = {ECCV})

@String(TOG   = {ACM TOG})

@String(ICLR  = {ICLR})

@inproceedings{lee2024guess,
  title={Guess the unseen: Dynamic 3d scene reconstruction from partial 2d glimpses},
  author={Lee, Inhee and Kim, Byungjun and Joo, Hanbyul},
  booktitle={Proceedings of the IEEE/CVF conference on computer vision and pattern recognition},
  pages={1062--1071},
  year={2024}
}

@article{ma2021pixel,
  title={Pixel codec avatars},
  author={Ma, Shugao and Simon, Tomas and Saragih, Jason and Wang, Dawei and Li, Yuecheng and De La Torre, Fernando and Sheikh, Yaser},
  journal={CVPR},
  year={2021}
}

@article{chupa,
    author    = {Kim, Byungjun and Kwon, Patrick and Lee, Kwangho and Lee, Myunggi and Han, Sookwan and Kim, Daesik and Joo, Hanbyul},
    title     = {Chupa: Carving 3D Clothed Humans from Skinned Shape Priors using 2D Diffusion Probabilistic Models},
    journal = {ICCV},
    year      = {2023}
}

@article{orts2016holoportation,
  title={Holoportation: Virtual 3d teleportation in real-time},
  author={Orts-Escolano, Sergio and Rhemann, Christoph and Fanello, Sean and Chang, Wayne and Kowdle, Adarsh and Degtyarev, Yury and Kim, David and Davidson, Philip L and Khamis, Sameh and Dou, Mingsong and others},
  journal={ACM Symposium on User Interface Software and Technology},
  pages={741--754},
  year={2016}
}

@article{cao2024mvinpainter,
  title={Mvinpainter: Learning multi-view consistent inpainting to bridge 2d and 3d editing},
  author={Cao, Chenjie and Yu, Chaohui and Wang, Fan and Xue, Xiangyang and Fu, Yanwei},
  journal={arXiv preprint arXiv:2408.08000},
  year={2024}
}

@misc{flux2024,
    author={Black Forest Labs},
    title={FLUX},
    year={2024},
    howpublished={\url{https://github.com/black-forest-labs/flux}},
}

@article{barda2024instant3dit,
  title={Instant3dit: Multiview Inpainting for Fast Editing of 3D Objects},
  author={Barda, Amir and Gadelha, Matheus and Kim, Vladimir G and Aigerman, Noam and Bermano, Amit H and Groueix, Thibault},
  journal={arXiv preprint arXiv:2412.00518},
  year={2024}
}

@article{xu2022vitpose,
  title={Vitpose: Simple vision transformer baselines for human pose estimation},
  author={Xu, Yufei and Zhang, Jing and Zhang, Qiming and Tao, Dacheng},
  journal={NeurIPS},
  year={2022}
}

@article{khanam2024yolov11,
  title={Yolov11: An overview of the key architectural enhancements},
  author={Khanam, Rahima and Hussain, Muhammad},
  journal={arXiv preprint arXiv:2410.17725},
  year={2024}
}

@article{paszke2019pytorch,
  title     = {PyTorch: An Imperative Style, High-Performance Deep Learning Library},
  author    = {Paszke, Adam and Gross, Sam and Massa, Francisco and Lerer, Adam and Bradbury, James and Chanan, Gregory and Killeen, Trevor and Lin, Zeming and Gimelshein, Natalia and Antiga, Luca and Desmaison, Alban and Kopf, Andreas and Yang, Edward and DeVito, Zachary and Raison, Martin and Tejani, Alykhan and Chilamkurthy, Sasank and Steiner, Benoit and Fang, Lu and Bai, Junjie and Chintala, Soumith},
  journal = {NeurIPS},
  year      = {2019},
  url       = {https://pytorch.org}
}

@article{ravi2020accelerating,
  title   = {Accelerating 3D Deep Learning with PyTorch3D},
  author  = {Ravi, Nikhila and Reizenstein, Jeremy and Novotny, David and Gordon, Taylor and Lo, Wan-Yen and Johnson, Justin and Gkioxari, Georgia},
  journal = {arXiv preprint arXiv:2007.08501},
  year    = {2020},
  url     = {https://pytorch3d.org}
}

@misc{vonplaten2022diffusers,
  title={Diffusers: State-of-the-art diffusion models},
  author={von Platen, Patrick and Lewis, Luke and Wolf, Thomas},
  year={2022},
  url={https://github.com/huggingface/diffusers}
}

@article{yin2023hi4d,
  title     = {Hi4D: 4D Instance Segmentation of Close Human Interaction},
  author    = {Yin, Yifei and Guo, Chen and Kaufmann, Manuel and Zarate, Juan Jose and Song, Jie and Hilliges, Otmar},
  journal = {CVPR},
  year      = {2023},
  url       = {https://yifeiyin04.github.io/Hi4D/},
  note      = {License: Non-commercial academic use only. See \url{https://hi4d.ait.ethz.ch/}}
}

@article{ho2023learning,
  title     = {Learning Locally Editable Virtual Humans},
  author    = {Ho, Hsuan-I and Xue, Lixin and Song, Jie and Hilliges, Otmar},
  journal = {CVPR},
  year      = {2023},
  url       = {https://custom-humans.github.io/}}

@article{zheng2021deepmulticap,
title={DeepMultiCap: Performance Capture of Multiple Characters Using Sparse Multiview Cameras},
author={Zheng, Yang and Shao, Ruizhi and Zhang, Yuxiang and Yu, Tao and Zheng, Zerong and Dai, Qionghai and Liu, Yebin},
journal={ICCV},
year={2021},
}

@article{zhang2023adding,
  title={Adding Conditional Control to Text-to-Image Diffusion Models}, 
  author={Lvmin Zhang and Anyi Rao and Maneesh Agrawala},
  journal={ICCV},
  year={2023},
}

@article{Deng2020RetinaFace,
  author={Deng, Jiankang and Guo, Jia and Ververas, Evangelos and Kotsia, Irene and Zafeiriou, Stefanos},
  journal={CVPR}, 
  title={RetinaFace: Single-Shot Multi-Level Face Localisation in the Wild}, 
  year={2020}
}

@article{pavlakos2019expressive,
  title={Expressive body capture: 3d hands, face, and body from a single image},
  author={Pavlakos, Georgios and Choutas, Vasileios and Ghorbani, Nima and Bolkart, Timo and Osman, Ahmed and Tzionas, Dimitrios and Black, Michael J},
  journal={CVPR},
  year={2019}
}

@article{huang2024closely,
  title={Closely interactive human reconstruction with proxemics and physics-guided adaption},
  author={Huang, Buzhen and Li, Chen and Xu, Chongyang and Pan, Liang and Wang, Yangang and Lee, Gim Hee},
  journal={CVPR},
  year={2024}
}

@article{chanrobust,
  title={Robust-PIFu: Robust Pixel-aligned Implicit Function for 3D Human Digitalization from a Single Image},
  author={Chan, Kennard and Liu, Fayao and Lin, Guosheng and Foo, Chuan-Sheng and Lin, Weisi},
  journal={ICLR},
  year={2024},
}

@article{bogo2016keep,
  title={Keep it SMPL: Automatic estimation of 3D human pose and shape from a single image},
  author={Bogo, Federica and Kanazawa, Angjoo and Lassner, Christoph and Gehler, Peter and Romero, Javier and Black, Michael J},
  journal={ECCV},
  year={2016}
}

@article{kanazawa2018end,
  title={End-to-end Recovery of Human Shape and Pose},
  author={Kanazawa, Angjoo and Black, Michael J and Jacobs, David W and Malik, Jitendra},
  journal={CVPR},
  year={2018}
}

@article{saito2019pifu,
  title={Pifu: Pixel-aligned implicit function for high-resolution clothed human digitization},
  author={Saito, Shunsuke and Huang, Zeng and Natsume, Ryota and Morishima, Shigeo and Kanazawa, Angjoo and Li, Hao},
  journal={ICCV},
  year={2019}
}

@article{saito2020pifuhd,
  title={Pifuhd: Multi-level pixel-aligned implicit function for high-resolution 3d human digitization},
  author={Saito, Shunsuke and Simon, Tomas and Saragih, Jason and Joo, Hanbyul},
  journal={CVPR},
  year={2020}
}

@article{mustafa2021multi,
  title={Multi-person implicit reconstruction from a single image},
  author={Mustafa, Armin and Caliskan, Akin and Agapito, Lourdes and Hilton, Adrian},
  journal={CVPR},
  year={2021}
}

@article{fieraru2021remips,
  title={Remips: Physically consistent 3d reconstruction of multiple interacting people under weak supervision},
  author={Fieraru, Mihai and Zanfir, Mihai and Szente, Teodor and Bazavan, Eduard and Olaru, Vlad and Sminchisescu, Cristian},
  journal={NeurIPS},
  year={2021}
}

@article{muller2024generative,
  title={Generative proxemics: A prior for 3d social interaction from images},
  author={M{\"u}ller, Lea and Ye, Vickie and Pavlakos, Georgios and Black, Michael and Kanazawa, Angjoo},
  journal={CVPR},
  year={2024}
}

@article{hassan2021populating,
  title={Populating 3D scenes by learning human-scene interaction},
  author={Hassan, Mohamed and Ghosh, Partha and Tesch, Joachim and Tzionas, Dimitrios and Black, Michael J},
  journal={CVPR},
  year={2021}
}

@article{li2019crowdpose,
  title={Crowdpose: Efficient crowded scenes pose estimation and a new benchmark},
  author={Li, Jiefeng and Wang, Can and Zhu, Hao and Mao, Yihuan and Fang, Hao-Shu and Lu, Cewu},
  journal={CVPR},
  year={2019}
}

@article{kocabas2020vibe,
  title={Vibe: Video inference for human body pose and shape estimation},
  author={Kocabas, Muhammed and Athanasiou, Nikos and Black, Michael J},
  journal={CVPR},
  year={2020}
}

@article{xiu2022icon,
  title={Icon: Implicit clothed humans obtained from normals},
  author={Xiu, Yuliang and Yang, Jinlong and Tzionas, Dimitrios and Black, Michael J},
  journal={CVPR},
  year={2022},
  organization={IEEE}
}

@article{loper2015smpl,
  title={SMPL: A skinned multi-person linear model},
  author={Loper, Matthew and Mahmood, Naureen and Romero, Javier and Pons-Moll, Gerard and Black, Michael J},
  journal={ACM Transactions on Graphics (TOG)},
  year={2015},
  publisher={ACM}
}

@article{joo2018total,
  title={Total capture: A 3d deformation model for tracking faces, hands, and bodies},
  author={Joo, Hanbyul and Simon, Tomas and Sheikh, Yaser},
  journal={CVPR},
  year={2018}
}

@article{rombach2022high,
  title={High-resolution image synthesis with latent diffusion models},
  author={Rombach, Robin and Blattmann, Andreas and Lorenz, Dominik and Esser, Patrick and Ommer, Bj{\"o}rn},
  journal={CVPR},
  year={2022}
}

@article{li2024pshuman,
  title={PSHuman: Photorealistic Single-view Human Reconstruction using Cross-Scale Diffusion},
  author={Li, Peng and Zheng, Wangguandong and Liu, Yuan and Yu, Tao and Li, Yangguang and Qi, Xingqun and Li, Mengfei and Chi, Xiaowei and Xia, Siyu and Xue, Wei and others},
  journal={arXiv preprint arXiv:2409.10141},
  year={2024}
}

@article{zhang2024sifu,
  title={Sifu: Side-view conditioned implicit function for real-world usable clothed human reconstruction},
  author={Zhang, Zechuan and Yang, Zongxin and Yang, Yi},
  journal={CVPR},
  year={2024}
}

@article{ho2024sith,
  title={Sith: Single-view textured human reconstruction with image-conditioned diffusion},
  author={Ho, I and Song, Jie and Hilliges, Otmar and others},
  journal={CVPR},
  year={2024}
}

@article{xiu2023econ,
  title={Econ: Explicit clothed humans optimized via normal integration},
  author={Xiu, Yuliang and Yang, Jinlong and Cao, Xu and Tzionas, Dimitrios and Black, Michael J},
  journal={CVPR},
  year={2023}
}

@article{qiu2025lhm,
  title={LHM: Large Animatable Human Reconstruction Model from a Single Image in Seconds},
  author={Qiu, Lingteng and Gu, Xiaodong and Li, Peihao and Zuo, Qi and Shen, Weichao and Zhang, Junfei and Qiu, Kejie and Yuan, Weihao and Chen, Guanying and Dong, Zilong and others},
  journal={arXiv preprint arXiv:2503.10625},
  year={2025}
}

@article{jiang2024multiply,
  title={Multiply: Reconstruction of multiple people from monocular video in the wild},
  author={Jiang, Zeren and Guo, Chen and Kaufmann, Manuel and Jiang, Tianjian and Valentin, Julien and Hilliges, Otmar and Song, Jie},
  journal={CVPR},
  year={2024}
}

@article{cha20243d,
  title={3D Reconstruction of Interacting Multi-Person in Clothing from a Single Image},
  author={Cha, Junuk and Lee, Hansol and Kim, Jaewon and Truong, Nhat Nguyen Bao and Yoon, Jaeshin and Baek, Seungryul},
  journal={WACV},
  year={2024}
}

@article{li2024era3d,
  title={Era3d: high-resolution multiview diffusion using efficient row-wise attention},
  author={Li, Peng and Liu, Yuan and Long, Xiaoxiao and Zhang, Feihu and Lin, Cheng and Li, Mengfei and Qi, Xingqun and Zhang, Shanghang and Xue, Wei and Luo, Wenhan and others},
  journal={NeurIPS},
  year={2024}
}

@article{wang2025meat,
  title={MEAT: Multiview Diffusion Model for Human Generation on Megapixels with Mesh Attention},
  author={Wang, Yuhan and Hong, Fangzhou and Yang, Shuai and Jiang, Liming and Wu, Wayne and Loy, Chen Change},
  journal={CVPR},
  year={2025}
}

@article{baradel2024multi,
  title={Multi-hmr: Multi-person whole-body human mesh recovery in a single shot},
  author={Baradel, Fabien and Armando, Matthieu and Galaaoui, Salma and Br{\'e}gier, Romain and Weinzaepfel, Philippe and Rogez, Gr{\'e}gory and Lucas, Thomas},
  journal={ECCV},
  year={2024}
}

@article{sun2022putting,
  title={Putting people in their place: Monocular regression of 3d people in depth},
  author={Sun, Yu and Liu, Wu and Bao, Qian and Fu, Yili and Mei, Tao and Black, Michael J},
  journal={CVPR},
  year={2022}
}

@article{yu2021function4d,
  title={Function4d: Real-time human volumetric capture from very sparse consumer rgbd sensors},
  author={Yu, Tao and Zheng, Zerong and Guo, Kaiwen and Liu, Pengpeng and Dai, Qionghai and Liu, Yebin},
  journal={CVPR},
  year={2021}
}

@article{zhou2022towards,
  title={Towards robust blind face restoration with codebook lookup transformer},
  author={Zhou, Shangchen and Chan, Kelvin and Li, Chongyi and Loy, Chen Change},
  journal={NeurIPS},
  year={2022}
}

@article{kirillov2023segment,
  title={Segment anything},
  author={Kirillov, Alexander and Mintun, Eric and Ravi, Nikhila and Mao, Hanzi and Rolland, Chloe and Gustafson, Laura and Xiao, Tete and Whitehead, Spencer and Berg, Alexander C and Lo, Wan-Yen and others},
  journal={ICCV},
  year={2023}
}

@article{khirodkar2024sapiens,
  title={Sapiens: Foundation for human vision models},
  author={Khirodkar, Rawal and Bagautdinov, Timur and Martinez, Julieta and Zhaoen, Su and James, Austin and Selednik, Peter and Anderson, Stuart and Saito, Shunsuke},
  journal={ECCV},
  year={2024}
}

@string{cvpr = {CVPR}}

@string{iccv = {ICCV}}

@string{eccv = {ECCV}}

@string{tog = {ACM TOG}}

@string{iclr = {ICLR}}

@article{ho2021classifier,
 author = {Ho, Jonathan and Salimans, Tim},
 journal = {NeurIPS},
 title = {Classifier-free diffusion guidance},
 year = {2021}
}

@String(WACV = {WACV})

@String(ICCV = {ICCV})

@String(CVPR = {CVPR})

@String(ICLR = {ICLR})

@article{kingma2014adam,
  title={Adam: A method for stochastic optimization},
  author={Kingma, Diederik P and Ba, Jimmy},
  journal={arXiv:1412.6980},
  year={2014}
}

@article{wilcoxon1945individual,
  title={Individual comparisons by ranking methods},
  author={Wilcoxon, Frank},
  journal={Biometrics bulletin},
  volume={1},
  number={6},
  pages={80--83},
  year={1945},
  publisher={JSTOR}
}
}

\onecolumn

\begin{center}
    {\Large \bfseries Human Interaction-Aware 3D Reconstruction from a Single Image\par}
    \vspace{0.5em}
    {\Large Supplementary Material\par}
\end{center}

\vspace{3em}





%
  

\vspace{-2em}
\begin{center}
 
\end{center}%

\section*{List of Contents} \label{list_of_contents}
\addcontentsline{toc}{section}{List of Contents}

\noindent \textbf{\ref{sec_sup:methods}\quad\nameref{sec_sup:methods}} \\
\quad \ref{sec_sup:segmentation}\quad\nameref{sec_sup:segmentation} \\
\quad \ref{sec_sup:pers2ortho}\quad\nameref{sec_sup:pers2ortho} \\
\quad \ref{sec_sup:hugmvd}\quad\nameref{sec_sup:hugmvd} \\
\quad \ref{sec_sup:huggr}\quad\nameref{sec_sup:huggr} \\
\quad \ref{sec_sup:texture}\quad\nameref{sec_sup:texture} \\

\noindent \textbf{\ref{sec_sup:evaluation}\quad\nameref{sec_sup:evaluation}} \\
\quad \ref{sec_sup:evalsetting}\quad\nameref{sec_sup:evalsetting} \\
\quad \ref{sec_sup:evaldataset}\quad\nameref{sec_sup:evaldataset} \\
\quad \ref{sec_sup:metrics}\quad\nameref{sec_sup:metrics} \\

\noindent \textbf{\ref{sec_sup:comparison}\quad\nameref{sec_sup:comparison}} \\
\quad \ref{sec_sup:baseline}\quad\nameref{sec_sup:baseline} \\
\quad \ref{sec_sup:predsmpl}\quad\nameref{sec_sup:predsmpl} \\
\quad \ref{sec_sup:separate}\quad\nameref{sec_sup:separate} \\
\quad \ref{sec_sup:interaction}\quad\nameref{sec_sup:interaction} \\
\quad \ref{sec_sup:scalability}\quad\nameref{sec_sup:scalability} \\
\quad \ref{sec_sup:generalization}\quad\nameref{sec_sup:generalization} \\
\quad \ref{sec_sup:novel_view}\quad\nameref{sec_sup:novel_view} \\
\quad \ref{sec_sup:robustness}\quad\nameref{sec_sup:robustness} \\
\quad \ref{sec_sup:videos}\quad\nameref{sec_sup:videos} \\

\vspace{-0.25em}
\noindent \textbf{\ref{sec_sup:components}\quad\nameref{sec_sup:components}} \\

\vspace{-0.25em}
\noindent \textbf{\ref{sec_sup:ablation}\quad\nameref{sec_sup:ablation}} \\
\quad \ref{sec_sup:abl_segmentation}\quad\nameref{sec_sup:abl_segmentation} \\
\quad \ref{sec_sup:abl_pers2ortho}\quad\nameref{sec_sup:abl_pers2ortho} \\
\quad \ref{sec_sup:abl_hugmvd}\quad\nameref{sec_sup:abl_hugmvd} \\
\quad \ref{sec_sup:abl_texture}\quad\nameref{sec_sup:abl_texture} \\
\quad \ref{sec_sup:abl_interaction}\quad\nameref{sec_sup:abl_interaction} \\
\quad \ref{sec_sup:abl_efficiency}\quad\nameref{sec_sup:abl_efficiency} \\
\quad \ref{sec_sup:abl_stat_sign}\quad\nameref{sec_sup:abl_stat_sign} \\

\noindent \textbf{\ref{sec_sup:licenses}\quad\nameref{sec_sup:licenses}} \\
\quad \ref{sec_sup:libs}\quad\nameref{sec_sup:libs} \\
\quad \ref{sec_sup:data}\quad\nameref{sec_sup:data} \\
\quad \ref{sec_sup:pretrained}\quad\nameref{sec_sup:pretrained} \\

\noindent \textbf{\ref{sec_sup:limitations}\quad\nameref{sec_sup:limitations}} \\
\quad \ref{sec_sup:limitation}\quad\nameref{sec_sup:limitation} \\
\quad \ref{sec_sup:impact}\quad\nameref{sec_sup:impact} \\
\normalsize

\clearpage
\twocolumn 

\newpage
\appendix

\renewcommand{\theequation}{S\arabic{equation}}
\renewcommand{\thefigure}{S\arabic{figure}}
\renewcommand{\thetable}{S\arabic{table}}

\section{Details on Methods and Implementation} \label{sec_sup:methods}
\subsection{Robust Instance Segmentation and SMPL-X Estimation }  \label{sec_sup:segmentation}

\noindent \textbf{Instance Segmentation.}  
We adopt a hybrid instance segmentation pipeline that combines detection, pose estimation, and promptable segmentation to produce per-person masks from a single image. This process is designed to be occlusion-aware and ensures that each human instance is segmented consistently, serving as the foundation for downstream SMPL-X fitting.

We begin by applying YOLOv11~\citep{khanam2024yolov11} to detect human instances in the input image, using only the “person” class to extract tight bounding boxes around each individual. For each detected bounding box, we then estimate 2D keypoints using ViTPose~\citep{xu2022vitpose} generated from BUDDI~\citep{muller2024generative} project. These keypoints provide reliable localization of body joints and are retained for subsequent matching and alignment purposes. 
To generate high-quality binary masks for each person, we pass the bounding boxes as prompts to the Segment Anything Model (SAM)~\citep{kirillov2023segment}, which produces accurate per-instance segmentations. These masks are then associated with individual people by solving a Hungarian assignment problem between keypoint-based anchor regions (e.g., head and feet) and the detected masks, ensuring proper instance-level alignment. 
To further improve segmentation quality, overlapping or duplicate detections are merged based on intersection-over-union (IoU) thresholds. Additionally, in cases of occlusion or ambiguous limb segmentation, we use the predicted keypoints along with SAM’s region prompting to correct mismatched or missing parts, particularly in the hand and foot regions.
This pipeline enables robust and scalable segmentation of multiple humans in a single view, even in the presence of occlusion or complex poses.

\noindent \textbf{SMPL-X Estimation (RoBUDDI).}  
We adopt BUDDI~\citep{muller2024generative}, a diffusion-based prior model, to estimate SMPL-X parameters for multi-human scenes. While BUDDI produces high-quality predictions for individual subjects, it exhibits limited effectiveness in handling collisions and interpenetrations. This is primarily because the penetration constraints are applied only in a second-stage refinement, after collisions have already occurred. As a result, scenes with dense interactions still suffer from body interpenetrations and inaccurate keypoint estimations in occluded regions. 

To overcome these limitations, we introduce two physics-inspired supervision terms during optimization: (1) an interpenetration loss that penalizes body collisions between interacting subjects, and (2) a visibility-aware keypoint loss that reduces errors in self- and inter-human occluded areas. These additions enable more robust and physically plausible multi-human pose estimation.
We refer to our enhanced approach as RoBUDDI, which integrates these geometry-level constraints into the fitting process. As shown in Tab.~\ref{tab:smplx} and Fig.~\ref{fig_supp_ablation_smplx}, RoBUDDI achieves both quantitatively superior accuracy and qualitatively improved physical realism compared to the baseline.

The interpenetration loss penalizes unrealistic mesh overlaps between specific body part pairs. We define a set $\mathrm{V}$ of tolerance pairs derived from contact regions in the initial SMPL-X meshes (e.g., left thigh vs.\ right calf). For each pair \((i,j)\in\mathrm{V}\), the nearest surface points are denoted by $s_1^{i,j}$ and $s_2^{i,j}$, and we use their separation $\lvert s_1^{i,j}-s_2^{i,j}\rvert$ in the loss. We apply a soft constraint around a threshold $\mathrm{tol}$, with $T=\max(0.25\,\mathrm{tol},10^{-5})$ acting as a smoothing temperature that controls the softness of the penalty:
\begin{equation}
\small
\mathcal{L}_{\mathrm{pen}}
= \gamma_{\mathrm{pen}}\cdot \operatorname{mean}_{\mathrm{V}}\!\Big[
T\,\ln\!\big(1 + e^{(\mathrm{tol}-\lvert s_1^{i,j}-s_2^{i,j}\rvert)/T}\big)
\Big].
\end{equation}
\normalsize
where \(\gamma_{\mathrm{pen}}\) sets the overall strength of the term, and $T$ preserves a smooth transition around while enforcing a strong penalty when $\lvert s_1^{i,j}-s_2^{i,j}\rvert<\mathrm{tol}$.

In addition, we introduce a visibility‐aware keypoint loss that adaptively downweights occluded joints. Given instance segmentation masks, each projected joint \(j\) is assigned:
\[
w_j =
\begin{cases}
1, & \text{if joint \(j\) is visible},\\
\alpha_{\text{occ}}, & \text{if joint \(j\) is occluded (\(\alpha_{\text{occ}}=0.1\))}.
\end{cases}
\]
Let \(\{\mathbf{u}_j\}\) and \(\{\hat{\mathbf{u}}_j\}\) be the ground‐truth and estimated 2D joint positions. We define
\[
\small
\mathcal{L}_{\mathrm{std}}
= \frac{1}{N}\sum_{j=1}^N \|\mathbf{u}_j - \hat{\mathbf{u}}_j\|^2,
\qquad
\mathcal{L}_{\mathrm{vis}}
= \frac{1}{N}\sum_{j=1}^N w_j\,\|\mathbf{u}_j - \hat{\mathbf{u}}_j\|^2.
\]
\normalsize
Here, \(N\) denotes the total number of 2D keypoints used in the reprojection loss.
We then combine them as below.
\[
\small
\mathcal{L}_{\mathrm{kp}}
= \gamma_{\mathrm{std}}\,\mathcal{L}_{\mathrm{std}}
\;+\;
\gamma_{\mathrm{vis}}\,\mathcal{L}_{\mathrm{vis}}
\]
\normalsize
Although conceptually similar to the visibility loss used in mesh reconstruction, where misclassified silhouette pixels are penalized, this formulation operates on joint reprojection errors rather than mask‐pixel discrepancies, yielding improved robustness to occlusion in keypoint alignment.

RoBUDDI estimates each person's 3D pose in multi-person scenarios, relative to the camera, providing all necessary geometric information for canonicalization. This effectively sets the camera's extrinsic rotation to $I$ and translation to $\vec{0}$.

The original BUDDI optimization takes approximately 60s per image and peaks at 12.48GB of GPU memory on an NVIDIA RTX A6000 GPU (batch size=1). 
Adding our interpenetration and visibility penalties increases runtime to 77s and peak memory to 15.16GB. All experiments use an interpenetration threshold \(tol=0.02\), a penetration loss weight \(\gamma_{\mathrm{pen}}=15\), an occlusion weight \(\alpha_{\text{occ}}=0.1\), and visibility‐aware keypoints and keypoint loss blend factors \(\gamma_{\mathrm{std}}=\gamma_{\mathrm{vis}}=0.5\).

\begin{figure*}[t]
    \centering
    \includegraphics[width=.9\linewidth]{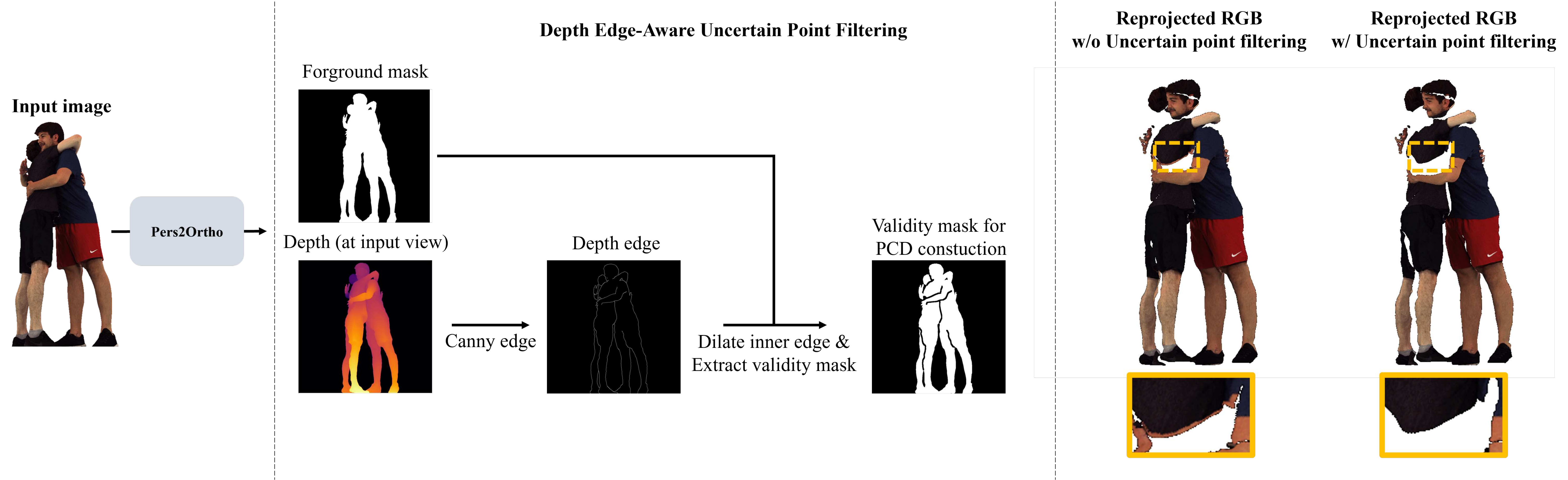}
    \vspace{-0.5em}
    \caption{Depth edge-aware filtering removes uncertain boundary regions to improve orthographic projection stability.}
    \label{fig_supp_method_p2o_uncertain_point_filtering}
    \vspace{-1em}
\end{figure*}

\begin{figure*}[t]
    \centering
    \includegraphics[width=0.7\linewidth]{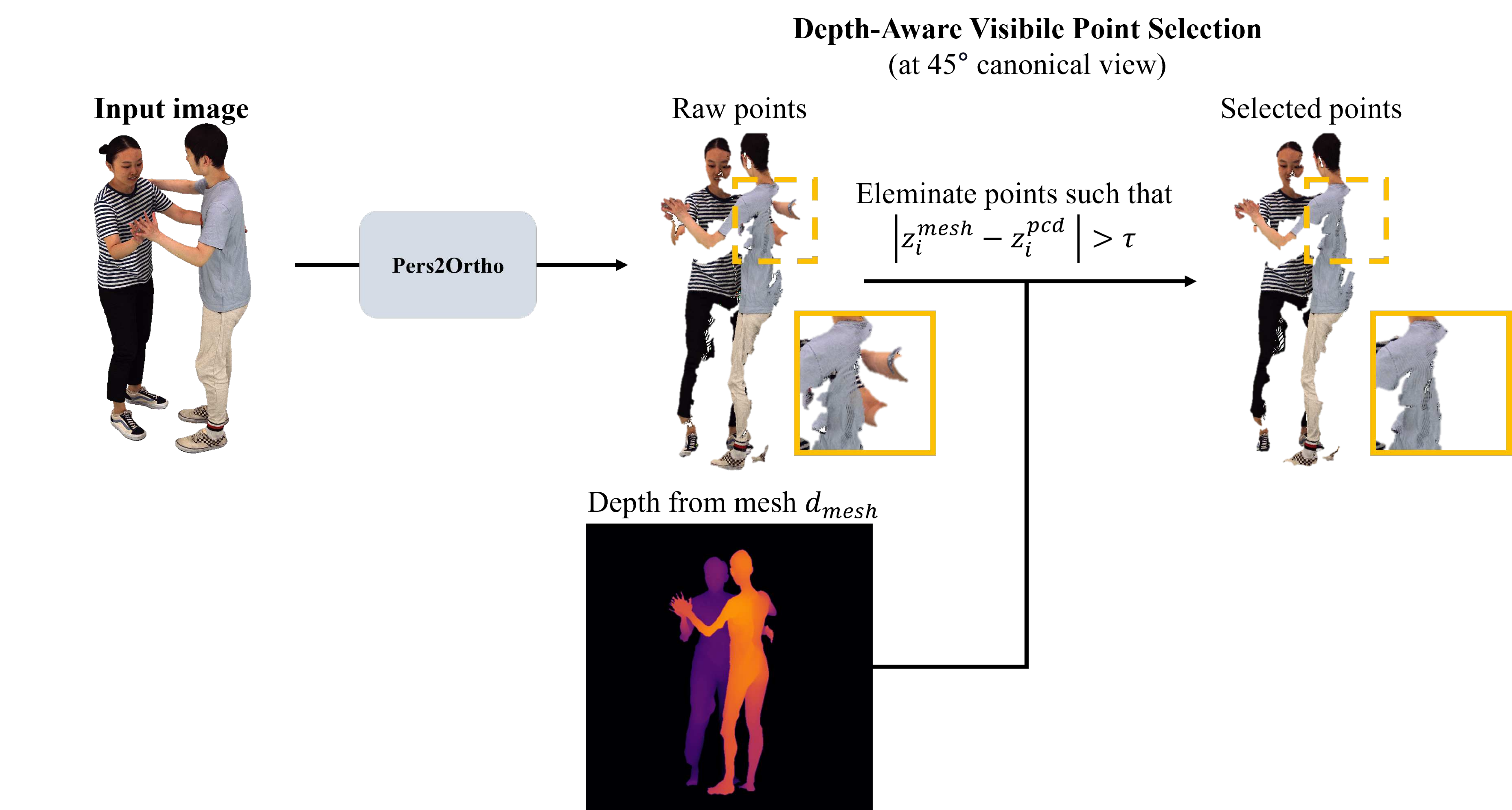}
    \vspace{-0.5em}
    \caption{Depth-aware filtering selects front-visible points for clean orthographic projections.}
    \label{fig_supp_method_p2o_visiblity_point_selection}
    \vspace{-1em}
\end{figure*}

\subsection{Canonical Perspective-to-Orthographic View Transform (Pers2Ortho)} \label{sec_sup:pers2ortho}

In addition to the primary transformation pipeline described in the main paper, we detail two depth-aware filtering strategies designed to suppress projection artifacts during the conversion from perspective to orthographic views.

\noindent \textbf{Depth Edge-Aware Uncertain Point Filtering.}  
As shown in Fig.~\ref{fig_supp_method_p2o_uncertain_point_filtering}, depth discontinuities often lead to jagged contours and ghosting artifacts near object boundaries. To address this, we detect depth edges using Canny edge detection applied to the rendered depth map. To reduce spurious detections near image borders, we erode the foreground mask before edge extraction. The resulting edge map is then dilated to encompass uncertain boundary regions. A refined validity mask is constructed by excluding these edge-dilated areas from the projection domain, ensuring that only stable, interior pixels are used in the orthographic projection.

\noindent \textbf{Depth-Aware Visible Point Selection.}  
As represented in Fig.~\ref{fig_supp_method_p2o_visiblity_point_selection}, to preserve geometric consistency during the projection of partial point clouds (PCD) into orthographic views, we filter for front-visible points using the rendered depth from the mesh as a geometric prior. This strategy eliminates occluded or background points, retaining only those lying in front of the mesh surface and visible from the target camera view.

Specifically, we project 3D world-space points onto the image plane and sample the mesh depth at corresponding pixel locations. A point is retained if:  
(1) its projected 2D coordinate lies within image bounds,  
(2) the absolute depth difference between the point and the mesh is below a threshold $\tau$ (set to $\tau = 0.02$), and  
(3) the sampled mesh depth is positive.

Formally, let $p_i \in \mathbb{R}^3$ be a 3D point from the PCD, and $z_i^\text{pcd}$ and $z_i^\text{mesh}$ denote the depths from the point cloud and mesh at the projected location, respectively. The point is retained if:
\begin{equation}
|z_i^\text{mesh} - z_i^\text{pcd}| < \tau, \quad \text{and} \quad z_i^\text{mesh} > 0
\end{equation}

This depth-aware visibility filtering yields cleaner foreground silhouettes and reduces projection noise by removing points that are geometrically inconsistent or lie behind the mesh surface.

We optimize the mesh for partial 3D reconstruction over 200 iterations with a learning rate of 0.02. 
The Pers2Ortho module, including the reprojection step, takes 16.20 seconds and consumes 14.4GB of VRAM on an NVIDIA A100. 
We apply a dilation operation with a kernel size of 5 for depth edges.

\subsection{Human Group-Instance Multi-View Diffusion (HUG-MVD)} \label{sec_sup:hugmvd}

\subsubsection{Training Datasets} \label{sec_sup:datasets}
We leverage one multi-human dataset~\citep{yin2023hi4d} and two single-human datasets~\citep{ho2023learning, yu2021function4d} to supervise our model with diverse human poses, interactions, and appearances.

\noindent \textbf{Hi4D}~\citep{yin2023hi4d} is a novel dataset targeting close-range, prolonged human-human interactions with physical contact. Capturing and disentangling such interactions is particularly challenging due to severe occlusions and topological ambiguities. To address this, Hi4D employs individually fitted neural implicit avatars and an alternating optimization scheme that jointly refines surface and pose during close contact. This enables automatic segmentation of fused 4D scans into individual humans. The dataset comprises 100 sequences across 20 subject pairs, totaling over 11K textured 4D scans, all annotated with accurate 2D/3D contact labels and registered SMPL-X models. For our experiments, we extract 1,272 scenes by sampling contact frames with a stride of 16.

\noindent \textbf{CustomHumans}~\citep{ho2023learning} contains high-quality static scans of 80 individuals, captured using a multi-view photogrammetry system with 53 RGB (12MP) and 53 IR (4MP) cameras. Each subject performs a set of predefined motions, including T-pose, hand gestures, and squats, in 10-second sequences at 30 FPS. From each sequence, 4–5 high-fidelity frames are selected, yielding over 600 3D scans. Each sample includes a 40K-face mesh, a 4K texture map, and accurately registered SMPL-X parameters, with a wide range of garment styles (120 total).

\noindent \textbf{THuman2.0}~\citep{yu2021function4d} offers 500 high-resolution human scans captured using a dense DSLR rig. Each sample consists of a detailed 3D mesh paired with a high-quality texture map, covering a wide range of body shapes and clothing types. This dataset serves as a clean source of diverse clothed human geometry.

\subsubsection{Rendering Procedure}  \label{sec_sup:rendering}
For each 3D human scene, we render 16 views in total, comprising 8 orthographic and 8 perspective images. Each view includes RGB images, normal maps, depth maps, and segmentation masks, rendered from both SMPL-X meshes and original scanned meshes. All views share the same azimuth angles to allow consistent comparison across projection types. Orthographic views are rendered with slight variation in elevation (randomly sampled in the range $[-10^\circ, +10^\circ]$) to improve robustness against vertical pose and camera variations. Perspective views, on the other hand, utilize elevation values randomly sampled from a broader range of $[-20^\circ, +45^\circ]$ and distances between $2.0$ and $6.0$ units to enhance variability and generalization.

Camera extrinsic parameters (rotation $\mathbf{R}$ and translation $\mathbf{T}$) are generated using a virtual camera located at the specified distance and orientation, always looking at the origin. These parameters are derived via the \texttt{look\_at\_view\_transform} function in PyTorch3D~\citep{ravi2020accelerating}.

Prior to rendering, all meshes are normalized to a canonical space. This is done by computing the bounding box of the vertex positions, determining the center and maximal axis length, and scaling the mesh such that it fits within a unit cube. For datasets like CustomHumans~\citep{ho2023learning} and THuman2.0~\citep{yu2021function4d}, we additionally introduce random jittering to the mesh center (on the $X$ and $Y$ axes) to prevent overfitting and encourage generalization. The normalized vertex positions $\mathbf{v}_{\text{norm}}$ are computed by:
\[
\mathbf{v}_{\text{norm}} = \frac{\mathbf{v} - \mathbf{c}}{s / 2}
\]
where $\mathbf{c}$ is the computed center of the bounding box and $s$ is the padded bounding box size.

For orthographic rendering, we fix the camera distance at $3.0$ units and maintain consistent scale across all views. For perspective rendering, focal lengths are automatically determined based on the normalized mesh size and camera distance, with additional jitter applied up to 20\% to simulate realistic monocular variation.

Rendering is performed using the PyTorch3D \texttt{MeshRenderer}, configured with either orthographic or perspective camera models. RGB images are rendered using a Phong shading model with ambient or point lighting depending on the dataset. Depth maps are extracted from the rasterizer’s $z$-buffer. Normal maps are generated by interpolating face vertex normals in view space. Instance segmentation masks are computed per-pixel using face-to-instance ID mappings. When contact information is available, contact masks are also rendered for the case of multi-human dataset by identifying faces associated with contact regions and projecting them to image space. Small holes in the resulting binary masks are filled using post-processing.

All outputs are rendered at a base resolution of $768 \times 768$ pixels. For training, we randomly select a reference view and sample six additional views at fixed relative azimuth angles of $\{0^\circ, 45^\circ, 90^\circ, 180^\circ, 270^\circ, 315^\circ\}$, resulting in a 6-view training input for each instance.

\subsubsection{Masking Strategy for Occlusion Simulation}   \label{sec_sup:masking}

\begin{figure*}[t]
    \centering
    \includegraphics[width=.9\linewidth]{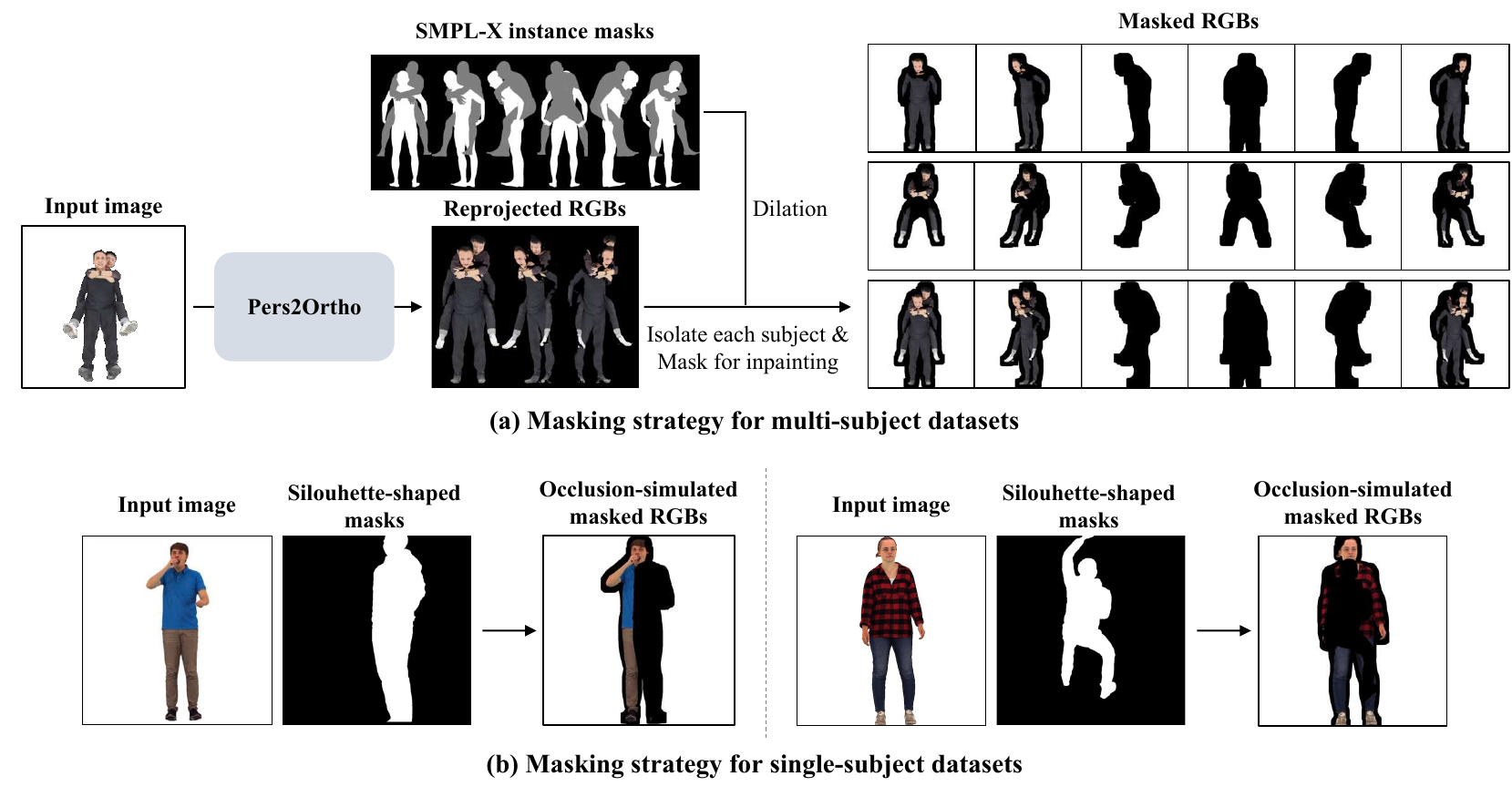}
    \vspace{-0.5em}
\caption{Illustration of masking strategy for occlusion simulation. (a) For multi-subject datasets, SMPL-X instance masks are used to isolate each subject and specify regions to inpaint. (b) For single-subject datasets, we simulate oclusion through silhouette-shaped masks.}
    \label{fig_supp_method_masking}
    \vspace{-1em}
\end{figure*}

As shown in ~\ref{fig_supp_method_masking}, to simulate realistic visibility and occlusion in multi-human 3D scenes, we construct masked images ${x^{(i)}_{\text{mask}}}$ from reprojected point clouds ${x^{(i)}_{\text{pcd}}}$ and visibility masks ${M}_{\text{vis}}$, which indicate regions of missing observation in each canonical view $\mathcal{C}_i$.

To define the canonical visibility views, we use the reprojected RGB images captured at \{0$^\circ$, 45$^\circ$, 315$^\circ$\}. For all other viewpoints, only the SMPL-X instance masks are used for occlusion simulation without relying on PCD.

During training, we generate masked RGB images $\{{x^{(i)}_{\text{mask}}}\}_{i=0}^{5}$ to guide inpainting or occlusion-aware reconstruction networks: background regions are labeled as 1, while occluded or missing areas are labeled as 0 and the visible region with the pixel values. These masks provide supervision for learning to reconstruct or inpaint plausible content in the occluded regions.

In multi-subject datasets such as Hi4D~\citep{yin2023hi4d}, where mutual occlusion naturally occurs, we use SMPL-X instance masks to isolate each subject. These masks are further dilated to account for peripheral structures such as garments and hair, ensuring that edge regions around the human silhouette are adequately covered for inpainting tasks. We apply a dilation operation with a kernel size of 61.

In single-subject datasets such as CustomHumans~\citep{ho2023learning} and THuman2.0~\citep{yu2021function4d}, we simulate occlusion by randomly selecting one of two masking strategies with equal probability (0.5): (1) silhouette-shaped masks that resemble human figures, randomly scaled and positioned near the image center to mimic the presence of an occluder, or (2) random hole-based masks, such as freeform or template-driven occlusions, which introduce unstructured masking artifacts. This augmentation scheme enables the model to generalize to a wide range of occlusion scenarios, even in the absence of multiple real subjects.

\subsubsection{Training Procedure}  \label{sec_sup:training}
\begin{figure*}[t]
    \centering
    \includegraphics[width=.9\linewidth]{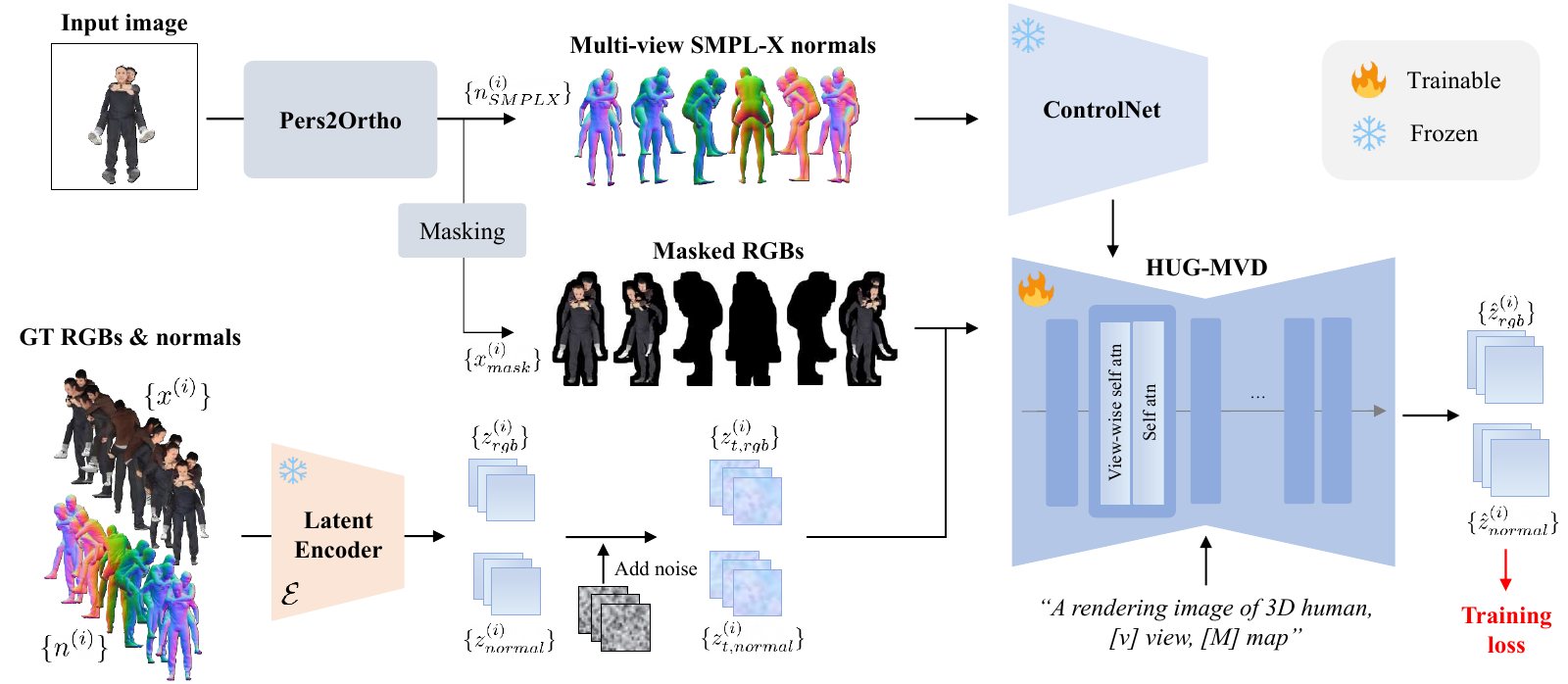}
    \vspace{-0.5em}
\caption{Illustration of the training procedure of HUG-MVD. The model takes occluded RGB images and SMPL-X normal maps from six views and learns to reconstruct complete RGB and normal views via a multi-view diffusion model. SMPL-X guidance is injected via ControlNet, with attention applied across views and modalities for coherent multi-human reconstruction.}
    \label{fig_supp_method_training}
    \vspace{-1em}
\end{figure*}

The objective of our training procedure is to reconstruct complete RGB and normal images from partially visible, occluded inputs by leveraging a multi-view diffusion model. An overview of the training process is illustrated in Fig.~\ref{fig_supp_method_training}. Each training sample consists of six canonical views per scene. The model receives the following inputs:  
(1) Occluded RGB images reprojected from point clouds using the masking strategies described earlier: $\{x^{(i)}_{\text{mask}}\}_{i=0}^{5}$, and  
(2) Corresponding SMPL-X normal maps providing geometric structure: $\{n^{(i)}_{\text{SMPLX}}\}_{i=0}^{5}$.  
The supervision targets are:  
(1) Ground-truth RGB images: $\{x^{(i)}\}_{i=0}^{5}$, and  
(2) Ground-truth normal maps: $\{n^{(i)}\}_{i=0}^{5}$.

We initialize the model from {PSHuman}~\citep{li2024pshuman}, a pre-trained diffusion model designed to synthesize six RGB and normal views from a single RGB image. Notably, PSHuman is trained exclusively on {single-human datasets}, including {THuman2.0}~\citep{yu2021function4d} and {CustomHumans}~\citep{ho2023learning}, and the open-source version of PSHuman does not support SMPL-X conditioning. To overcome this limitation and extend the model to multi-human scenes with explicit geometry input, we integrate {ControlNet}~\citep{zhang2023adding} into the architecture. This enables the model to utilize SMPL-X normal maps as structural guidance during training.

Our model operates in the latent space defined by the variational autoencoder (VAE) from Stable Diffusion 2.1~\citep{rombach2022high}. Each ground-truth RGB image $x^{(i)}$ and normal map $n^{(i)}$ is encoded into latent variables $z^{(i)}_{\text{rgb}}$ and $z^{(i)}_{\text{normal}}$ using the VAE encoder. The model then predicts the residual noise added to these latents during the forward diffusion process. The training objective is defined as:

\begin{equation}
\small
\begin{split}
\mathcal{L}_{\text{diff}} = \sum_{i=0}^{5} \Bigl( 
    &\mathbb{E}_{t, \epsilon} \Bigl[ \|\epsilon - \epsilon_\theta(z_{t,\text{rgb}}^{(i)}, t, x^{(i)}_{\text{mask}}, n^{(i)}_{\text{SMPLX}})\|_2^2 \Bigr] \\
    &+ \mathbb{E}_{t, \epsilon} \Bigl[ \|\epsilon - \epsilon_\theta(z_{t,\text{normal}}^{(i)}, t, x^{(i)}_{\text{mask}}, n^{(i)}_{\text{SMPLX}})\|_2^2 \Bigr]
\Bigr)
\label{eq:diffusion_loss}
\end{split}
\end{equation}
\normalsize

Here, $z_{t, \text{rgb}}^{(i)}$ and $z_{t, \text{normal}}^{(i)}$ denote the noisy latent variables at timestep $t$, and $\epsilon$ is the Gaussian noise sampled during training. The denoising model $\epsilon_\theta$ learns to recover the clean signal conditioned on the masked RGB inputs and the SMPL-X geometry.

Two key attention modules are employed to support effective multi-view generation.  
First, following prior work~\citep{li2024era3d,li2024pshuman}, we apply row-wise multi-view self-attention independently within each modality (RGB or normal), allowing the model to correlate across the six canonical views.  
Second, to enable information exchange between RGB and normal modalities, we apply self-attention across all latent tokens, allowing cross-modality attention between RGB and normal features at each view. This joint RGB-normal attention mechanism is also inspired by~\citep{li2024era3d,li2024pshuman}.

During training, only the U-Net parameters are optimized, while the remaining components such as the CLIP image encoder and the VAE are kept frozen. Text prompt embeddings are injected via cross-attention using the template: \texttt{``a rendering image of 3D human, [V] view, [M] map''}, where [V] indicates the view direction (e.g., \texttt{front}, \texttt{left}, \texttt{face}) and [M] specifies the modality (\texttt{color} or \texttt{normal}).

We also experimented with classifier-free guidance (CFG) by applying conditioning dropout, following prior work~\citep{ho2021classifier}. However, since CFG yielded only marginal performance improvements at test time, it was not used during inference.

To address occlusion and mesh collision issues in multi-human scenes, we optionally apply contact-aware binary masks on Hi4D~\citep{yin2023hi4d} samples. These masks suppress gradient updates in regions of body-to-body contact where the mesh supervision is less reliable due to penetration artifacts or annotation noise.

The model is trained on a single NVIDIA A100 GPU (80GB) with a batch size of 16 and gradient accumulation steps of 8. We use the Adam~\citep{kingma2014adam}  optimizer with a learning rate of $5 \times 10^{-6}$, $\beta_1 = 0.9$, and $\beta_2 = 0.999$. Training follows a two-stage curriculum: (1) pre-training without inpainting masks for 1,000 steps, followed by (2) fine-tuning with inpainting masks for another 1,000 steps to simulate occlusion. The entire training takes approximately two days, with peak GPU memory usage of around 62GB. A DDPM scheduler with 1,000 diffusion steps is used throughout training.

\subsubsection{Inference Procedure} \label{sec_sup:inference}
\begin{figure*}[t]
    \centering
    \includegraphics[width=.9\textwidth]{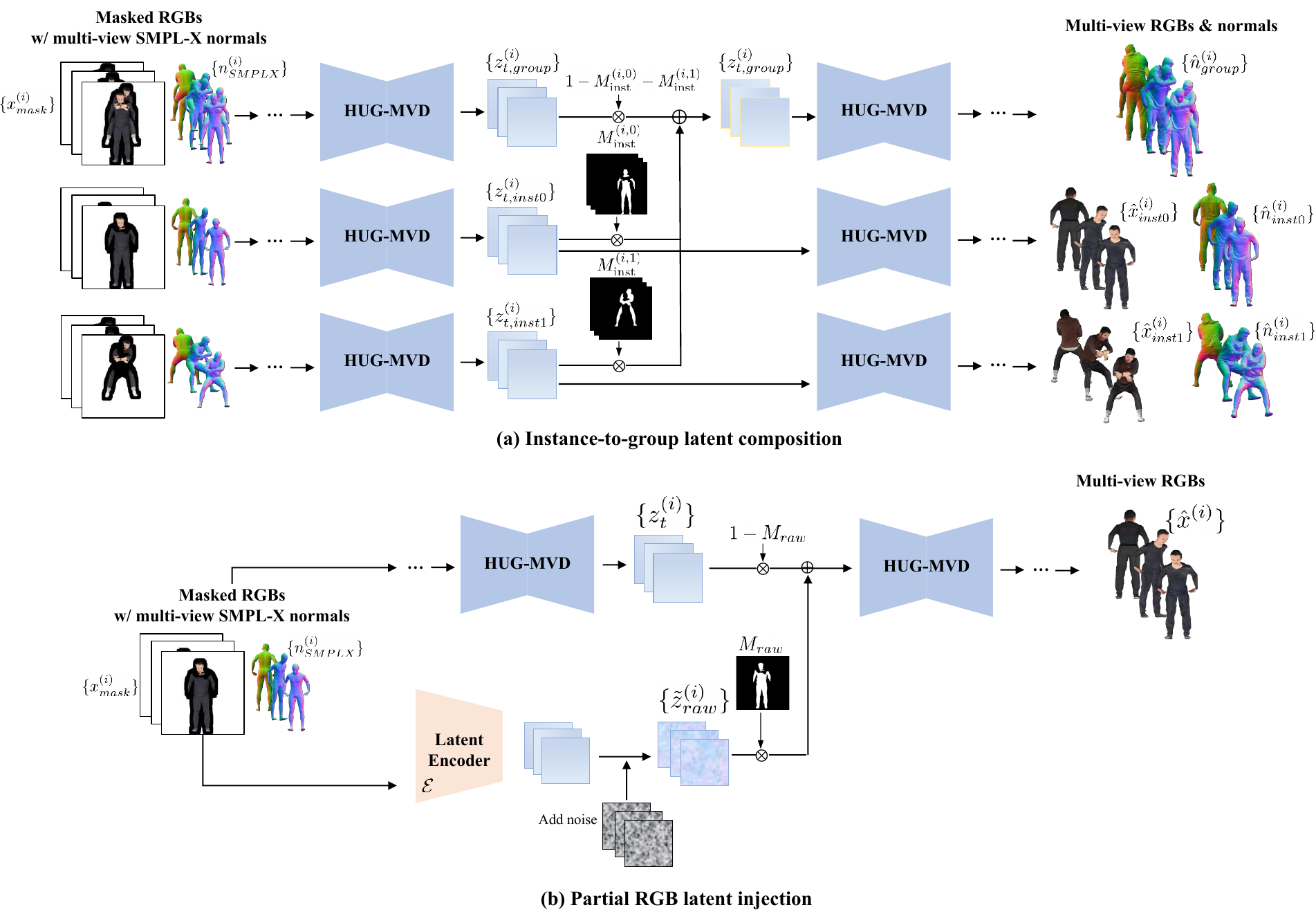}
    \vspace{-0.5em}
\caption{Illustration of the inference procedure of HUG-MVD. (a) Instance-to-group latent composition and (b) partial RGB latent injection for RGB synthesis.}
    \label{fig_supp_method_inference}
    \vspace{-1.5em}
\end{figure*}
At inference time, the model predicts complete normal maps $\{\hat{n}^{(i)}\}$ and synthesizes complete RGB images $\{\hat{x}^{(i)}\}$ across all views from partially visible RGB inputs and canonical-view SMPL-X normal maps as illustrated in Fig.~\ref{fig_supp_method_inference}.

To maintain consistency between group-level and instance-level reconstructions, we perform latent-space composition. At each diffusion timestep $t$, instance-specific latents $z_{t,\text{inst}(k)}^{(i)}$ are softly injected into the group-level latent $z_{t,\text{group}}^{(i)}$ using binary masks $\mathbbm{1}_{\text{inst}}^{(i,k)}$ and a blending ratio $\alpha_{\text{gi}} = 0.8$:

\begin{equation}
\small
\begin{split}
z_{t, \text{group}}^{(i)} \leftarrow &\sum_{k=1}^{K} \Bigl[
    \alpha_{\text{gi}} \cdot \mathbbm{1}_{\text{inst}}^{(i,k)} \cdot z_{t, \text{inst}(k)}^{(i)} 
    + (1 - \alpha_{\text{gi}}) \cdot \mathbbm{1}_{\text{inst}}^{(i,k)} \cdot z_{t, \text{group}}^{(i)}
\Bigr] \\
&\quad + \left(1 - \sum_{k=1}^{K} \mathbbm{1}_{\text{inst}}^{(i,k)}\right) \cdot z_{t, \text{group}}^{(i)},
\label{eq:inst2group_compose}
\end{split}
\end{equation}
\normalsize

This mechanism allows high-frequency details from instance-specific predictions to be integrated into the global group representation, improving surface continuity in multi-human scenes.

Also, to further enhance RGB quality, we inject latent signals from partially visible RGB inputs. At each diffusion timestep $t$, we generate a noisy version $\tilde{z}_{\text{raw}}$ of the raw RGB latent (restricted to visible regions) and blend it into the current latent $z_t$ using a binary mask $\mathbf{m}_{\text{raw}}$ and mixing ratio $\alpha_{\text{pcd}} = 0.8$:

\begin{equation}
\small
z_t \leftarrow \mathbf{m}_{\text{raw}} \cdot \left[
\alpha_{\text{pcd}} \cdot \tilde{z}_{\text{raw}} + (1 - \alpha_{\text{pcd}}) \cdot z_t
\right] + (1 - \mathbf{m}_{\text{raw}}) \cdot z_t,
\end{equation}
\normalsize

This operation is applied exclusively to the RGB branch and aims to reinforce reliable visual priors in visible areas, improving fidelity in occluded or ambiguous regions. We apply this injection selectively to low-confidence views (e.g., non-source views) to avoid overwriting already plausible outputs.

We use a DDIM scheduler with $\eta = 1.0$ and perform 40 denoising steps per sample. 
We also used $ \alpha_{\text{gi}}=\alpha_{\text{pcd}}=0.8$.  Inference for all group-level and instance-level multi-view RGB and normal maps takes 60.16 seconds, using 34.76GB of VRAM on an NVIDIA A100.

\subsection{Human Group-Instance Geometry Reconstruction (HUG-GR)} \label{sec_sup:huggr}

\begin{figure}[t]
    \centering
    \includegraphics[width=0.95\linewidth]{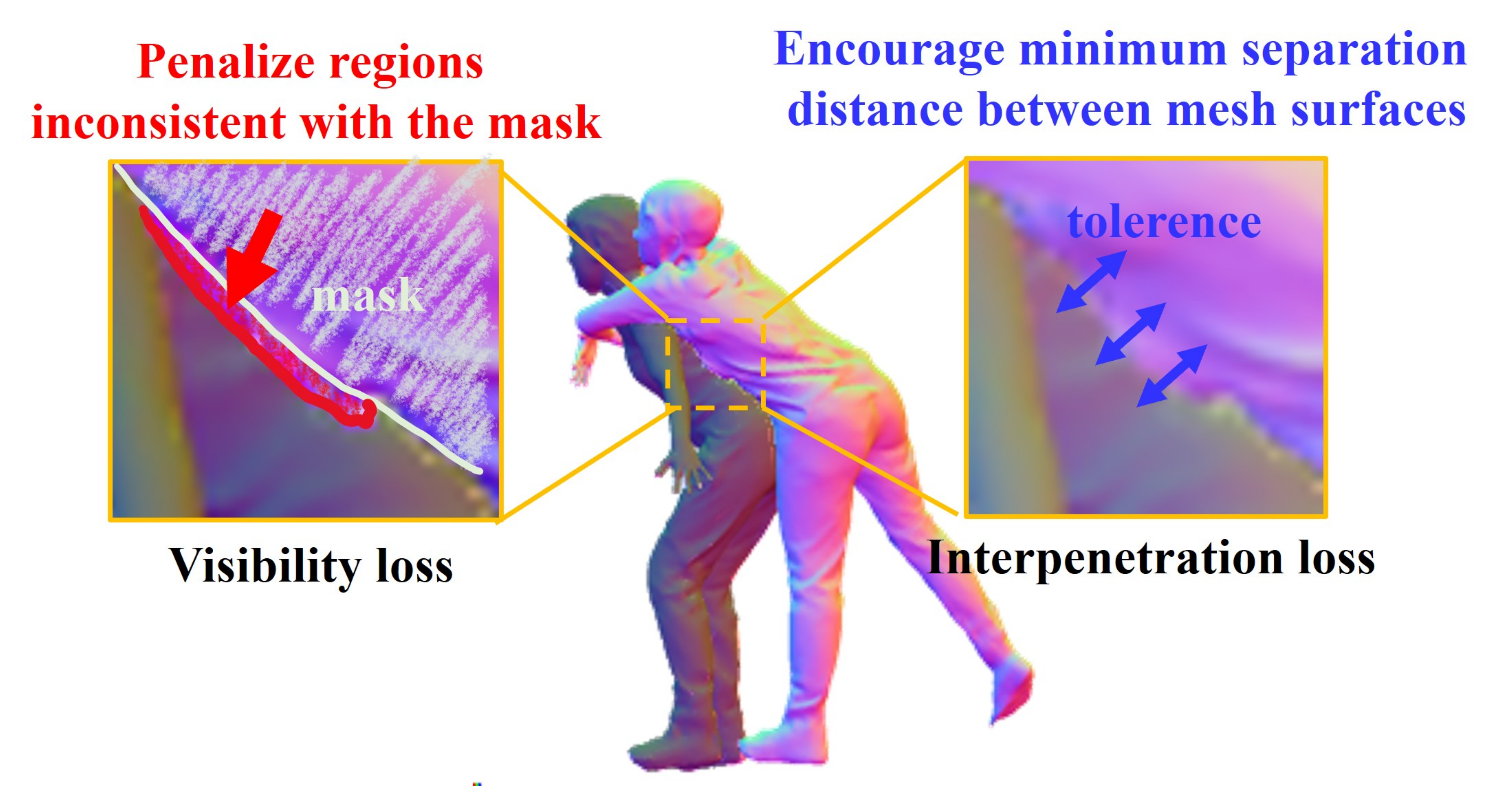}
    \vspace{-0.5em}
    \caption{Illustration of the interpenetration loss and visibility loss. The interpenetration loss penalizes body-part collisions (blue arrows), while the visibility loss enforces alignment between rendered masks and ground-truth visibility (red region).}
    \label{fig_supp_method_loss_hug_gr}
    \vspace{-1em}
\end{figure}

Here, we provide a detailed explanation of the two geometry-level supervision terms—\textit{interpenetration loss} and \textit{visibility loss}. As illustrated in Fig.~\ref{fig_supp_method_loss_hug_gr}, these losses play complementary roles in enhancing geometric plausibility and part-level visibility consistency during group-instance reconstruction.
Z

\noindent \textbf{Interpenetration Loss.} To prevent anatomically implausible overlaps between articulated body parts, we define an interpenetration loss that penalizes violations among predefined part pairs $(i,j)\in\mathrm{V}$, where $\mathrm{tol}$ is the tolerance set encoding pairs subject to collision constraints inspired by~\citep{huang2024closely}. 
To determine $\mathrm{V}$, we first compute a contact map based on vertex-to-vertex distances on the initial SMPL-X meshes, which identifies all potential contact regions. The resulting contact regions constitutes our tolerance set. 
For each pair, let $s_1^{i,j}$ and $s_2^{i,j}$ be the closest surface points on parts $i$ and $j$. We apply a smooth barrier around a tolerance $\mathrm{tol}$  (default $= 5\times10^{-4}$), with $T=\max(0.25\,\mathrm{tol},10^{-5})$ acting as a smoothing temperature:
\begin{equation}
\small
\mathcal{L}_{\mathrm{pen}}
= \operatorname{mean}_{(i,j)\in\mathrm{V}}\!\Big[
T\,\ln\!\big(1 + e^{(\mathrm{tol}-\lvert s_1^{i,j}-s_2^{i,j}\rvert)/T}\big)
\Big].
\end{equation}
\normalsize

This term encourages a minimum surface separation between adjacent parts (e.g., thighs vs. calves), helping to reduce penetration artifacts while preserving flexibility for naturally close configurations such as seated or folded poses.

\noindent \textbf{Visibility Loss.}  
To improve spatial alignment in crowded scenes, we supervise visibility using rendered segmentation masks. For each body part \(b\) in instance \(k\), we penalize visibility mismatches using:

\begin{equation}
\small
\mathcal{L}_{\mathrm{vis}} =
  \frac{1}{2B}
  \sum_{k=1}^{K} \sum_{b=1}^{B}
     \frac{E^k_b}{M^k_b + \epsilon},
\end{equation}
\normalsize

where \(E^k_b\) is the number of incorrectly occluded pixels and \(M^k_b\) the total visible pixels in the ground truth. This encourages accurate silhouette and occlusion boundaries, particularly in group interactions.

\noindent \textbf{Adaptive Region-Specific Optimization.}  
To balance global stability and preservation of local details, we apply region-specific optimization strategies. In particular, lower learning rates are used for vertices located in semantically and geometrically complex regions such as the hands and face. This allows the model to better preserve high-frequency features provided in the initial SMPL-X mesh in these areas while still allowing flexible carving of geometric features, such as clothing, in other regions. We determine how close a vertice is to a complex region using the optimized SMPL-X joint positions of the hands and face. And use sigmoid blending to derive the actual learning rate. 
Formally, the vertice-wise adaptive learning rate $\alpha{}_{\mathrm{v}}$ for vertice $v$ is:
\begin{equation}
\small
\alpha_{\mathrm{v}} = \alpha_{\mathrm{base}} \cdot
    \frac{1}{e^{-(200 d_v + 10)} + 1},
\end{equation}
Where $\alpha{}_{\mathrm{base}}$ is the base learning rate and $d_v$ is the minimum distance between $v$ and the set of all SMPL-X joint vertices in consideration denoted as $J_{\text{SMPL-X}}$. Then, $d_v$ is:
\begin{equation}
\small
d_v = \min_{j \in J_{\text{SMPL-X}}}\|v - j\|
\end{equation}
Thus, we assign lower learning rates to vertices with smaller $d_v$ (i.e. vertices closer to hands or the face). As illustrated in Fig. 6(b),
this adaptive strategy results in sharper reconstructions of fine regions (e.g., fingers, facial contours) while maintaining coherence in broader anatomical parts like the torso or limbs.

We optimize the mesh over 200 iterations with a learning rate of 0.01, \(\lambda_{\text{group}} = 1.0\), \(\lambda_{\text{inst}} = 0.2\), \(\lambda_{\text{pen}} = 30.0\) and \(\lambda_{\text{vis}} = 1.0\).
HUG-GR takes 125.47 seconds and consumes 7.58GB of VRAM on an A100 GPU.

\subsection{Occlusion- and View-Aware Texture Fusion} \label{sec_sup:texture}

\begin{figure*}[t]
    \centering
    \includegraphics[width=.9\textwidth]{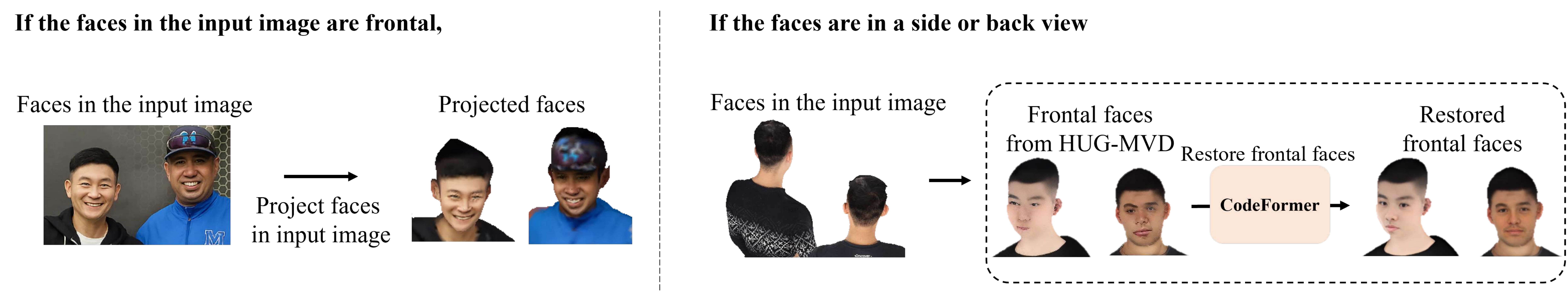}
    \vspace{-0.5em}
   \caption{View-aware face restoration enhances frontal views using landmark-guided inpainting.}
    \label{fig_supp_method_tex_face_restroation}
    \vspace{-1em}
\end{figure*}

\begin{figure*}[t]
    \centering
    \includegraphics[width=0.8\textwidth]{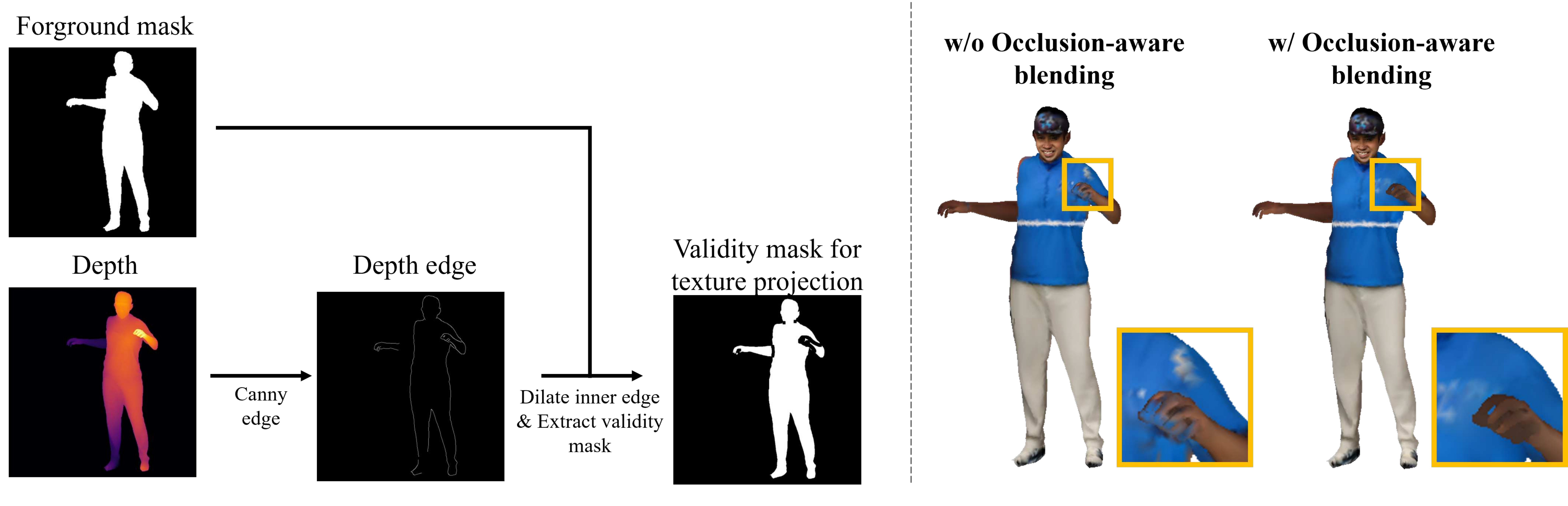}
    \vspace{-0.5em}
\caption{Illustration of the occlusion-aware blending strategy. Edge-aware confidence masks, computed from depth discontinuities, suppress artifacts near occlusion boundaries, resulting in cleaner silhouettes and improved cross-view consistency.}
    \label{fig_supp_method_tex_occ_aware_blending}
    \vspace{-1em}
\end{figure*}

To construct coherent and high-quality full-body textures, we fuse multi-view RGB predictions into a unified texture. To improve texture fidelity and suppress artifacts, we introduce two important enhancements: \textit{view-aware face restoration} and \textit{occlusion-aware blending}.

\noindent \textbf{View-Aware Face Restoration.}  
Faces captured from extreme angles or under occlusion often exhibit degraded appearance. To address this, as shown in Fig.~\ref{fig_supp_method_tex_face_restroation}, we first analyze each view using facial landmarks and SMPL-X head orientation to estimate the relative frontalness of the face. Among the six RGB predictions per instance, we select the two most frontal views. If the source view is used for the face, we directly use its content. If not, we perform face inpainting on the most frontal view using CodeFormer~\citep{zhou2022towards}, where a soft circular mask is generated using warped 5-point facial landmarks. In both cases, the enhanced face region is warped back and blended into the original view using inverse affine transformation. This step improves the final texture synthesis especially in the face region.

\noindent \textbf{Occlusion-Aware Blending.}
As illustrated in Fig.~\ref{fig_supp_method_tex_occ_aware_blending}, to prevent ghosting and bleeding artifacts near occlusion boundaries, we employ edge-aware confidence masking guided by view-dependent depth maps. Depth edges are first extracted using the Canny filter, and the resulting edge map is dilated with a fixed kernel to define an exclusion zone. We retain only the pixels that lie within foreground regions and are sufficiently distant from detected depth discontinuities. These reliable pixels are used to generate a binary confidence mask $C_i$ for each view. The final contribution of a view’s texture projection $T_i$ is modulated by this mask as $T_i' = C_i \cdot T_i$. This occlusion-aware blending strategy effectively suppresses unstable regions near self-occlusion edges, resulting in cleaner object silhouettes and improved consistency across views. We apply a dilation operation with a kernel size of 21.

Our texture fusion takes  14.49 seconds and consumes 4.95GB of VRAM on an A100 GPU.

\twocolumn[\vspace*{0pt}]
\section{Evaluation Details} \label{sec_sup:evaluation}

\subsection{Evaluation Settings} \label{sec_sup:evalsetting}
We evaluate our method in comparison to prior works across three categories:
methods of single human reconstruction from a single image,
methods of multi-human reconstruction from multi-view images, and
methods of multi-human reconstruction from videos.
Since there is no publicly available baseline implementation that directly performs multi-human reconstruction from a single image as represented in Tab.~\ref{tab_related_methods}, 
to ensure a fair comparison, we follow the evaluation protocol of~\citep{cha20243d} by adapting related methods in each category under consistent settings. To isolate the effect of SMPL-X prediction from the reconstruction process, all main comparisons are conducted using ground-truth SMPL-X. Results using predicted SMPL-X are included in Sec.~\ref{sec_sup:baseline} of the supplementary.

\begin{table*}[!h]\centering
\caption{Comparison of recent 3D human reconstruction methods. HUG3D supports multi-human reconstruction from a single image with both geometry and texture.}
\vspace{-0.5em}
\label{tab_related_methods}
\scriptsize
\centering
\begin{adjustbox}{width=0.85\linewidth}
\begin{tabular}{lccccc}
\toprule
Method & Multi- or Single Human & Input Type & Geometry & Texture & Publicly Available \\
\midrule
ECON~\citep{xiu2023econ} & Single human & Single image & \checkmark & \xmark & \checkmark \\
SiTH~\citep{ho2024sith} & Single human & Single image & \checkmark & \checkmark & \checkmark \\
SIFU~\citep{zhang2024sifu} & Single human & Single image & \checkmark & \checkmark & \checkmark \\
PSHuman~\citep{li2024pshuman} & Single human & Single image & \checkmark & \checkmark & \checkmark \\
DeepMultiCAP~\citep{zheng2021deepmulticap} & Multi-human & Multi-view images & \checkmark & \xmark & \checkmark \\
Multiply~\citep{jiang2024multiply} & Multi-human & Video & \checkmark & \checkmark & \checkmark \\
Cha et al.~\citep{cha20243d} & Multi-human & Single image & \checkmark & \xmark & \xmark \\
\textbf{HUG3D (Ours)} & Multi-human & Single image & \checkmark & \checkmark & \checkmark~(upon acceptance) \\
\bottomrule
\end{tabular}
\end{adjustbox}
\vspace{-1em}
\end{table*}

\noindent \textbf{Single human reconstruction from single image.} For methods of originally designed for single human reconstruction~\citep{zhang2024sifu, ho2024sith, li2024pshuman, xiu2023econ}, we adapted them to the multi-human setting as follows. Ground-truth instance segmentation masks were used to isolate each person in the input image. Each individual was then reconstructed independently using the corresponding method. Since the outputs lie in different coordinate frames, we performed a canonicalization procedure to align all reconstructions into a shared space. Specifically, for each instance, we first predicted the SMPL-X mesh using the method's native estimator. We then computed a similarity transformation—comprising scale, rotation, and translation—that aligns the ground-truth SMPL-X mesh to the predicted one. The ground-truth SMPL-X mesh was transformed into the predicted space before reconstruction, and the reconstruction output was transformed back to the ground-truth space via the inverse transformation, allowing the reconstructed scene to be composited consistently. We also evaluate PSHuman-multi, which applies the single-person reconstruction pipeline PSHuman~\citep{li2024pshuman} directly to uncropped multi-person images. Since we use ground-truth SMPL-X for all evaluations, we omit the SMPL-X optimization process for baselines that originally involve it.

\noindent \textbf{Multi-human reconstruction from multi-view image.} For multi-view baselines~\citep{zheng2021deepmulticap}, we provided only a single view as input for inference, to ensure comparability with our single-image reconstruction setting.

\noindent \textbf{Multi-human reconstruction from videos.} Similarly, for video-based baselines~\citep{jiang2024multiply}, we provided only a single image as the first frame for inference.

\subsection{Evaluation Dataset} \label{sec_sup:evaldataset}
\noindent \textbf{MultiHuman~\citep{jiang2024multiply}.} To facilitate both quantitative and qualitative evaluation of reconstructed meshes, we rendered perspective-view images from the MultiHuman dataset using a multi-view setup. Our evaluation covers a total of 20 two-person scenes, including 6 closely interactive cases (sequences 8, 23, 24, 250, 252, 253) and 14 naturally interactive scenes (sequences 12, 16, 17, 18, 19, 20, 22, 30, 226, 244, 249, 251, 255, 256). For ablation studies, we focus on the closely interactive cases.

For each scene, we rendered the meshes from 4 distinct camera viewpoints, generated by sampling a random azimuth and adding fixed offsets of \(\{0^\circ,\,90^\circ,\,180^\circ,\,270^\circ\}\), resulting in views uniformly distributed around the subject. The elevation angles were randomly sampled in the range \([-20^\circ, \,45^\circ]\), and the camera-to-subject distances were sampled uniformly from \([2.0, 6.0]\), simulating varying levels of zoom and perspective distortion.

To ensure scale-invariant and consistently framed rendering, each mesh was normalized to fit within a unit cube centered at the origin. This was achieved by computing the mesh's axis-aligned bounding box and uniformly scaling it based on the maximum side length.

\noindent \textbf{In-the-wild.} For our qualitative evaluation with in-the-wild images, we leveraged OpenAI's Sora service to obtain a diverse set of test images. Sora performs a web-based image search for user-specified concepts, reconstructs novel scenes by referencing those search results, and synthesizes new images that reflect real-world variation. The resulting Sora outputs—whose content is derived from Internet-sourced photos—were then used as our "in-the-wild" evaluation set, ensuring that our method is tested on unconstrained, naturally diverse imagery.

\subsection{Evaluation Metrics}  \label{sec_sup:metrics}
We employ a comprehensive set of metrics to evaluate both the geometric and texture quality of reconstructed multi-human meshes. These metrics cover surface accuracy, physical realism, and perceptual quality. Here, P and Q are point clouds sampled from the predicted and ground-truth meshes.

\noindent \textbf{Chamfer Distance (CD).}
Chamfer Distance measures the bidirectional discrepancy between the predicted and ground-truth surfaces. We uniformly sample 100,000 points from each mesh surface and compute the average closest-point distance from the predicted points to the ground-truth surface and vice versa.
The final CD score is defined as:
\[
\small
\text{CD}(P, Q) = \frac{1}{|P|} \sum_{p \in P} \min_{q \in Q} \|p - q\|_2 + \frac{1}{|Q|} \sum_{q \in Q} \min_{p \in P} \|q - p\|_2.
\]
\normalsize

A lower CD indicates a more accurate reconstruction that closely matches the geometry of the ground truth in both completeness and precision.

\noindent \textbf{Point-to-Surface Distance (P2S).}
P2S measures the unidirectional accuracy of the predicted surface with respect to the ground-truth shape. Specifically, it measures the average Euclidean distance from each point sampled on the predicted mesh to the closest point on the ground-truth surface:
\[
\text{P2S}(P \rightarrow Q) = \frac{1}{|P|} \sum_{p \in P} \min_{q \in Q} \|p - q\|_2^2.
\]
P2S emphasizes surface accuracy without penalizing missing parts, and lower values indicate closer alignment to the reference shape.

\noindent \textbf{Normal Consistency (NC).}
NC measures the angular similarity between surface normals on the predicted and ground-truth meshes. For each point, we compare the normal vector at that point with the normal vector at the closest point on the opposite surface. The final score is averaged bidirectionally:
\[
\small
\begin{split}
\text{NC}(P, Q) = \frac{1}{2|P|} \sum_{p \in P} \left(1 - \langle \mathbf{n}_p, \mathbf{n}_{\text{NN}(p, Q)} \rangle \right)
\\ + \frac{1}{2|Q|} \sum_{q \in Q} \left(1 - \langle \mathbf{n}_q, \mathbf{n}_{\text{NN}(q, P)} \rangle \right),
\end{split}
\]
\normalsize
where $\langle \cdot, \cdot \rangle$ denotes the dot product between unit normals, and $\text{NN}(\cdot)$ returns the nearest neighbor in the opposite set. A higher NC indicates better preservation of surface orientations and local detail.

\noindent \textbf{F-score.}
F-score evaluates both precision and recall of the predicted surface points with respect to a ground-truth reference under a distance threshold $\tau$. We use $\tau=1$cm. Precision measures the percentage of predicted points that lie within $\tau$ of the ground-truth surface, while recall measures the converse. F-score is defined as the harmonic mean of the two:
\[
\text{F-score} = \frac{2 \cdot \text{Precision} \cdot \text{Recall}}{\text{Precision} + \text{Recall}}.
\]
This metric rewards reconstructions that are both accurate and complete.

\noindent \textbf{Bounding Box IoU (bbox-IoU).}
We compute the 3D Intersection-over-Union (IoU) of axis-aligned bounding boxes of the predicted and ground-truth meshes:
\[
\text{IoU}_{\text{bbox}} = \frac{\text{vol}(B_{\text{pred}} \cap B_{\text{gt}})}{\text{vol}(B_{\text{pred}} \cup B_{\text{gt}})},
\]
where $B_{\text{pred}}$ and $B_{\text{gt}}$ are the predicted and ground-truth bounding boxes, respectively. This metric evaluates global layout similarity and spatial coverage.

\noindent \textbf{L2 Normal Error.}
We assess surface detail preservation by computing the per-pixel $L_2$ distance between rendered normal maps of the predicted and ground-truth meshes. This is done across four orthographic views at azimuth angles $\{0^\circ, 90^\circ, 180^\circ, 270^\circ\}$:
\[
\text{L2-NormErr} = \frac{1}{N} \sum_{i=1}^{N} \| \mathbf{n}_i^{\text{pred}} - \mathbf{n}_i^{\text{gt}} \|_2^2.
\]
We also report this error computed within occluded regions, to specifically assess reconstruction quality under visual occlusion.

\noindent \textbf{Contact Precision (CP).}
To evaluate the physical plausibility of multi-human reconstruction, we measure the alignment of predicted inter-body contact regions with the ground truth. This metric quantifies how accurately the predicted contact points reflect the true contact between two human bodies.

Let $\hat{M}_1$ and $\hat{M}_2$ be the predicted meshes, and $M_1$ and $M_2$ the corresponding ground-truth meshes. Denote their vertex sets as $\hat{V}_1$, $\hat{V}_2$, $V_1$, and $V_2$, respectively. A vertex is considered in contact if it lies within a threshold distance $\delta$ from the other mesh.

First, we define the ground-truth contact region $\mathcal{C}_\text{gt}$ as:
\[
\mathcal{C}_\text{gt} = \left\{ v \in V_1 \cup V_2 \,\middle|\, \min_{v' \in V_2 \cup V_1} \|v - v'\|_2 < \delta \right\},
\]
and similarly, the predicted contact region $\mathcal{C}_\text{pred}$ as:
\[
\mathcal{C}_\text{pred} = \left\{ \hat{v} \in \hat{V}_1 \cup \hat{V}_2 \,\middle|\, \min_{\hat{v}' \in \hat{V}_2 \cup \hat{V}_1} \|\hat{v} - \hat{v}'\|_2 < \delta \right\}.
\]

We then compute precision by counting the proportion of predicted contact points that are close to the ground-truth contact region:
\[
\text{CP} = \frac{1}{|\mathcal{C}_\text{pred}|} \sum_{\hat{v} \in \mathcal{C}_\text{pred}} \mathbf{1}\left[ \text{NN}(\hat{v}, V_\text{gt}) \in \mathcal{C}_\text{gt} \right],
\]
where $\text{NN}(\hat{v}, V_\text{gt})$ denotes the nearest vertex to $\hat{v}$ among all ground-truth vertices.

We set the contact threshold $\delta = 0.01$ meter. A higher CP indicates better prediction of physically plausible inter-human contacts.

\noindent \textbf{Texture Fidelity.}
To assess the perceptual quality of the reconstructed texture, we evaluate the rendered mesh images against ground-truth renderings using three standard image similarity metrics: PSNR, SSIM, and LPIPS.

Given the predicted image $\hat{I}$ and the ground-truth image $I$ rendered from the same view, we compute:

\textbf{Peak Signal-to-Noise Ratio (PSNR)}:
  \[
  \text{PSNR}(I, \hat{I}) = 10 \cdot \log_{10} \left( \frac{(L_{\max})^2}{\text{MSE}(I, \hat{I})} \right),
  \]
  where $L_{\max} = 255$ and MSE denotes the mean squared error between pixel values.
  A higher PSNR indicates better reconstruction.

\textbf{Structural Similarity Index Measure (SSIM).}
  \[
  \text{SSIM}(I, \hat{I}) = \frac{(2\mu_I\mu_{\hat{I}} + c_1)(2\sigma_{I\hat{I}} + c_2)}{(\mu_I^2 + \mu_{\hat{I}}^2 + c_1)(\sigma_I^2 + \sigma_{\hat{I}}^2 + c_2)},
  \]
  where $\mu$, $\sigma^2$, and $\sigma_{I\hat{I}}$ denote means, variances, and covariances of local patches. SSIM captures perceptual similarity in terms of luminance, contrast, and structure.

\textbf{Learned Perceptual Image Patch Similarity (LPIPS).}
  LPIPS compares features from a pretrained deep network (e.g., AlexNet) between $\hat{I}$ and $I$, and correlates better with human judgment of perceptual similarity. Lower LPIPS indicates better quality.

These metrics are computed over four rendered views \(\{0^\circ,\,90^\circ,\,180^\circ,\,270^\circ\}\), under orthographic projection. We also report masked versions of these metrics that are evaluated only on occluded foreground regions, allowing more fine-grained assessment under challenging interaction scenarios.

\noindent \textbf{Occlusion-aware Metrics.}
To evaluate reconstruction quality under challenging visibility conditions, we compute occlusion-aware variants of image-based metrics and surface normal metrics by restricting the evaluation to regions occluded by another human instance.

Let $I^{(i)}$ and $\hat{I}^{(i)}$ denote the ground-truth and predicted images of instance $i \in \{0, 1\}$, and let $M^{(j)}$ be the binary mask of the other instance $j \neq i$. A pixel $(x,y)$ is considered occluded in instance $i$ if it belongs to $M^{(j)}$ and the corresponding pixel in $I^{(i)}$ is not background (i.e., not white):
\[
\resizebox{0.96\linewidth}{!}{$
\mathcal{O}^{(i)} = \left\{ (x,y) \,\middle|\, M^{(j)}(x,y) = 1 \;\wedge\; I^{(i)}(x,y) \neq background \right\}.
$}
\]
We then compute each metric by applying the occlusion mask $\mathcal{O}^{(i)}$ to both predicted and ground-truth images:
\begin{align*}
\text{Occ-PSNR}^{(i)} &= \text{PSNR}\left(\hat{I}^{(i)}|_{\mathcal{O}^{(i)}}, I^{(i)}|_{\mathcal{O}^{(i)}}\right), \\
\text{Occ-SSIM}^{(i)} &= \text{SSIM}\left(\hat{I}^{(i)}|_{\mathcal{O}^{(i)}}, I^{(i)}|_{\mathcal{O}^{(i)}}\right)
\end{align*}

For surface normal comparisons, let $N^{(i)}$ and $\hat{N}^{(i)}$ be the ground-truth and predicted normal maps of instance $i$. The occlusion-aware $L_2$ Normal Error is defined as:
\[
\resizebox{\linewidth}{!}{$
\text{Occ-L2-NormErr}^{(i)} = \frac{1}{|\mathcal{O}^{(i)}|} \sum_{(x,y) \in \mathcal{O}^{(i)}} \left\| \hat{N}^{(i)}(x,y) - N^{(i)}(x,y) \right\|_2^2.
$}
\]

All occlusion-aware metrics are averaged over both instances and across the four canonical viewpoints to provide a robust estimate of reconstruction performance in visually occluded regions.

\newpage
\section{Additional Results of 3D Multi-Human Reconstruction} \label{sec_sup:comparison}

\subsection{Qualitative Comparison Including Additional Baselines} \label{sec_sup:baseline}

\begin{figure*}[!t]
    \centering
    \includegraphics[width=0.7\textwidth]{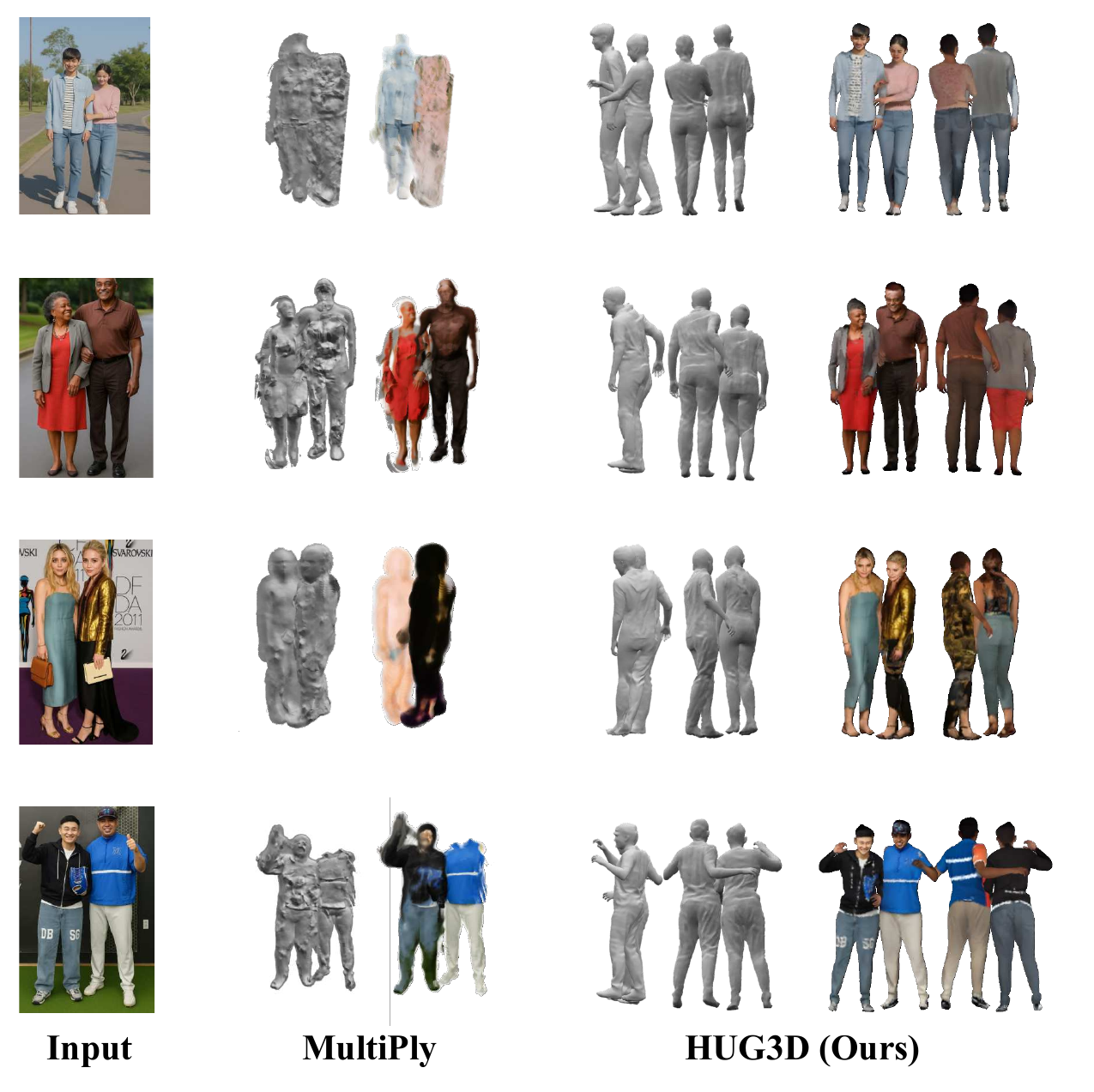}
    \vspace{-.5em}
\caption{Additional qualitative comparison on multi-human 3D reconstruction from a single in-the-wild image. HUG3D
outperforms baselines by correcting perspective distortion, preserving inter-human contact, and
hallucinating plausible textures under heavy occlusion.}
    \label{fig_supp_quali_wild1}
    \vspace{-.5em}
\end{figure*}

\begin{figure*}[!t]
    \centering
    \includegraphics[width=0.9\textwidth]{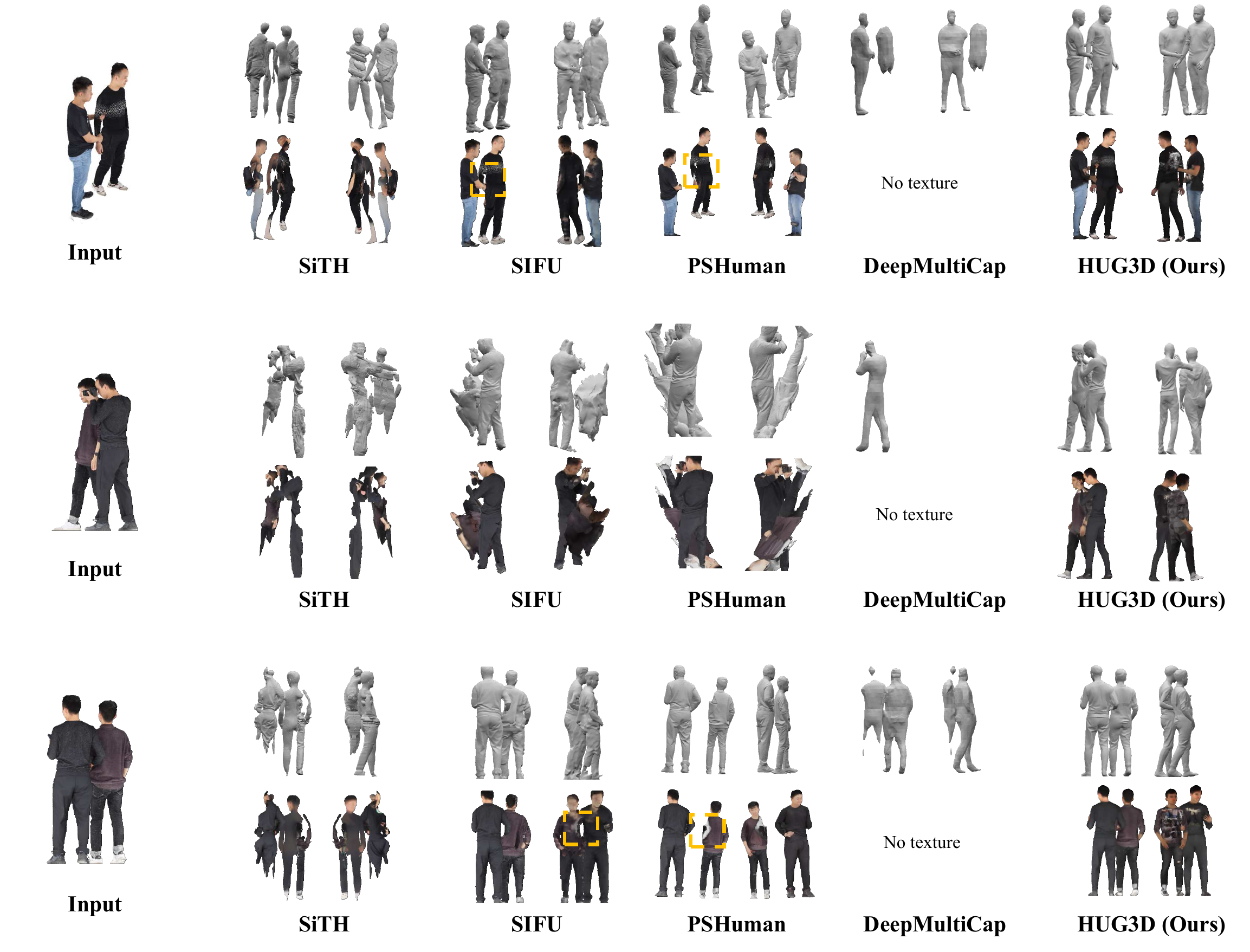}
    \vspace{-.5em}
\caption{Additional qualitative comparison on multi-human 3D reconstruction from a single image in the MultiHuman dataset. Yellow boxes highlight broken geometry, missing texture, and incorrect inter-human interactions. HUG3D
outperforms baselines by correcting perspective distortion, preserving inter-human contact, and
hallucinating plausible textures under heavy occlusion.}
    \label{sec_sup:fig_supp_quali_multihuman1}
    \vspace{-1em}
\end{figure*}

In addition to the baselines presented in the main paper, we include two additional baselines for comparison: DeepMultiCap~\citep{zheng2021deepmulticap}, a method designed for multi-human reconstruction from multi-view images, and Multiply~\citep{jiang2024multiply}, a method for multi-human reconstruction from videos.
Fig.~\ref{fig_supp_quali_wild1} presents additional qualitative comparisons on the in-the-wild images, while Fig.~\ref{sec_sup:fig_supp_quali_multihuman1} shows results on the MultiHuman dataset. In the in-the-wild setting, where SMPL-X predictions are used instead of ground-truth, our method continues to produce high-quality reconstructions, demonstrating robustness to SMPL-X estimation errors. Across all baselines, we observe common failure modes: incomplete geometry and missing textures in occluded regions, severe interpenetration or failure to preserve contact due to the lack of inter-person modeling, and inability to correct perspective distortion in images with complex viewpoints. In contrast, HUG3D consistently delivers robust multi-human reconstructions that preserve contact, correct geometric distortion, and hallucinate plausible textures even under severe occlusion.

\subsection{Results with Predicted SMPL-X} \label{sec_sup:predsmpl}
\begin{table}[h]
\centering
\vspace{-.5em}
\caption{End-to-end evaluation using predicted masks, SMPL-X parameters, and camera estimates from RoBUDDI. Despite operating on predicted inputs, HUG3D outperforms existing baselines.}
\vspace{-.5em}
\label{tab_pred_smpl}
\begin{adjustbox}{width=0.65\linewidth}
\begin{tabular}{cccc}
\toprule
Method & CD$\downarrow$ & PSNR$\uparrow$ & CP$\uparrow$ \\
\midrule
SIFU     & 26.268 & 10.037 & 0.006 \\
SiTH     & 24.607 & 9.332  & 0.012 \\
PSHuman  & 23.315 & 9.251  & 0.011 \\
\bottomrule
\end{tabular}
\end{adjustbox}
\end{table}
\vspace{-.5em}

In the main paper, we evaluate reconstruction performance using ground-truth SMPL-X parameters to decouple reconstruction quality from pose estimation errors, following common protocols in prior work (e.g., PSHuman~\citep{li2024pshuman}, ECON~\citep{xiu2023econ}). To further assess robustness under realistic conditions, we additionally report end-to-end results using predicted masks, SMPL-X parameters, and camera estimates obtained via RoBUDDI. As shown in Tab.~\ref{tab_pred_smpl}, HUG3D consistently outperforms all baselines even when operating on predicted inputs.

\subsection{Separate Results for Each Instance} \label{sec_sup:separate}

Table~\ref{tab_each_instance} shows per-instance comparisons of geometry and texture metrics. Our method consistently outperforms baselines across all measures, achieving better geometric accuracy (e.g., lowest CD, P2S, and Norm $L2$; highest NC and F-score) and texture quality (highest PSNR/SSIM, lowest LPIPS). This instance-level analysis further highlights the effectiveness of our unified framework in capturing both fine-grained geometry and high-quality appearance.

\subsection{Results Depending on Level of Interaction} \label{sec_sup:interaction}

Tables~\ref{tab_interaction_geo}--\ref{tab_occluded} compare results across two interaction levels: Closely interactive and Naturally interactive. Our method consistently outperforms others in geometry, texture, and occluded regions. It demonstrates superior geometric fidelity (e.g., CD, NC, F-score), texture quality (PSNR, SSIM, LPIPS), and robustness under occlusions, regardless of interaction level. These results highlight the resilience and generalizability of our approach under varying interaction conditions.

\subsection{Scalability to Larger Human Groups}
\label{sec_sup:scalability}

\begin{figure*}[!t]
    \centering
    \includegraphics[width=.85\textwidth]{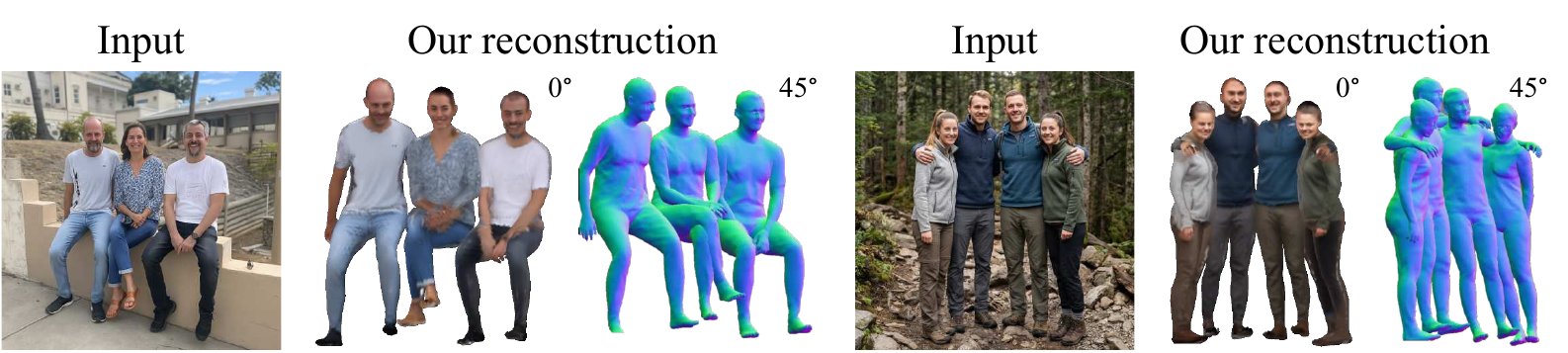}
    \vspace{-.5em}
    \caption{Qualitative results on scenes with three or more interacting humans. HUG3D reconstructs plausible multi-person interactions while preserving consistent geometry and contact relationships, demonstrating its scalability to larger human groups.}
    \label{fig_supp_3plus}
\end{figure*}

We further evaluate the scalability of HUG3D on scenes containing three or more interacting humans. Although the model is trained only on single-person and pair interactions, it generalizes to larger groups through the joint diffusion and reconstruction framework. As shown in Fig.~\ref{fig_supp_3plus}, HUG3D successfully reconstructs plausible multi-person interactions with three or more subjects while preserving consistent geometry and contact relationships. 

\subsection{Generalization to Out-of-Distribution Humans} \label{sec_sup:generalization}

\begin{figure*}[!t]
    \centering
    \includegraphics[width=1.\textwidth]{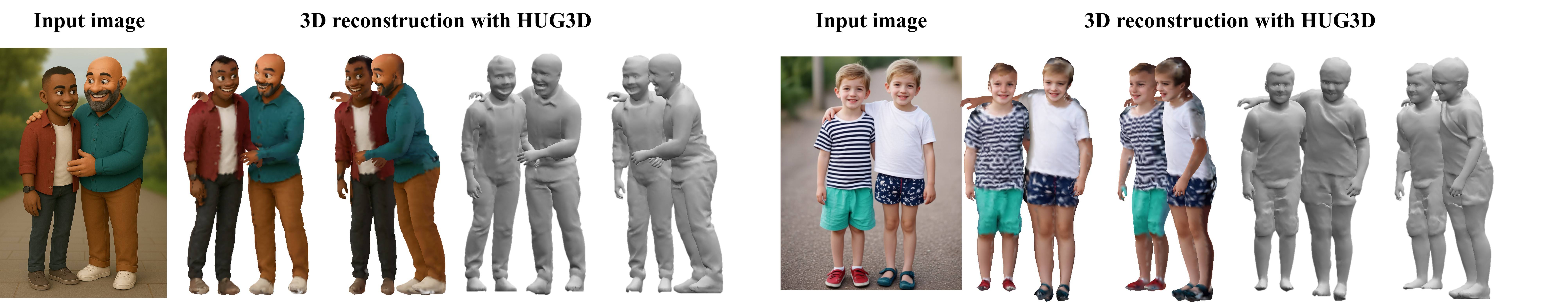}
    \vspace{-1.5em}
    \caption{Qualitative results demonstrating HUG3D's generalization capability to novel human types, including stylized 3D characters and children. Despite domain differences, our method produces structurally plausible and semantically consistent outputs.}
    \label{fig_supp_generalization}
\end{figure*}

To assess the robustness of our method, we tested HUG3D on novel human inputs, including stylized 3D characters and children—categories not present during training. As shown in Fig.~\ref{fig_supp_generalization}, while minor mismatches in body proportions may occur due to distribution shifts, our model still generates geometrically plausible and semantically coherent outputs. These results highlight the strong generalization ability of HUG3D, even in challenging and unseen scenarios.

\subsection{Results from Multiple Views} \label{sec_sup:novel_view}

\noindent Figs.~\ref{fig_supp_novel_view_normal} and~\ref{fig_supp_novel_view_rgb} show qualitative renderings of our reconstructed textured 3D mesh from a broad set of viewpoints. We visualize both training views (with gray backgrounds) and novel views (with white backgrounds), sampled across varying camera positions: elevations of \{-45°, 0°, 45°\} and azimuths of \{0°, 45°, 90°, 135°, 180°, 225°, 270°, 315°\}. These results demonstrate the model’s strong generalization capability to unseen perspectives for both normal maps and RGB images.

\subsection{Robustness to Intermediate Errors} \label{sec_sup:robustness}

\begin{figure*}[!t]
    \centering
    \includegraphics[width=.6\textwidth]{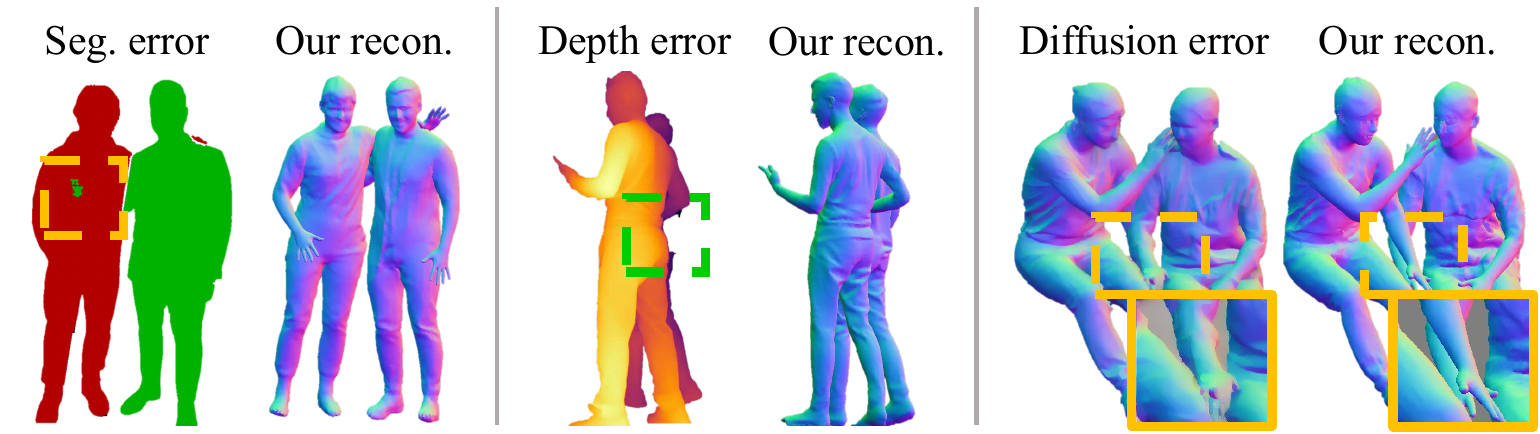}
    \vspace{-0.5em}
    \caption{Robustness to errors in intermediate predictions. Despite inaccuracies in segmentation, depth estimation, or diffusion outputs, HUG3D maintains plausible multi-human reconstructions with consistent geometry and interaction relationships.}
    \label{fig_supp_robustness}
\end{figure*}
\begin{figure*}[!t]
    \centering
    \includegraphics[width=.6\textwidth]{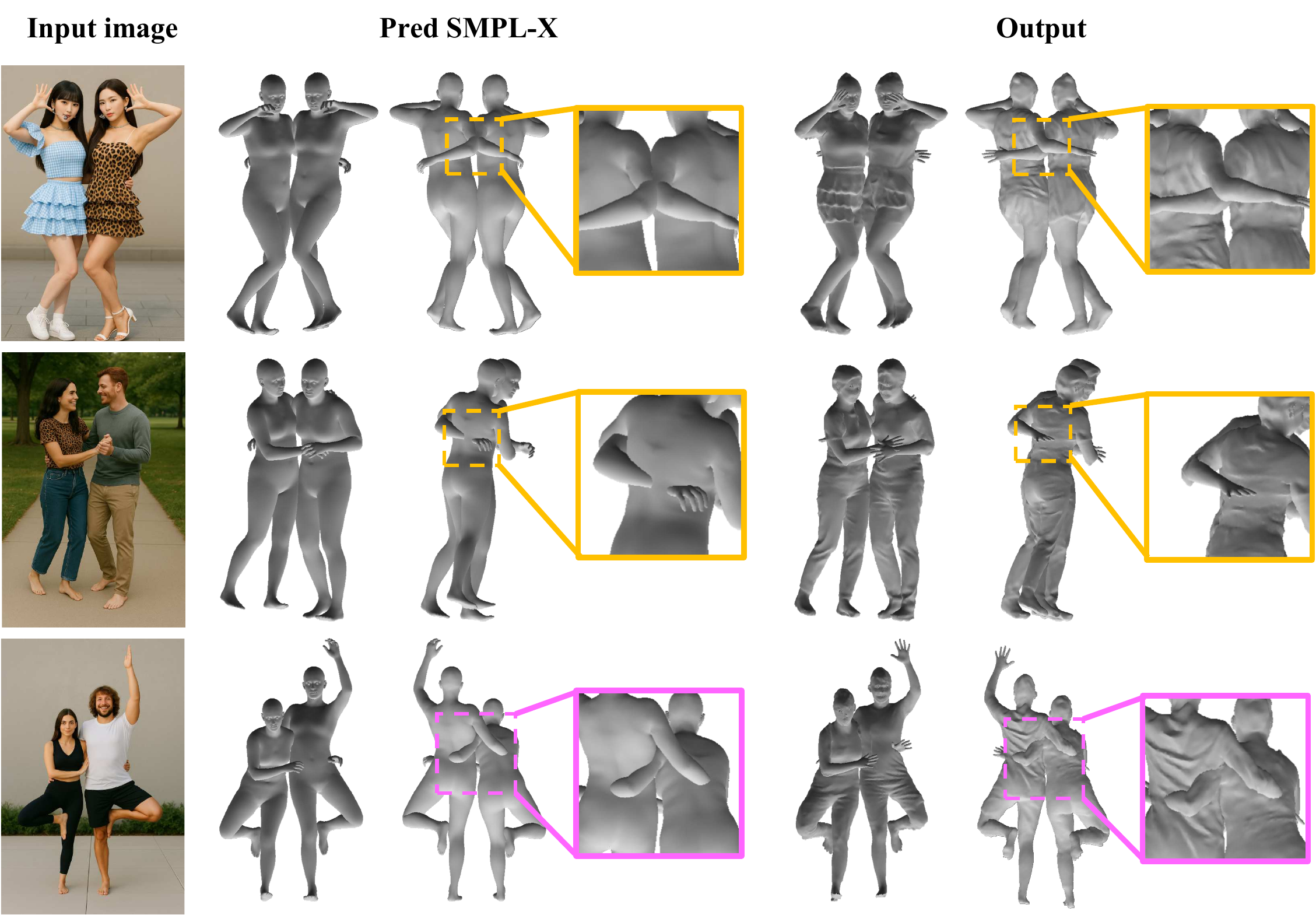}
    \vspace{-0.5em}
    \caption{Robustness to SMPL-X estimation errors. The top two rows demonstrate that our method reduces interpenetration artifacts even with inaccurate SMPL-X estimates. The last row shows a failure case arising from severely corrupted SMPL-X initialization.}
    \label{fig_supp_smpl_robustness}
    \vspace{-1em}
\end{figure*}
\begin{figure*}[!t]
    \centering
    \includegraphics[width=0.8\textwidth]{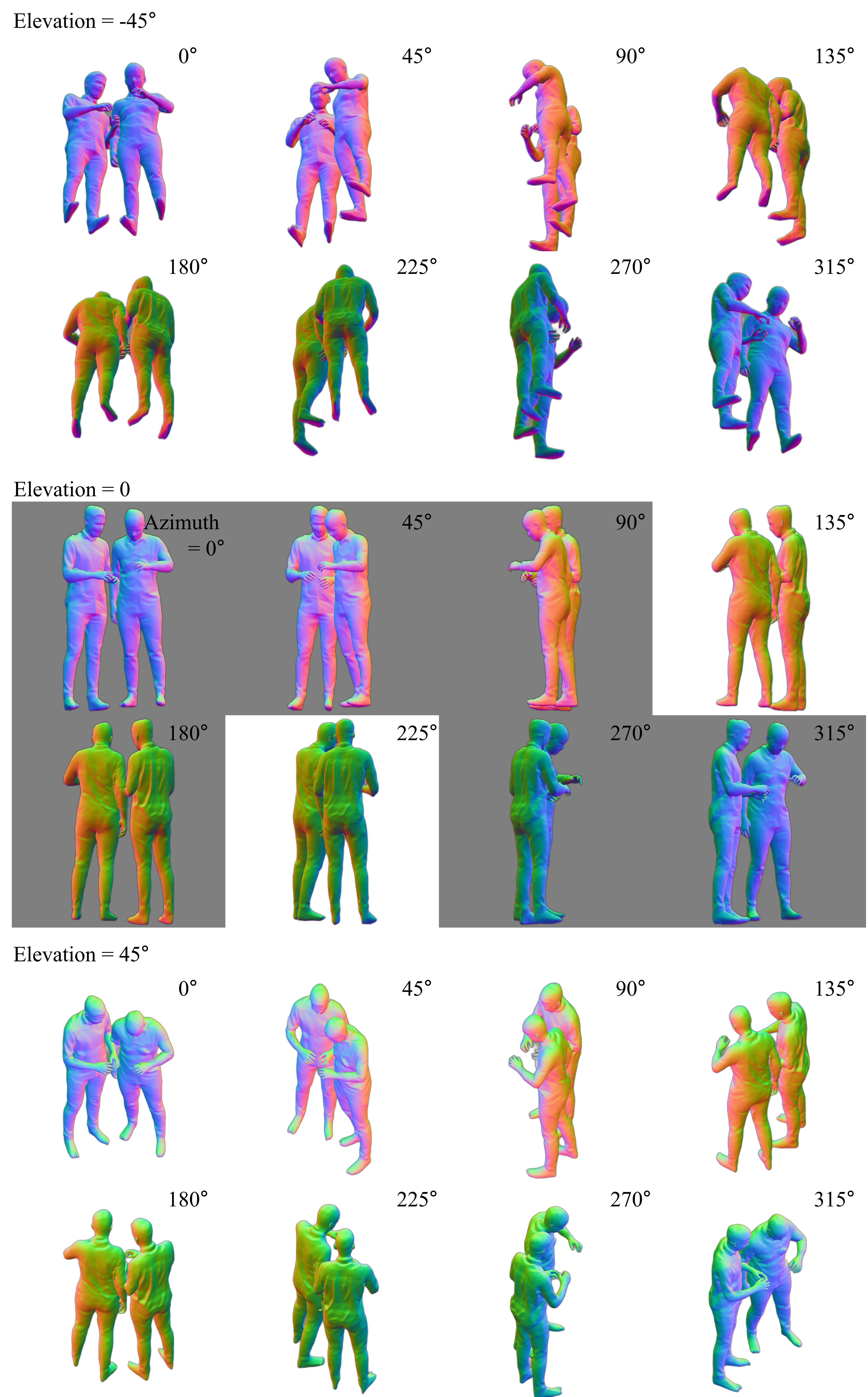}
    \vspace{-0.5em}
    \caption{Normal maps rendered from multiple viewpoints of our reconstructed textured 3D mesh, including both training views (gray background) and novel views (white background).}
    \label{fig_supp_novel_view_normal}
    \vspace{-1em}
\end{figure*}
\begin{figure*}[!t]
    \centering
    \includegraphics[width=0.8\textwidth]{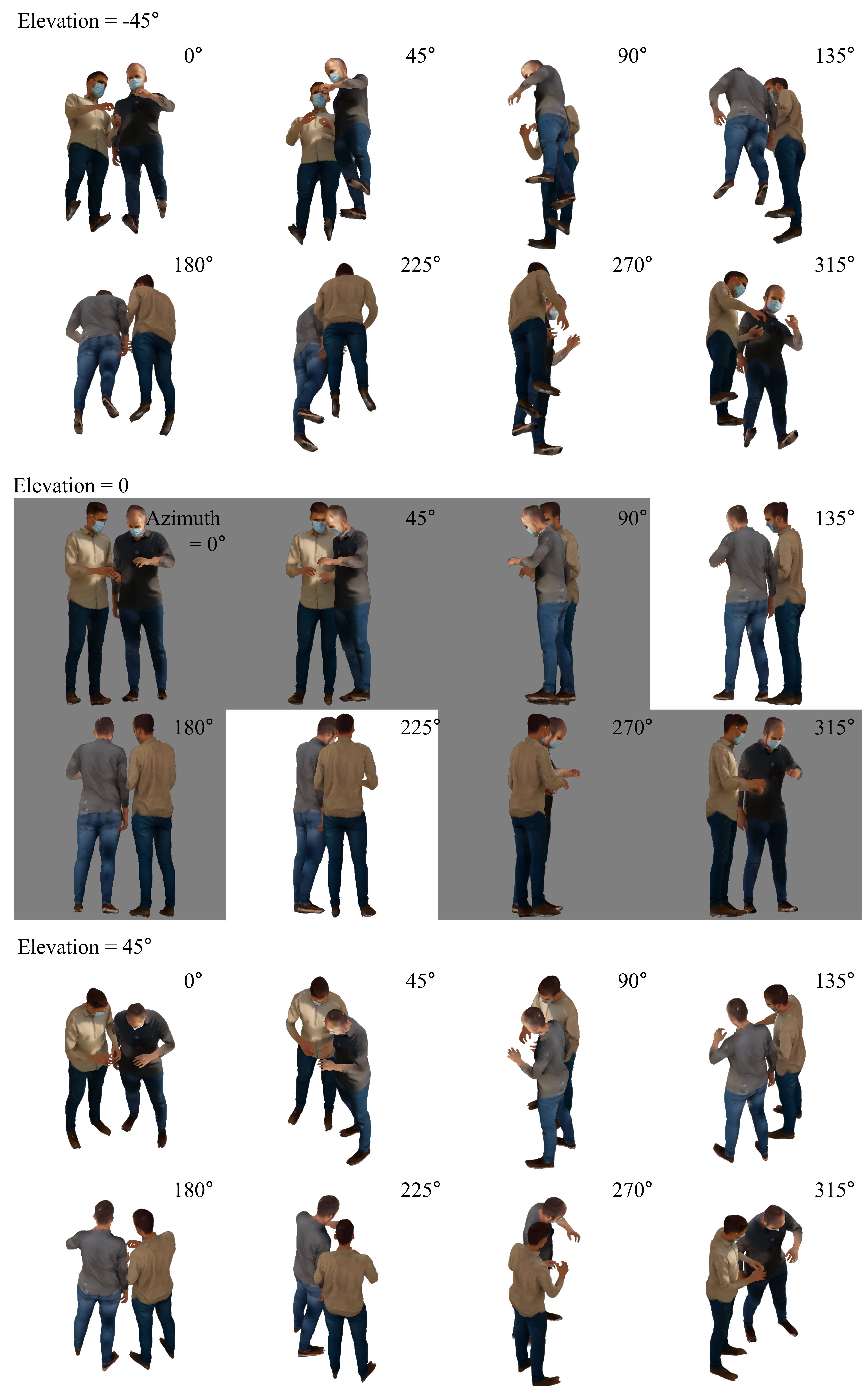}
    \vspace{-0.5em}
    \caption{RGB renderings from multiple viewpoints of our reconstructed textured 3D mesh, including both training views (gray background) and novel views (white background).}
    \label{fig_supp_novel_view_rgb}
    \vspace{-1em}
\end{figure*}

As shown in Fig.~\ref{fig_supp_robustness} and Fig.~\ref{fig_supp_smpl_robustness}, our approach demonstrates robustness to inaccuracies in intermediate stages such as segmentation, depth estimation, diffusion predictions, and SMPL-X estimation. This robustness is enabled by our multi-view diffusion prior and physics-based, interaction-aware geometry reconstruction.

While our approach remains stable under moderate errors (the top two rows of Fig.~\ref{fig_supp_smpl_robustness}), reconstruction quality degrades when the SMPL-X initialization is heavily corrupted, as shown in the last row of Fig.~\ref{fig_supp_smpl_robustness}.

\subsection{Videos}
\label{sec_sup:videos}
We provide an accompanying supplementary video that better visualizes the key advantages of our method, HUG3D. The video highlights that HUG3D produces physically plausible, high-fidelity 3D reconstructions of interacting people from a single image. The video is available on our project page: 
\href{https://jongheean11.github.io/HUG3D_project}{\textcolor{magenta}{jongheean11.github.io/HUG3D\_project}}.

\clearpage

\begin{table*}[h]\centering
\caption{Quantitative comparison of geometry and texture for each instance}
\label{tab_each_instance}
\scriptsize
\centering
\begin{adjustbox}{width=0.9\linewidth}
\begin{tabular}{c|cccccc|ccc}
\toprule
Method & CD ↓ & P2S ↓ & NC ↑ & F-score ↑ & bbox-IoU ↑ & Norm $L2$ ↓  
& PSNR ↑ & SSIM ↑ & LPIPS ↓  \\
\midrule
SIFU         & 6.367 & 2.292 & 0.753 & 30.203 & 0.659 & 0.018 & 17.222 & 0.882 & 0.127 \\
SiTH         & 9.642 & 3.166 & 0.712 & 21.740 & 0.541 & 0.024 & 16.090 & 0.881 & 0.143 \\
PSHuman      & 16.876 & 6.384 & 0.614 & 9.561  & 0.402 & 0.039 & 13.720 & 0.857 & 0.188 \\
DeepMultiCap & 13.314 & 2.952 & 0.754 & 18.898 & 0.442 & 0.026 & 15.25 & 0.880 & 0.161 \\
\textbf{Ours} & \textbf{3.531} & \textbf{1.719} & \textbf{0.816} & \textbf{42.946} & \textbf{0.801} & \textbf{0.012} & \textbf{18.659} & \textbf{0.894} & \textbf{0.102} \\
\bottomrule
\end{tabular}
\end{adjustbox}
\vspace{-1em}
\end{table*}

\begin{table*}[t]
\caption{Quantitative comparison of geometry depending on level of interaction.}
\label{tab_interaction_geo}
\scriptsize
\centering
\begin{adjustbox}{width=0.9\linewidth}
\begin{tabular}{ccccccccc}
\toprule
Interaction & Method & CD ↓ & P2S ↓ & NC ↑ & F-score ↑ & bbox-IoU ↑ & Norm $L2$ ↓ & CP ↑  \\
\midrule
\multirow{5}{*}{Closely} 
& SIFU         & 7.267 & 2.750 & 0.724 & 24.335 & 0.757 & 0.033 & 0.117 \\
& SiTH         & 10.908 & 3.491 & 0.697 & 19.216 & 0.694 & 0.044 & 0.281 \\
& PSHuman      & 14.920 & 5.518 & 0.616 & 10.572 & 0.631 & 0.065 & 0.049 \\
& DeepMultiCap & 9.6697 & 2.745 & 0.764 & 20.471 & 0.606 & 0.039 & 0.123 \\
& \textbf{Ours} & \textbf{4.315} & \textbf{2.121} & \textbf{0.811} & \textbf{37.243} & \textbf{0.838} & \textbf{0.022} & \textbf{0.326} \\
\midrule
\multirow{5}{*}{Natural} 
& SIFU         & 4.895 & 2.069 & 0.768 & 31.510 & 0.788 & 0.026 & 0.076 \\
& SiTH         & 8.486 & 3.044 & 0.714 & 21.877 & 0.715 & 0.038 & 0.068 \\
& PSHuman      & 15.884 & 6.350 & 0.617 & 9.370 & 0.671 & 0.070 & 0.017 \\
& DeepMultiCap & 17.081 & 2.463 & 0.741 & 17.018 & 0.470 & 0.052 & 0.064 \\
& \textbf{Ours} & \textbf{3.340} & \textbf{1.585} & \textbf{0.816} & \textbf{42.957} & \textbf{0.849} & \textbf{0.018} & \textbf{0.184} \\
\bottomrule
\end{tabular}
\end{adjustbox}
\vspace{-1em}
\end{table*}

\begin{table*}[t]\centering
\scriptsize
    \begin{minipage}[t]{0.48\linewidth}
        \centering
        \caption{{Quantitative comparison of texture depending on level of interaction.}}
        \label{tab_texture}
        \begin{adjustbox}{width=0.95\linewidth}
        \begin{tabular}{ccccc}
        \toprule
        Interaction & Method & PSNR ↑ & SSIM ↑ & LPIPS ↓ \\
        \midrule
        \multirow{4}{*}{Closely} 
        & SIFU         & 14.369 & 0.781 & 0.223 \\
        & SiTH         & 13.683 & 0.785 & 0.243 \\
        & PSHuman      & 11.722 & 0.747 & 0.293 \\
        & \textbf{Ours}& \textbf{16.454} & \textbf{0.805} & \textbf{0.179} \\
        \midrule
        \multirow{4}{*}{Natural}
        & SIFU         & 15.586 & 0.799 & 0.192 \\
        & SiTH         & 13.851 & 0.790 & 0.228 \\
        & PSHuman      & 11.278 & 0.740 & 0.309 \\
        & \textbf{Ours}& \textbf{16.741} & \textbf{0.818} & \textbf{0.166} \\
        \bottomrule
        \end{tabular}
        \end{adjustbox}
    \end{minipage}
    \hfill
    \begin{minipage}[t]{0.45\linewidth}
        \centering
        \caption{Quantitative comparison within occluded regions depending on level of interaction.}
        \label{tab_occluded}
        \begin{adjustbox}{width=\linewidth}
        \begin{tabular}{ccccc}
        \toprule
        Interaction & Method & Norm $L2$ ↓ & PSNR ↑ & SSIM ↑ \\
        \midrule
        \multirow{4}{*}{Closely}
        & SIFU         & 0.223 & 5.745 & 0.569 \\
        & SiTH         & 0.218 & 5.900 & 0.551 \\
        & PSHuman      & 0.258 & 4.344 & 0.529 \\
        & DeepMultiCap & 0.219 & - & - \\
        & \textbf{Ours}& \textbf{0.153} & \textbf{8.082} & \textbf{0.610} \\
        \midrule
        \multirow{4}{*}{Natural}
        & SIFU         & 0.184 & 6.359 & 0.554 \\
        & SiTH         & 0.187 & 6.557 & 0.532 \\
        & PSHuman      & 0.249 & 4.757 & 0.501 \\
        & DeepMultiCap & 0.216 & -     & -     \\
        & \textbf{Ours}& \textbf{0.138} & \textbf{8.358} & \textbf{0.599} \\
        \bottomrule
        \end{tabular}
        \end{adjustbox}
    \end{minipage}
\end{table*}

\clearpage

\newpage
\section{Results from Each Component} \label{sec_sup:components}
\begin{figure*}[!t]
    \centering
    \includegraphics[width=0.8\textwidth]{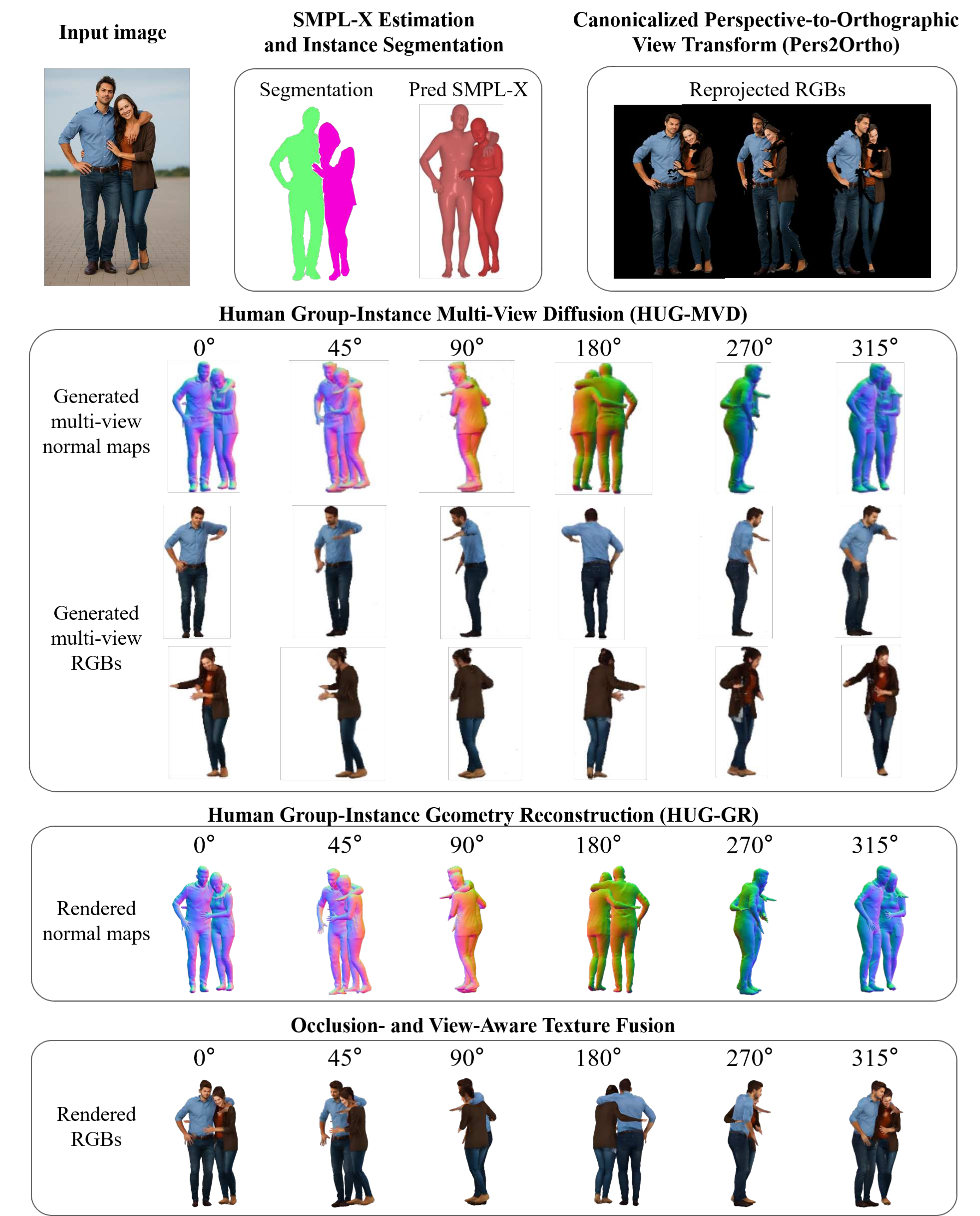}
    \vspace{-0.5em}
    \caption{Example of results from each component of our HUG3D. }
    \label{fig_supp_each}
    \vspace{-1em}
\end{figure*}

Fig.~\ref{fig_supp_each} presents qualitative outputs from each stage of our framework.  
(1) \textit{SMPL-X Estimation and Instance Segmentation} produce parametric body models and segmentation masks.  
(2) \textit{Canonical Perspective-to-Orthographic View Transformation (Pers2Ortho)} enables reprojection of RGB images to a shared canonical view.  
(3) \textit{Human Group-Instance Multi-View Diffusion (HUG-MVD)} generates multi-view consistent RGB and normal maps.  
(4) \textit{Human Group-Instance Geometry Reconstruction (HUG-GR)} reconstructs accurate 3D meshes of multiple human subjects.  
(5) \textit{Occlusion- and View-Aware Texture Fusion} synthesizes high-quality textured meshes by integrating multi-view information while handling occlusions and viewpoint variations.

\section{Additional Ablation Studies and Analysis} \label{sec_sup:ablation}

\subsection{Robust SMPL-X Estimation (RoBUDDI)} \label{sec_sup:abl_segmentation}

\begin{table}[!h]
  \centering
  \scriptsize
  \caption{Quantitative comparison of SMPL-X fitting accuracy on the MultiHuman dataset~\citep{zheng2021deepmulticap}. Metrics include mean per-joint position error (MPJPE), its Procrustes-aligned variant (PA-MPJPE), and mean vertex error (MVE).}
  \label{tab:smplx}
  \begin{adjustbox}{width=0.9\linewidth}
    \begin{tabular}{cccc}
      \toprule
      Method & MPJPE ↓ & PA-MPJPE ↓ & MVE ↓ \\
      \midrule
      BEV    & 13.178 & 12.704 & 10.570 \\
      BUDDI  & 13.162 & 12.695 & 10.591 \\
      \textbf{Ours (RoBUDDI)} & \textbf{13.139} & \textbf{12.673} & \textbf{10.566} \\
      \bottomrule
    \end{tabular}
  \end{adjustbox}
\end{table}

We evaluate our proposed RoBUDDI on the MultiHuman dataset~\citep{zheng2021deepmulticap} and compare it against BEV~\citep{sun2022putting} and BUDDI~\citep{muller2024generative}. As shown in Tab.~\ref{tab:smplx}, RoBUDDI achieves lower MPJPE, PA-MPJPE, and MVE, demonstrating superior accuracy in 3D pose and shape estimation. 

In addition to quantitative improvements, our method shows qualitative benefits as illustrated in Fig.~\ref{fig_supp_ablation_smplx}. While BUDDI suffers from interpenetration artifacts between closely interacting subjects (yellow arrows), our RoBUDDI, enhanced with interpenetration and visibility-aware losses, yields more physically plausible and realistic 3D reconstructions.

\subsection{Canonical Perspective-to-Orthographic View Transform (Pers2Ortho)} \label{sec_sup:abl_pers2ortho}

\noindent \textbf{Depth Edge-Aware Uncertain Point Filtering.}  
As shown in Fig.~\ref{fig_supp_method_p2o_uncertain_point_filtering}, removing uncertain points helps reduce jagged contours and ghosting artifacts near object boundaries after reprojection.

\noindent \textbf{Depth-Aware Visible Point Selection.}  
As shown in Fig.~\ref{fig_supp_method_p2o_visiblity_point_selection}, this strategy filters out occluded or background points, retaining only those in front of the mesh surface and visible from the target camera view.

\subsection{Human Group-Instance Multi-View Diffusion (HUG-MVD)} \label{sec_sup:abl_hugmvd}

\noindent \textbf{Perspective multi-view diffusion vs. Orthographic multi-view diffusion.} We compare the results of multi-view diffusion models trained directly on perspective images with those trained on orthographic images in Fig.~\ref{fig_supp_ablation_mvd_pers} with the same training settings. Due to the limited amount of ground-truth group data, the geometric complexity of group scenes, and the constrained capacity of the base diffusion model, models trained on perspective images often fail to produce plausible outputs (Fig.~\ref{fig_supp_ablation_mvd_pers}(a)). These failures manifest as artifacts such as twisted limbs and mixed clothing textures. In contrast, our multi-view diffusion model trained on orthographic images performs significantly better under limited data settings (Fig.~\ref{fig_supp_ablation_mvd_pers}(b)). This highlights the effectiveness of our strategy, which first transforms perspective images into orthographic views and then applies multi-view diffusion in a canonical space.

\noindent \textbf{Comparison with state-of-the-art diffusion-based inpainting.}  
Fig.~\ref{fig_supp_ablation_inp} shows a detailed comparison between current state-of-the-art diffusion-based inpainting methods~\citep{rombach2022high, flux2024} and our HUG-MVD, which achieves multi-view consistent inpainting and generation. Unlike existing inpainting approaches, HUG-MVD (1) produces multi-view consistent results, (2) maintains pose consistency without anatomical errors, and (3) additionally performs normal map inpainting—representing a significant advancement. 

While several prior works address multi-view consistent inpainting, their tasks differ and are less suitable for human occlusion completion. For instance, MVInpainter~\citep{cao2024mvinpainter} requires a consistent multi-view background and an inpainted object visible in the first view, and Instant3Dit~\citep{barda2024instant3dit} demands a 3D object as input. In contrast, our method only requires a single occluded human image as input, which is a more challenging and realistic scenario.

\noindent \textbf{Training MVD with occlusion-aware masks and freeform masks.}  
In Fig.~\ref{fig_supp_ablation_mask}, we evaluate our inpainting training strategy for HUG-MVD by comparing mask generation schemes. Since freeform masks do not reflect real world occlusion patterns, it biases the model towards learning mask artifacts and producing visibly unnatural inpainted regions. In contrast, our method enables more natural, artifact-free inpainting.

\subsection{Occlusion- and View-Aware Texture Fusion} \label{sec_sup:abl_texture}

\noindent \textbf{View-Aware Face Restoration.}  
As shown in Fig.~\ref{fig_supp_method_tex_face_restroation}, our method effectively refines facial regions captured from extreme angles or under occlusion, such as back views, which often exhibit degraded appearances.

\noindent \textbf{Occlusion-Aware Blending.}  
As illustrated in Fig.~\ref{fig_supp_method_tex_occ_aware_blending}, our method effectively prevents ghosting and bleeding artifacts near occlusion boundaries.

\begin{table*}[t]\centering
\caption{Comparison of inference time and memory usage across baseline methods. Our method achieves superior performance while operating within the range of existing methods.}
\label{tab:efficiency_comparison}
\begin{adjustbox}{width=0.8\linewidth}
\begin{tabular}{@{}cccccccc@{}}
\toprule
 Metric & SIFU & ECON & SiTH & PSHuman & DeepMultiCap & MultiPly & HUG3D (Ours) \\
\midrule
Elapsed Time (s) & 333.79 & 80.09 & 148.01 & 128.47 & 42.35 & 27907.67 & 216.32 \\
Peak VRAM (GB)   & 7.31  & 5.44  & 17.79  & 32.12  & 1.37  & 5.75     & 34.76  \\
\bottomrule
\end{tabular}
\end{adjustbox}
\vspace{-.5em}
\end{table*}

\begin{table*}[t]
\centering
\caption{Average elapsed time and peak VRAM usage for each pipeline stage.}
\label{tab:efficiency_ours}
\begin{adjustbox}{width=0.55\linewidth}
\begin{tabular}{@{}ccccc@{}}
\toprule
Metric & Pers2Ortho & HUG-MVD & HUG-GR & Texture Fusion \\
\midrule
Elapsed Time (s) & 16.20 & 60.16 & 125.46 & 14.49 \\
Peak VRAM (GB)   & 14.40  & 34.76 & 7.58  & 4.95  \\
\bottomrule
\end{tabular}
\end{adjustbox}
\end{table*}

\begin{table*}[h!]
\centering
\caption{Wilcoxon signed-rank test results (p-values) across all evaluation metrics, confirming statistically significant improvements of our method over baselines.}
\label{tab:significance}
\resizebox{\textwidth}{!}{
\begin{tabular}{cccccccccccccc}
\toprule
Method & CD & P2S & NC & F-score & bbox-IoU & Norm $L2$ & CP & PSNR & SSIM & LPIPS & Occ.Norm $L2$ & Occ.PSNR & Occ.SSIM \\
\midrule
SIFU          & 1.8e-09 & 3.3e-08 & 3.9e-14 & 3.9e-10 & 1.6e-07 & 2.0e-39 & 1.1e-05 & 3.8e-27 & 1.7e-38 & 5.1e-37 & 5.7e-10 & 5.9e-10 & 1.9e-11 \\
SiTH          & 3.6e-14 & 3.8e-14 & 3.6e-14 & 5.8e-14 & 4.6e-13 & 1.4e-51 & 8.2e-04 & 1.1e-48 & 2.2e-38 & 1.5e-51 & 1.3e-14 & 4.2e-14 & 1.4e-13 \\
PSHuman       & 3.6e-14 & 3.6e-14 & 3.6e-14 & 5.2e-14 & 2.7e-13 & 1.4e-51 & 1.2e-08 & 1.4e-51 & 1.7e-51 & 1.4e-51 & 1.9e-18 & 1.4e-18 & 5.2e-17 \\
DeepMultiCap  & 7.8e-13 & 1.1e-08 & 2.6e-12 & 1.3e-13 & 1.4e-13 & 1.4e-46 & 5.3e-06 & 3.5e-50 & 1.2e-27 & 5.2e-47 & 5.9e-11 & 1.1e-14 & 1.5e-14 \\
\bottomrule
\end{tabular}}
\end{table*}

\subsection{Interaction-Aware Modeling}
\label{sec_sup:abl_interaction}
\begin{table}[h!]
  \centering
  \vspace{-.5em}
  \caption{Ablation study on interaction-aware modeling. Removing either component degrades performance, confirming that both stages are essential for accurate multi-human interaction reconstruction.}
  \label{tab:interaction}
  \vspace{-.5em}
  \begin{adjustbox}{width=0.7\linewidth}
    \begin{tabular}{cccc}
    \toprule
    Method & CD$\downarrow$ & P2S$\downarrow$ & NC$\uparrow$ \\
    \midrule
    w/o HUG-MVD & 10.810  & 6.993 & 0.529 \\
    w/o HUG-GR  & 11.746  & 7.920 & 0.524 \\
    \textbf{Ours} & \textbf{10.627} & \textbf{6.877} & \textbf{0.537} \\
    \bottomrule
    \end{tabular}
  \end{adjustbox}
\end{table}
\vspace{-.5em}

Tab.~\ref{tab:interaction} reports metrics computed specifically within contact regions to isolate the contribution of each stage. \textit{w/o HUG-MVD} is trained solely on single-human data and removes the joint multi-human diffusion inference, eliminating the generative priors required for modeling interactions. \textit{w/o HUG-GR} removes the interaction-aware geometric reconstruction stage by disabling group-normal and physics-based losses, which reduces physical plausibility and contact precision. These results confirm that interaction-aware modeling in both stages is essential.

\subsection{Efficiency Analysis} \label{sec_sup:abl_efficiency}

We provide a comparison of inference time and memory usage across baseline methods in Tab.~\ref{tab:efficiency_comparison}. While our end-to-end inference time of 216 seconds per image is within the range of existing methods, this represents a reasonable trade-off, as our approach substantially outperforms them in reconstruction fidelity and physical plausibility. The primary runtime overhead arises from contact-mask computation in HUG-GR, which is necessary for accurate interaction modeling. Disabling it reduces runtime by 58.6\% but degrades interaction realism (Fig. 6(b)). Moreover, the runtime scales linearly with the number of subjects, rather than exponentially.

We also measured the elapsed time and peak VRAM usage for each stage in our proposed method as shown in Tab.~\ref{tab:efficiency_ours} with NVDIA A100. We observed that the most significant bottlenecks were identified to be HUG-GR (time-wise) and HUG-MVD (peak VRAM-wise), with each stage consuming 125.46 seconds and 34.76 GB of VRAM respectively. 

\subsection{Statistical Significance Analysis} \label{sec_sup:abl_stat_sign}

We conducted Wilcoxon signed-rank tests~\citep{wilcoxon1945individual} to assess statistical significance across all metrics in 
Tabs. 1, 2, 3.
As shown in Tab. ~\ref{tab:significance}, the p-values confirm that our method consistently outperforms all baselines with statistically significant (p value < 0.001) differences across geometry, texture, and occlusion handling metrics.

\

\begin{figure*}[!t]
    \centering
    \includegraphics[width=1.0\textwidth]{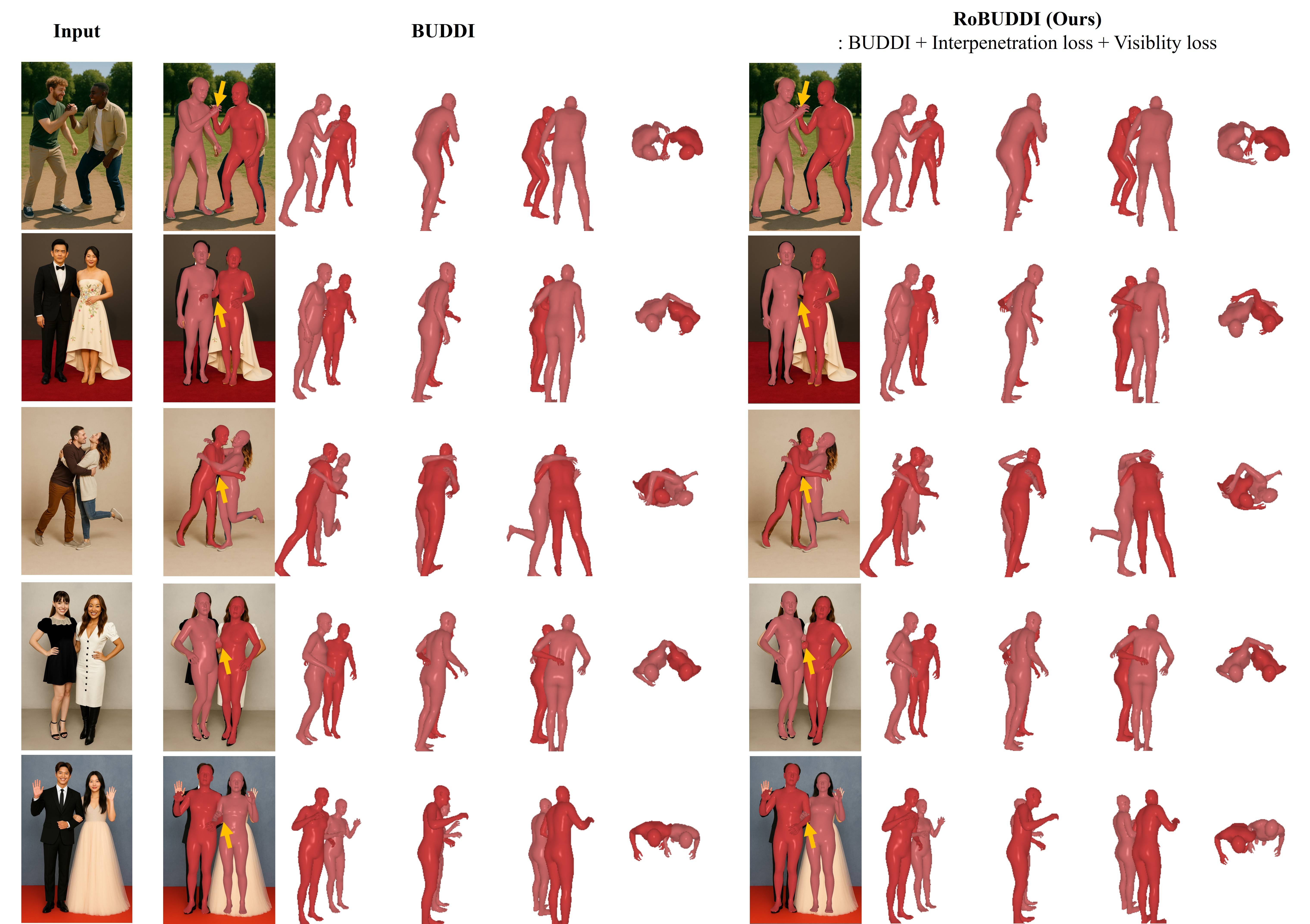}
    \vspace{-0.5em}
    \caption{Qualitative comparison of our SMPL-X fitting (RoBUDDI) against BUDDI~\citep{muller2024generative}. BUDDI exhibits visible interpenetrations between interacting subjects (yellow arrows), whereas our RoBUDDI produces more physically plausible results.}
    \label{fig_supp_ablation_smplx}
    \vspace{-1em}
\end{figure*}

\begin{figure*}[!t]
    \centering
    \includegraphics[width=.9\textwidth]{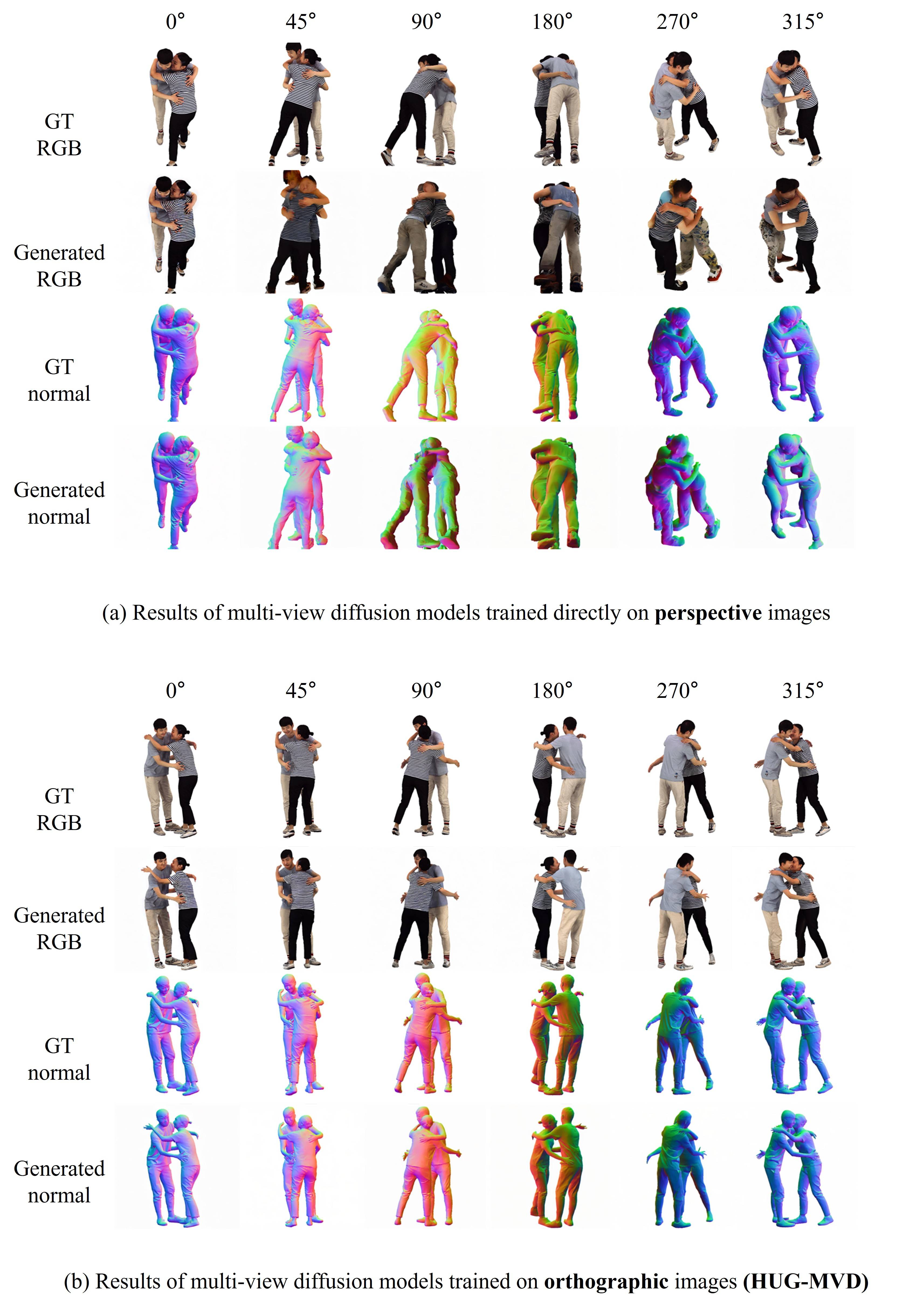}
    \vspace{-0.5em}
    \caption{Comparison of results from multi-view diffusion models trained directly on perspective images vs. models trained on orthographic images.}
    \label{fig_supp_ablation_mvd_pers}
    \vspace{-1em}
\end{figure*}

\begin{figure*}[!t]
    \centering
    \includegraphics[width=0.9\textwidth]{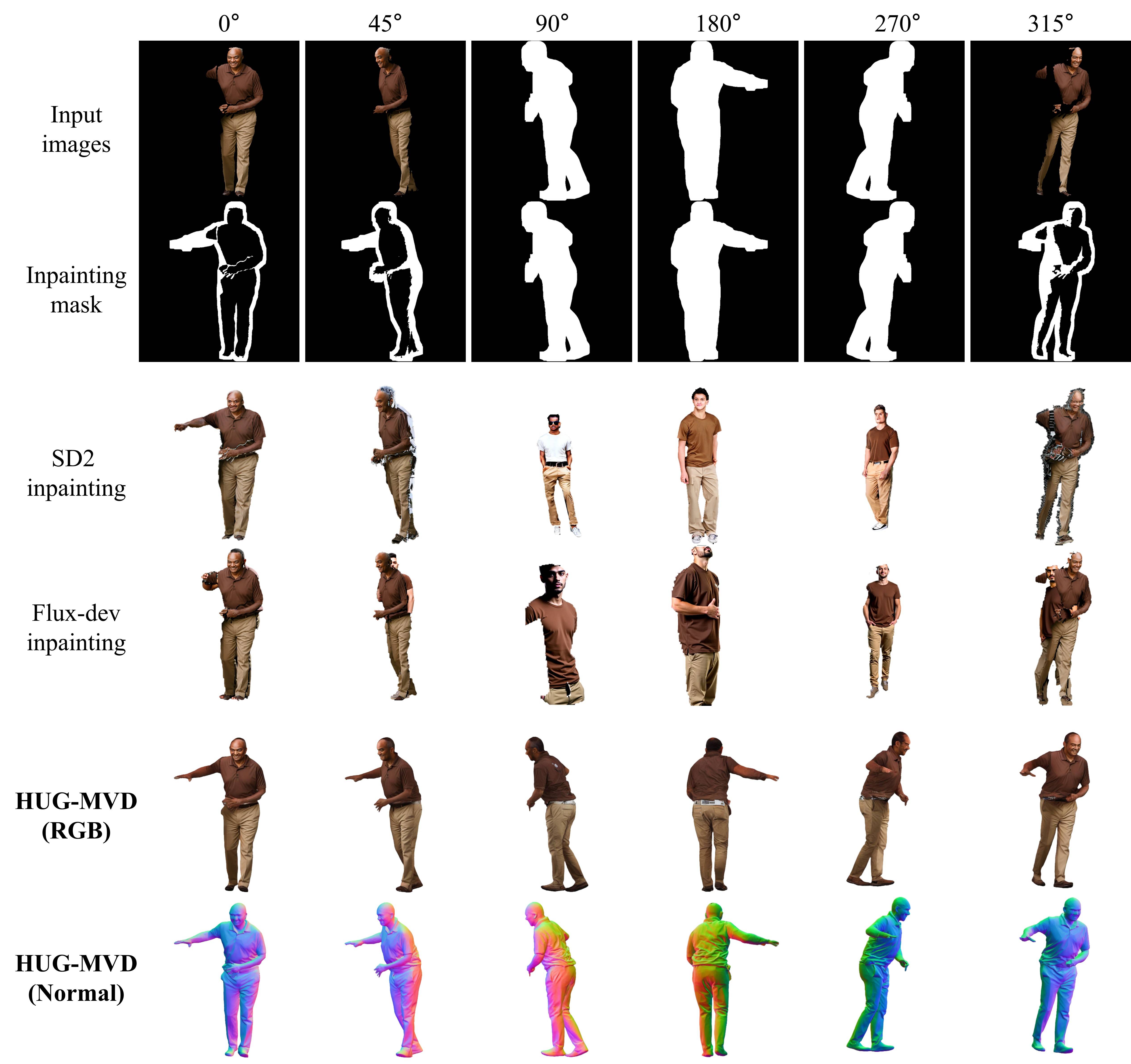}
    \vspace{-0.5em}
    \caption{Comparison between state-of-the-art diffusion-based inpainting methods and our HUG-MVD framework for multi-view consistent inpainting and generation.}
    \label{fig_supp_ablation_inp}
    \vspace{-1em}
\end{figure*}

\begin{figure*}[!t]
    \centering
    \includegraphics[width=0.95\textwidth]{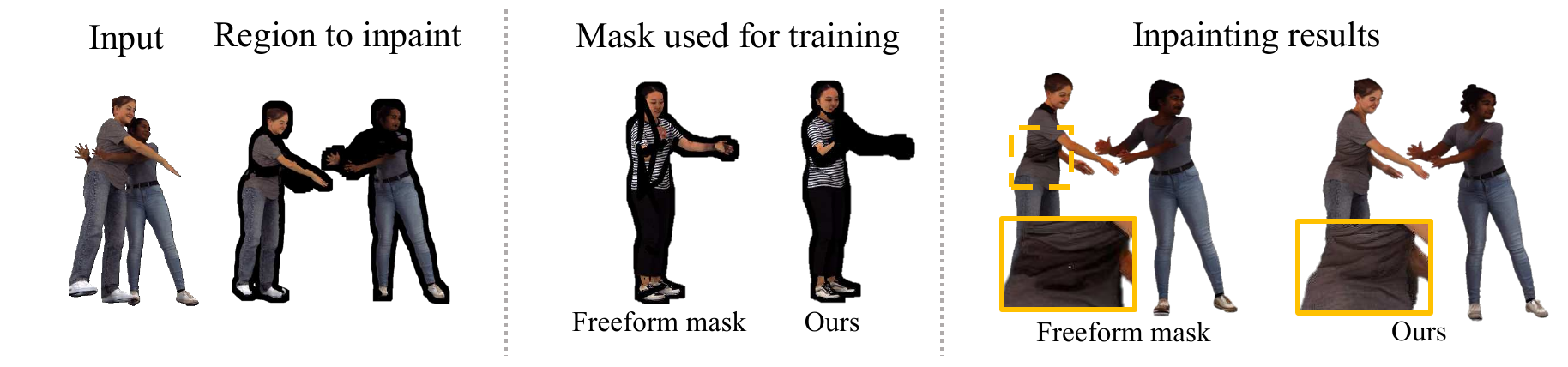}
    \vspace{-0.5em}
    \caption{Ablation on mask types for training HUG-MVD. Our occlusion-simulated masks enhance inpainting compared to freeform masks. 
}
    \label{fig_supp_ablation_mask}
    \vspace{-1em}
\end{figure*}

\clearpage
\balance
\section{ Licenses for Existing Assets} \label{sec_sup:licenses}


\subsection{Libraries} \label{sec_sup:libs}
The libraries used in this work are shown in Tab.~\ref{tab:licence_libs}.

\subsection{Datasets} \label{sec_sup:data}
The datasets used in this work are shown in Tab.~\ref{tab:licence_data}.

\subsection{Pretrained Models} \label{sec_sup:pretrained}
The pretrained models used in this work are shown in Tab.~\ref{tab:pretrained}.

\begin{table*}[h]
\centering
\caption{Libraries used in the paper}\label{tab:licence_libs}
\begin{adjustbox}{width=0.8\linewidth}
\begin{tabular}{@{} cc}
\toprule
\textbf{Library} & \textbf{Link to license}  \\
\midrule
Pytorch~\citep{paszke2019pytorch} & \url{https://github.com/pytorch/pytorch/blob/main/LICENSE} \\
Pytorch3D~\citep{ravi2020accelerating} & \url{https://github.com/facebookresearch/pytorch3d/blob/main/LICENSE} \\
Diffusers~\citep{vonplaten2022diffusers} & \url{https://github.com/huggingface/diffusers/blob/main/LICENSE} \\
\bottomrule
\end{tabular}
\end{adjustbox}
\end{table*}

\begin{table*}[h]
\centering
\caption{Datasets used in the paper}\label{tab:licence_data}
\begin{adjustbox}{width=0.95\linewidth}
\begin{tabular}{@{} cc}
\toprule
\textbf{Dataset} & \textbf{Link to license}  \\
\midrule
Hi4D~\citep{yin2023hi4d} & \url{https://hi4d.ait.ethz.ch} \\
CustomHumans~\citep{ho2023learning} & \url{https://custom-humans.ait.ethz.ch/} \\
THuman2.0~\citep{yu2021function4d} & \url{https://github.com/ytrock/THuman2.0-Dataset/blob/main/THUman2.1_Agreement.pdf} \\
MultiHuman~\citep{zheng2021deepmulticap} & \url{https://github.com/y-zheng18/MultiHuman-Dataset} \\
\bottomrule
\end{tabular}
\end{adjustbox}

\end{table*}

\begin{table*}[!h]
\centering
\caption{Pretrained Models used in the paper}\label{tab:pretrained}
\begin{adjustbox}{width=0.95\linewidth}
\begin{tabular}{@{} cc}
\toprule
\textbf{Pretrained model} & \textbf{Link to license}  \\
\midrule
Stable Diffusion 2.1 Unclip~\citep{rombach2022high} & \url{https://huggingface.co/stabilityai/stable-diffusion-2/blob/main/LICENSE-MODEL} \\
PSHuman~\citep{li2024pshuman} & \url{https://github.com/pengHTYX/PSHuman/blob/main/LICENSE.txt} \\
ControlNet~\citep{zhang2023adding} & \url{https://github.com/lllyasviel/ControlNet/blob/main/LICENSE} \\
CodeFormer~\citep{zhou2022towards} & \url{https://github.com/sczhou/CodeFormer/blob/master/LICENSE} \\
Face detector~\citep{Deng2020RetinaFace} & \url{https://github.com/serengil/retinaface/blob/master/LICENSE} \\
\bottomrule
\end{tabular}
\end{adjustbox}

\end{table*}

\section{Limitations, Impact and Safeguards}  \label{sec_sup:limitations}

\subsection{Limitations}  \label{sec_sup:limitation}
While our approach demonstrates strong performance across various scenarios, we acknowledge several aspects that offer room for future improvement.

First, our method is trained under ambient lighting assumptions with consistent illumination across multiple views. In certain challenging cases such as low-light scenes or strong lighting contrast, minor failures may occur, as shown in the top-left of Fig. 4 in the main paper, the back side of the person appears relatively dark. We note that our method is a proof of concept, and these issues can potentially be mitigated through data augmentation, more diverse training data, or incorporation of synthetic lighting variations.

Second, our method focuses on inter-human occlusion but does not yet explicitly model object-induced occlusions. In cases where a person is holding or interacting with a object, the model may fail to recognize the object as separate and instead reconstruct it as part of the body, resulting in distorted geometry (see bottom-right of Fig. 4 in the main paper). We plan to address this limitation in future work by incorporating object-aware reasoning.

Third, in cases where the model relies on predicted SMPL-X input, errors in the pose estimation can lead to discrepancies between the input image and the reconstructed mesh. The model may tend to follow the predicted SMPL-X pose, resulting in slightly misaligned geometry with the input image. Nonetheless, as shown in Tab.~\ref{tab:smplx} and Fig.~\ref{fig_supp_ablation_smplx}, our method remains robust under such conditions and outperforms existing baselines.

Finally, there is currently no publicly available baseline that directly matches our problem setting—multi-human reconstruction from a single image. To allow meaningful comparisons, we carefully adapted related methods across different input modalities (e.g., single-human or multi-view approaches). Although these comparisons are not perfectly aligned, they offer reasonable context. We also note that relevant recent work such as ~\citep{cha20243d} was not included due to the lack of released implementation.
\subsection{Impact and Safeguards} \label{sec_sup:impact}
This work can have significant potential across fields such as virtual reality, gaming, telepresence, digital fashion, and medical imaging. However, the ability to generate lifelike 3D representations from minimal input raises important ethical concerns around consent, data ownership, and control over digital likenesses. Moreover, the generated normal maps or 3D mesh can be used to infer sensitive biological data of the individual. It is therefore essential to limit access to the model through controlled licensing agreements and establish guidelines centered on the consent of the input image provider to minimize these concerns. 



\end{document}